\documentclass{article}

\usepackage{amsmath, amsthm, amssymb}

\usepackage{expl3}  %
\ExplSyntaxOn
  _to_str:N
\ExplSyntaxOff
\usepackage{ifthen}
\usepackage{pgffor}

\usepackage[acronym,nowarn,section,nonumberlist]{glossaries}
\glsdisablehyper  %
\makeglossaries

\setacronymstyle{long-short}

\newacronym{auc}{AUC}{Area Under the Curve}

\newacronym{bert}{BERT}{Bidirectional Encoder Representations from Transformers}
\newacronym{bow}{BOW}{Bag of Words}
\newacronym{boe}{BOE}{Bag of Embeddings}
\newacronym{borep}{BOREP}{Bag of Random Embedding Projections}
\newacronym{bn}{BN}{Bayesian Network}

\newacronym{cdf}{CDF}{Cumulative Distribution Function}
\newacronym{cnn}{CNN}{Convolutional Neural Network}
\newacronym{cka}{CKA}{Centered Kernel Alignment}
\newacronym{cvae}{CVAE}{Conditional Variational Auto-Encoder}

\newacronym{dag}{DAG}{Directed Acyclic Graph}
\newacronym{dgm}{DGM}{Directed Graphical Model}
\newacronym{dnn}{DNN}{Deep Neural Network}

\newacronym{ecfp4}{ECFP4}{Extended Connectivity Fingerprint}
\newacronym{ehr}{EHR}{Electronic Health Record}
\newacronym{esn}{ESN}{Echo State Network}

\newacronym{fft}{FFT}{Fast Fourier Transform}
\newacronym{fs}{FS}{FastSent}
\newacronym{ft}{FT}{Fourier Transform}
\newacronym{flashAttn}{\textsc{FlashAttention}}{Flash Attention}

\newacronym{gat}{GAT}{Graph Attention Network}
\newacronym{gcn}{GCN}{Graph Convolutional Network}
\newacronym{gdl}{GDL}{Geometric Deep Learning}
\newacronym{ggnn}{GGNN}{Gated Graph Neural Network}
\newacronym{glove}{GloVe}{Global Vectors for Word Representation}
\newacronym{gnn}{GNN}{Graph Neural Network}
\newacronym{gru}{GRU}{Gated Recurrent Unit}

\newacronym{hmm}{HMM}{Hidden Markov Model}

\newacronym{idf}{IDF}{Inverse Document Frequency}
\newacronym{ig}{IG}{Information Gain}
\newacronym{isan}{ISAN}{Input Switched Affine Network}

\newacronym{jsd}{JSD}{Jensen-Shannon Divergence}

\newacronym{kld}{KLD}{Kullback-Leibler Divergence}
\newacronym{knn}{KNN}{$K$-Nearest Neighbour}

\newacronym{lhs}{L.H.S.}{left hand side}
\newacronym{lstm}{LSTM}{Long Short Term Memory Network}

\newacronym{map}{MAP}{Maximum a Posteriori}
\newacronym{mc}{MC}{Monte Carlo}
\newacronym{mdl}{MDL}{Minimum Description Length}
\newacronym{ml}{ML}{Machine Learning}
\newacronym{mlp}{MLP}{Multi-Layer Perceptron}
\newacronym{mle}{MLE}{Maximum Likelihood Estimation}
\newacronym{mnist}{MNIST}{Modified National Institute of Standards and Technology}

\newacronym{nlp}{NLP}{Natural Language Processing}
\newacronym{nn}{NN}{Neural Network}
\newacronym{nll}{NLL}{Negative Log Likelihood}

\newacronym{ode}{ODE}{Ordinary Differential Equation}
\newacronym{oh}{OH}{One-Hot}

\newacronym{pdf}{PDF}{Probability Density Function}
\newacronym{pgm}{PGM}{Probabilistic Graphical Model}

\newacronym{relu}{ReLU}{Rectified Linear Unit}
\newacronym{rdf}{RDF}{Resource Description Framework}
\newacronym{rgb}{RGB}{Red Green Blue}
\newacronym{rgc}{RGC}{Relational Graph Convolution}
\newacronym{rgat}{RGAT}{Relational Graph Attention Network}
\newacronym{rgcn}{RGCN}{Relational Graph Convolutional Network}
\newacronym{rhs}{R.H.S.}{Right Hand Side}
\newacronym{rnn}{RNN}{Recurrent Neural Network}
\newacronym{roc}{ROC}{Receiver Operating Characteristic}

\newacronym{sgd}{SGD}{Stochastic Gradient Descent}
\newacronym{sg}{SG}{SkipGram}
\newacronym{simclr}{SimCLR}{Simple framework for Contrastive Learning of visual Representations}
\newacronym{st}{ST}{SkipThought}
\newacronym{sts}{STS}{Semantic Textual Similarity}
\newacronym{softmaxattn}{\texttt{SoftmaxAttn}}{Softmax Dot Product Attention}
\newacronym{termfreq}{TF}{Term Frequency}
\newacronym{tfidf}{TF-IDF}{Term Frequency-Inverse Document Frequency}
\newacronym[plural=TPPs, descriptionplural={Temporal Point Processes}]{tpp}{TPP}{Temporal Point Process}

\newacronym{vae}{VAE}{Variational Auto-Encoder}

\newacronym{wirgat}{WIRGAT}{Within-Relation Graph Attention}
\newacronym{wl}{WL}{Weisfeiler-Lehman}

\usepackage{bm}
\usepackage[T1]{fontenc}  %
\usepackage[type1]{libertine}
\usepackage{mathtools}
\usepackage{tcolorbox}
\usepackage[makeroom]{cancel}   %
\usepackage{algorithm}
\usepackage{algpseudocode}

\usepackage{enumitem}

\usepackage{multirow}
\usepackage{array}
\usepackage{makecell}

\DeclareRobustCommand{\mathup}[1]{\begingroup\changegreek\mathrm{#1}\endgroup}
\DeclareRobustCommand{\mathbfup}[1]{\begingroup\changegreekbf\mathbf{#1}\endgroup}
\DeclareRobustCommand{\mathbit}[1]{\bm{\mathit{#1}}}

\DeclareMathAlphabet{\mathsfit}{\encodingdefault}{\sfdefault}{m}{sl}
\SetMathAlphabet{\mathsfit}{bold}{\encodingdefault}{\sfdefault}{bx}{n}
\newcommand{\tens}[1]{\bm{\mathsfit{#1}}}

\newcommand{\constantvector}{\bm}               %

\newcommand{\constantmatrix}{\bm}               %
\newcommand{\constantmatrixgreek}{\mathbit}

\newcommand{\randomscalar}{\textnormal}         %
\newcommand{\randomscalargreek}{\mathup}

\newcommand{\randomvector}{\mathbf}             %
\newcommand{\randomvectorgreek}{\mathbfup}

\newcommand{\randommatrix}{\mathbf}             %
\newcommand{\randommatrixgreek}{\mathbfup}

\newcommand{\graphstyle}{\mathcal}              %

\newcommand{\tensorstyle}{\tens}                %

\newcommand{\setstyle}{\mathbb}                %

\definecolor{lightgray}{rgb}{220,219,217}
\definecolor{eggshell}{rgb}{0.94, 0.92, 0.84}
\definecolor{lightblue}{rgb}{0.68, 0.91, 1.00}
\definecolor{applelightblue}{rgb}{0.83, 0.90, 0.94}

\def\alphabet{a,b,c,d,e,f,g,h,i,j,k,l,m,n,o,p,q,r,s,t,u,v,w,x,y,z}
\def\Alphabet{A,B,C,D,E,F,G,H,I,J,K,L,M,M,O,P,Q,R,S,T,U,V,W,X,Y,Z}
\def\greekalphabet{alpha,beta,gamma,delta,epsilon,varepsilon,zeta,eta,theta,vartheta,iota,kappa,varkappa,lambda,mu,nu,xi,pi,varpi,rho,varrho,sigma,varsigma,tau,upsilon,phi,varphi,chi,psi,omega}
\def\GreekAlphabet{Gamma,Delta,Theta,Lambda,Xi,Pi,Sigma,Upsilon,Phi,Psi,Omega}

\makeatletter
\def\changegreek{\@for\next:=\greekalphabet
	\do{\expandafter\let\csname\next\expandafter\endcsname\csname\next up\endcsname}}
\def\changegreekbf{\@for\next:=\greekalphabet
	\do{\expandafter\def\csname\next\expandafter\endcsname\expandafter{%
			\expandafter\bm\expandafter{\csname\next up\endcsname}}}}
\makeatother

\foreach \x in \alphabet {
	\expandafter\xdef\csname v\x\endcsname{\noexpand\ensuremath{\noexpand\constantvector{\x}}}

	\expandafter\xdef\csname ev\x\endcsname{\noexpand\ensuremath{\noexpand\x}}

	\ifthenelse{\equal{\x}{m}}{}{
		\expandafter\xdef\csname r\x\endcsname{\noexpand\ensuremath{\noexpand\randomscalar{\x}}}
	}

	\expandafter\xdef\csname rv\x\endcsname{\noexpand\ensuremath{\noexpand\randomvector{\x}}}
}

\foreach \x in \greekalphabet {
	\expandafter\xdef\csname v\x\endcsname{\noexpand\ensuremath{\noexpand\constantvector{\csname \x\endcsname}}}

	\expandafter\xdef\csname ev\x\endcsname{\noexpand\ensuremath{\noexpand{\csname \x \endcsname}}}

	\expandafter\xdef\csname r\x\endcsname{\noexpand\ensuremath{\noexpand\randomscalargreek{\csname \x\endcsname}}}

	\expandafter\xdef\csname rv\x\endcsname{\noexpand\ensuremath{\noexpand\randomvectorgreek{\csname \x\endcsname}}}
}

\foreach \x in \Alphabet {
	\expandafter\xdef\csname m\x\endcsname{\noexpand\ensuremath{\noexpand\constantmatrix{\x}}}

	\expandafter\xdef\csname em\x\endcsname{\noexpand\ensuremath{\noexpand\x}}

	\expandafter\xdef\csname rm\x\endcsname{\noexpand\ensuremath{\noexpand\randommatrix{\x}}}

	\expandafter\xdef\csname t\x\endcsname{\noexpand\ensuremath{\noexpand\tensorstyle{\x}}}

	\expandafter\xdef\csname g\x\endcsname{\noexpand\ensuremath{\noexpand\graphstyle{\x}}}

	\ifthenelse{\equal{\x}{E}}{}{
		\expandafter\xdef\csname s\x\endcsname{\noexpand\ensuremath{\noexpand\setstyle{\x}}}
	}

}

\foreach \x in \GreekAlphabet {
	\expandafter\xdef\csname m\x\endcsname{\noexpand\ensuremath{\noexpand\constantmatrixgreek{\csname \x\endcsname}}}

	\expandafter\xdef\csname rm\x\endcsname{\noexpand\ensuremath{\noexpand\randommatrixgreek{\csname \x\endcsname}}}
}

\DeclareMathOperator*{\softmax}{Softmax}

\DeclarePairedDelimiter{\ceil}{\lceil}{\rceil}

\newcommand{\sigmoid}{\sigma}

\newcommand{\bb}[1]{{\boldsymbol{#1}}}

\def\eps{{\epsilon}}

\newcommand{\attn}{\mathrm{Attn}}
\newcommand{\sigmoidattn}{\mathrm{SigmoidAttn}}
\newcommand{\softmaxattn}{\mathrm{SoftmaxAttn}}

\newcommand{\reading}[2]{{#1}{\pm{#2}}}

\usepackage{titletoc}
\usepackage[page,header]{appendix}
\usepackage{setspace}
\noappendicestocpagenum

\usepackage{microtype}
\usepackage{graphicx}
\usepackage{subcaption}
\usepackage{booktabs} %

\usepackage{xcolor}
\definecolor{lm_core_en}{rgb}{0.7137,0.3333,0.3333}
\definecolor{lm_code}{rgb}{0.4118,0.6431,0.4314}
\definecolor{lm_math}{rgb}{0.3254,0.4431,0.6666}

\usepackage{hyperref}
\hypersetup{
    colorlinks,
    linkcolor={black!10!black},
    citecolor={blue!50!black},
    urlcolor={blue!80!black}
}

\usepackage[preprint]{neurips_2024}

\usepackage{amsmath}
\usepackage{amssymb}
\usepackage{mathtools}
\usepackage{amsthm}

\usepackage{adjustbox}

\usepackage{titlesec}
\titlespacing{\paragraph}{0pt}{0pt}{1em} %

\usepackage[nameinlink]{cleveref}
\usepackage{enumitem}
\setlist[enumerate,1]{label=(\arabic*),ref=(\arabic*)}
\setlist[enumerate,2]{label=(\alph*),ref=(\arabic{enumi}-\alph*)}
\crefname{enumi}{Step}{Steps}     %
\crefformat{equation}{(#2#1#3)} %
\crefmultiformat{equation}{(#2#1#3)}%
{ and~(#2#1#3)}{, (#2#1#3)}{ and~(#2#1#3)}

\crefname{section}{Sec.}{Sec.}
\Crefname{section}{Section}{Sections}
\crefname{appendix}{App.}{App.}
\Crefname{appendix}{Appendix}{Appendices}
\crefname{table}{Tab.}{Tab.}
\Crefname{table}{Table}{Tables}
\crefname{figure}{Fig.}{Fig.}
\Crefname{figure}{Figure}{Figures}
\crefname{algorithm}{Alg.}{Alg.}
\Crefname{algorithm}{Algorithm}{Algorithms}
\crefname{theorem}{Thm.}{Thm.}
\Crefname{theorem}{Theorem}{Theorems}
\crefname{lemma}{Lemma}{Lemmas}
\Crefname{lemma}{Lemma}{Lemmas}
\crefname{definition}{Def.}{Def.}
\Crefname{definition}{Definition}{Definitions}
\crefname{remark}{Rmk.}{Rmk.}
\Crefname{remark}{Remark}{Remarks}

\theoremstyle{plain}
\newtheorem{theorem}{Theorem}[section]
\newtheorem{proposition}[theorem]{Proposition}
\newtheorem{lemma}[theorem]{Lemma}

\theoremstyle{definition}
\newtheorem{definition}[theorem]{Definition}

\theoremstyle{remark}

\usepackage{multirow}
\usepackage{multicol}

\usepackage[textsize=tiny]{todonotes}

\title{Theory, Analysis, and Best Practices for \\ Sigmoid Self-Attention}

\newcommand*\samethanks[1][\value{footnote}]{\footnotemark[#1]}
\author{
  Jason Ramapuram\thanks{Primary contributor. For a detailed breakdown of author contributions see \Cref{sec:attribution}.}
  \And
  Federico Danieli\samethanks
  \And 
  Eeshan Dhekane\samethanks \And
  Floris Weers\samethanks \And
  Dan Busbridge\samethanks \And
  Pierre Ablin\samethanks \And  
  Tatiana Likhomanenko\samethanks \AND
  Jagrit Digani \And
  Zijin Gu \And
  Amitis Shidani \And
  Russ Webb \And
  \vspace{-0.5cm} \\ \phantom{Mystery} \\
  Apple \\ \\
  \footnotesize\texttt{\{jramapuram, f\_danieli, eeshan, floris\_weers, dbusbridge,} \\
  \footnotesize\texttt{p\_ablin, antares, digani, zgu26, amitis\_shidani, rwebb\}@apple.com} \\
\\
}

\begin{document}

\maketitle

\vspace{-0.3in}
\begin{abstract}
    Attention is a key part of the transformer architecture. It is a sequence-to-sequence mapping that transforms each sequence element into a weighted sum of values. The weights are typically obtained as the softmax of dot products between keys and queries. Recent work has explored alternatives to softmax attention in transformers, such as ReLU and sigmoid activations. In this work, we revisit sigmoid attention and conduct an in-depth theoretical and empirical analysis. Theoretically, we prove that transformers with sigmoid attention are universal function approximators and benefit from improved regularity compared to softmax attention. Through detailed empirical analysis, we identify stabilization of large initial attention norms during the early stages of training as a crucial factor for the successful training of models with sigmoid attention, outperforming prior attempts. We also introduce \textsc{FlashSigmoid}, a hardware-aware and memory-efficient implementation of sigmoid attention yielding a \emph{17\%} inference kernel speed-up over \textsc{FlashAttention2} on H100 GPUs \footnote{Code is available at \url{https://github.com/apple/ml-sigmoid-attention}.
    }. Experiments across language, vision, and speech show that properly normalized sigmoid attention matches the strong performance of softmax attention on a wide range of domains and scales, which previous attempts at sigmoid attention were unable to fully achieve. Our work unifies prior art and establishes best practices for sigmoid attention as a drop-in softmax replacement in transformers.

\end{abstract}

\section{Introduction}
\label{sec:intro}

The success of modern machine learning can be largely attributed to the attention mechanism \citep{DBLP:journals/corr/BahdanauCB14,DBLP:conf/nips/VaswaniSPUJGKP17}. Attention uses a sequence-to-sequence (seq-to-seq) map to build context-aware token representations.
Classically, attention relies on the softmax function ($\softmaxattn$) to recover token representations as data-dependent convex combinations of values.

Despite its widespread use and effectiveness, softmax in $\softmaxattn$ is not without limitations. For instance, the softmax function can sometimes lead to a concentration of attention on just a few features \citep{DBLP:conf/iclr/YangDSC18,DBLP:conf/icml/GaneaGBS19}, potentially neglecting other informative aspects of the input data. Moreover, applying $\softmaxattn$ requires performing a \textit{row-wise} reduction along the length of the input sequence, which in the case of efficient attention kernels \citep{DBLP:conf/nips/DaoFERR22,DBLP:journals/corr/abs-2307-08691}, slows down computations.
In this work, we relax this constraint by substituting the \textit{row-wise} softmax operation with an \textit{element-wise} sigmoid nonlinearity. We highlight that the central problem with na\"ive sigmoid attention ($\sigmoidattn$) is that of large initial attention norms and propose solutions to alleviate it. \textbf{Our contributions are as follows:}

\begin{enumerate}[itemsep=0pt,leftmargin=*]
    \item We prove $\sigmoidattn$ is a universal function approximator on seq-to-seq tasks (\cref{sec:ufa}).
    \item We analyze $\sigmoidattn$'s regularity and provide its worst-case Jacobian bound (\cref{sec:regularity}).
    \item We extend \textsc{FlashAttention2} \citep{DBLP:conf/nips/DaoFERR22,DBLP:journals/corr/abs-2307-08691} with the sigmoid kernel, reducing kernel inference wall-clock time by up to 17\% and real world inference by up to 8\% (\cref{sec:FlashSigmoidHardwareAwareImplementation}).
    \item We show that $\sigmoidattn$ matches $\softmaxattn$ in various tasks and domains (\cref{sec:experiments}).
\end{enumerate}

\section{Sigmoid Attention}
\label{sec:methods}
Let $\mX \in \mathbb{R}^{n \times d}$ be the input sequence of $n$ vectors, where each vector has dimension $d$. We define three learnable weight matrices $\mW_q \in \mathbb{R}^{d \times d_{qk}}$, $\mW_k \in \mathbb{R}^{d \times d_{qk}}$, and $\mW_v \in \mathbb{R}^{d \times d_v}$, which are used to compute the queries $\mQ \in \mathbb{R}^{n \times d_{qk}}$, keys $\mK \in \mathbb{R}^{n \times d_{qk}}$, and values $\mV \in \mathbb{R}^{n \times d_v}$ as follows:
\begin{equation}
\mQ = \mX \mW_q, \quad \mK = \mX \mW_k, \quad \text{and} \quad \mV = \mX \mW_v.
\end{equation}
Self-attention~\citep{DBLP:journals/corr/BahdanauCB14,DBLP:conf/nips/VaswaniSPUJGKP17} can be compactly written as
\begin{equation}
\label{eq:attn_short}
\softmaxattn(\mX) = \softmax(\mQ \mK^T / \sqrt{d_{qk}}) \mV,
\end{equation}
where the $\softmax$ function \textit{normalizes each row }of the input matrix.
We  replace the $\softmax$ with
\begin{tcolorbox}[colback=applelightblue, colframe=black, boxrule=1pt, arc=5mm, boxsep=1mm, left=0mm, top=0mm, right=2mm, valign=center]
\begin{align}
\begin{split}
    \label{eq:sigmoid_attn}
    \sigmoidattn(\mX) = \sigma(\mQ\mK^T / \sqrt{d_{qk}})\mV,\\
    \text{with }\sigma:u\mapsto \mathrm{sigmoid}(u + b)\coloneqq (1+e^{-(u+b)})^{-1}.
\end{split}
\end{align}
\end{tcolorbox}
Here, $\sigma$ is applied \textit{element-wise} to the input matrix in \cref{eq:sigmoid_attn}.
The activation function $\sigma$ has a hyper-parameter $b\in\mathbb{R}$. In \cref{app:sigmoid_bias}, we discuss an intuitive way to choose the order-optimal bias term, resulting in $b = -\log(n)$.
This choice of $b$ allows us to make sense of $\sigmoidattn$ for any sequence length.
Indeed, letting $(\vy_1, \dots, \vy_n) = \sigmoidattn(\mX)$ be the output sequence, we have
\begin{equation}
    \label{eq:sigmoid_attn_sequence}
    \vy_i = \sum_{j=1}^n \frac{\exp(\langle \mW_q\vx_i, \mW_k\vx_j\rangle)}{\exp(\langle \mW_q\vx_i, \mW_k\vx_j\rangle) + n}\mW_v\vx_j
    \xrightarrow[n\to+\infty]{} \int \exp(\langle \mW_q\vx_i, \mW_k\vx \rangle)\mW_v\vx d\mu(\vx),
\end{equation}
where 
$\mu = \frac1n\sum_{j=1}^n\delta_{\vx_j}$ is the empirical measure corresponding to $\mX$.
Notably, \cref{eq:sigmoid_attn_sequence} still makes sense in the infinite length limit, where the measure $\mu$ is not a sum of Diracs. \citet{wortsman2023replacing} do not use a bias, and propose a $n^{-1}$ normalization for various attention activations, such as sigmoid and ReLU, but leave the reason as an open question.
Our variable bias has a similar effect in the large $n$ limit, and we posit that recovering a finite output limit as $n$ increases is the why it works in practice.

A multi-head version of \cref{eq:sigmoid_attn} is obtained by combining the outputs of several $\sigmoidattn$, as follows:
\begin{equation}
    \left[\sigmoidattn_1(\bb{X}),\dots,\sigmoidattn_h(\bb{X})\right]\bb{W}_o,
\end{equation}
for a learnable output weight matrix $\bb{W}_o\in\mathbb{R}^{hd_v\times d}$, where $h$ denotes the number of \emph{heads}.

\section{Theoretical Properties of Sigmoid Attention}
\label{sec:theory}
We analyze $\sigmoidattn$, with two objectives: (1) showing that a transformer architecture remains a universal function approximator when $\sigmoidattn$ replaces $\softmaxattn$, and (2) recovering a measure of regularity of $\sigmoidattn$ by computing its Lipschitz constant.
\subsection{Are Transformers with Sigmoid Attention Universal Approximators?}
\label{sec:ufa}
\cite{Yun_UAP} demonstrate that classical transformers can approximate continuous sequence-to-sequence functions to arbitrary precision, a property known as the \emph{Universal Approximation Property} (UAP). UAP is highly desirable as it provides proof of an architecture's generalizability and representation capability.
As $\sigmoidattn$ modifies the transformer architecture, it is crucial to theoretically guarantee that this modification does not impact the representation capability and that UAP is retained. We provide this guarantee with the following theorem.
\begin{theorem}[UAP for $\sigmoidattn$]
    \label{thm::UAP}
    We denote with $\mathcal{T}^{h,d_v,r}_{\sigma}$ the class of transformer networks obtainable by combining an arbitrary number of $\sigmoidattn$ layers (each of $h$ heads of dimension $d_v$) followed by FFN layers of hidden dimension $r$.
    For any given continuous, permutation-equivariant function $f:\Omega\subset\mathbb{R}^{n\times d}\to\mathbb{R}^{n\times d}$ with compact support $\Omega$, and for any arbitrarily small error $\varepsilon$, there exists a transformer network $g\in\mathcal{T}_\sigma^{4,1,4}$ such that
    \begin{equation}
        \left(\int_{\Omega}\|f(\bb{X})-g(\bb{X})\|^p_p d\bb{X}\right)\leq\varepsilon,\qquad\text{for}\quad 1\leq p<\infty.
    \end{equation}
\end{theorem}
\Cref{thm::UAP} is the exact counterpart of \cite[Thm.~2]{Yun_UAP}, which shows UAP for classical transformers. Our proof largely follows the same path, an outline of the original proof provided in \cref{app:UAP_proof}. Here, we present an overview of the main adaptations required to prove \cref{thm::UAP} for $\sigmoidattn$, with further details in \cref{sec::proof_modified_sigmoid,sec::proof_contextual_mapping_top}.

\paragraph{Sigmoid Attention layers can implement contextual mappings:} A key step in proving \cref{thm::UAP} is showing that, even with $\sigmoidattn$, a sequence of transformer blocks can implement a \emph{Contextual Mapping} \cite[Def.~3.1]{Yun_UAP}. A contextual mapping characterizes a function that maps each input sequence element to an output \emph{uniquely} dependent on the \emph{whole} sequence. This property allows a transformer to capture and store global context within each token, even if each layer only performs pairwise comparisons. Subsequent layers can then use this global information to map individual tokens to the correct output, ultimately approximating any arbitrary sequence-to-sequence function.

In \cite{Yun_UAP}, the contextual mapping is assembled by modifying individual transformer blocks: each block is tuned to react to a specific input token. By stacking a sequence of these blocks, a transformer can be turned into an accumulator, mapping a given input token sequence to a unique global index. This outcome is achieved via a \emph{selective shift layer} \cite[App.~B.5]{Yun_UAP}:
\begin{equation}
    \Psi(\bb{X};b,b')_{i,1}\coloneqq \begin{cases}
        \max_k \bb{X}_{k,1}-\min_k\bb{X}_{k,1}&\text{if}\quad b<\bb{X}_{i,1}<b'\\
        0&\text{otherwise},
    \end{cases}
    \label{eqn::shift_operation_original}
\end{equation}
and can be approximated using classic attention.
Although $\sigmoidattn$ cannot directly approximate~\cref{eqn::shift_operation_original}, our accumulator definition relies on an equivalent selective shift operation:
\begin{equation}
    \Psi_\sigma(\bb{X};b,b')_{i,1}\coloneqq\begin{cases}
        \sum_{k:\bb{X}_{k,1}> b'} \bb{X}_{k,1} &\text{if}\quad b<\bb{X}_{i,1}<b' \\
        0 &\text{otherwise},
    \end{cases}
    \label{eqn::shift_operation_ours}
\end{equation}
which can be approximated by $\sigmoidattn$ (described in \cref{sec::proof_modified_sigmoid}). In~\cref{sec::proof_contextual_mapping}, we show that~\cref{eqn::shift_operation_ours} shares similar properties with~\cref{eqn::shift_operation_original}, allowing us to use the original proof framework in \cite{Yun_UAP} and demonstrate that UAP holds in our case as well.

Our proof is largely equivalent to that in \cite{Yun_UAP}, with two relevant differences: to approximate \cref{eqn::shift_operation_ours}, we require $\sigmoidattn$ with \textit{at least four heads} and shifts included in both query and key definitions. In contrast, $\softmaxattn$ requires \textit{at least two heads} to approximate~\cref{eqn::shift_operation_original}, with shifts only in the query definition. However, this is primarily a theoretical requirement for the proof and does not affect performance. Notably, the total number of parameters required by both architectures for the approximation follows the same tight scaling of \cite{Yun_UAP}.

\subsection{Regularity of Sigmoid Attention}
\label{sec:regularity}
As with any layer in a neural network, the regularity of $\sigmoidattn$ is important to study, as it gives insights into the robustness of the corresponding network and the ease of optimizing it.
The most standard way to quantify the regularity of a layer function $\phi$ is to compute its \emph{Lipschitz constant} over a set $\mathcal{X}$, that is a constant $C>0$ such that for all $\mX, \mY\in \mathcal{X}$, it holds $\|\phi(\mX) - \phi(\mY)\|\leq C \|\mX - \mY\|$, where $\|\cdot\|$ is the standard Frobenius norm.
The \emph{local} Lipschitz constant is the spectral norm of the Jacobian of $\phi$ at $\mX$.
The two are related: the Lipschitz constant of $\phi$ over $\mathcal{X}$ is the greatest local Lipschitz constant for all $\mX\in \mathcal{X}$.
We turn to the theorem giving the regularity of $\sigmoidattn$:
\begin{theorem}
\label{thm:regularity}
    Define $A = \{\langle \mW_q \vx_i \mW_k \vx_j\rangle|,\enspace i, j\in \{1,\dots,n\}\}\subset\mathbb{R}$ the set of attention weights,  and the scaled activation norms $\sigma_{\infty} = n\times\sup_{u\in A} |\sigma(u)|$ and $\sigma'_{\infty} = n\times \sup_{u\in A} |\sigma'(u)|$.
    Then, the Jacobian of $\sigmoidattn$ at $\mX = (\vx_1, \dots, \vx_n)$ has a spectral norm of at most:
    \begin{equation}
        \|\mW_v\|_2\left(\sigma_{\infty} + 2\sigma'_{\infty} \|\mW_q^T \mW_k\|_2\left(\frac1n\sum_{i=1}^n\|\vx_i\|_2^2\right)\right).
    \end{equation}
\end{theorem}
The proof is found in \cref{app:lipschitz_proof}.
In $\sigmoidattn$, if we assume that the attention weights $\langle \mW_q \vx_i, \mW_k \vx_j\rangle$ are all bounded by a constant $\mu$ --- this is true, e.g., if the activations are bounded --- we get $\sigma_{\infty}\leq \exp(\mu)$ and $\sigma'_{\infty}\leq\exp(\mu)$ thanks to the choice of $b = -\log(n)$.
The bound in \cref{thm:regularity} depends only on the \emph{average} squared-norm of the input sequence $\vx_i$, while classical results for the study of attention all rely on the largest value of $\|\vx_i\|^2_2$~\citep{kim2021lipschitz,castin2023understanding}. 
This is another consequence of the simplicity of sigmoid attention and is due to the removal of the normalizing constant in $\softmaxattn$.
Our result implies that if all $\vx_i$ are within a ball of radius $R$ then the Lipschitz constant of $\sigmoidattn$ grows at most like $R^2$, but it is stronger since we can apply this to unbounded distributions $\vx_i$; it matters only that the second moment is bounded.
This result contrasts sharply with the bounds obtained for $\softmaxattn$: \citet[Thm.~3.4.]{castin2023understanding} show that there exists a sequence $\mX = (\vx_1, \dots, \vx_n)$ with $\|\vx_i\|_2\leq R$ for all $i$ such that the spectral norm of the Jacobian of $\attn$ at $\mX$ is at least $cR^2\exp(cR^2)$ for some constant $c>0$.
On the other hand, our bound scales in $R^2$: this means that the local Lipschitz constant of $\sigmoidattn$ is much lower than the worst local Lipschitz constant of $\softmaxattn$.
Note that this result does not inform us of the practical average case Lipschitz constant, which is likely to be much lower for both Softmax and Sigmoid attention.
Upper bounds on the Lipschitz constant of $\sigmoidattn$ are of particular interest to study the dynamics of attention, as done, e.g., in \citep{geshkovski2024emergence,geshkovski2023mathematical}

\subsection{Computational Complexity of Sigmoid and Softmax.}
\label{sec:parameter_and_flops}
\vspace{-0.1in}
\begin{table}[h!]
\small
\centering
\caption{Forward floating operations per token per attention head. 
$n_\text{ctx}$ and $d_\text{head}$ are the context length and head dimension respectively. 
$\Delta$ measures the compute difference between sigmoid and softmax. 
$c$ accounts for causal ($c=(n_\text{ctx}+1)/2n_\text{ctx}\sim1/2$), or standard ($c=1$) attention.
Typical values from the 1B LLM results are $n_\text{ctx}=2048$, $d_\text{head}=64$.
Sigmoid and softmax share the same number of floating operations (softmax: max-subtraction (2), exponentiation, summation, division; Sigmoid: bias-add, sign-flip, exponentiation, addition, division).
Remaining differences are due implementation details, and are subleading ($\sim1\%$) compared to other attention operations like computing attention logits $\mL$ (shown below). This analysis precludes hardware aware improvements (\Cref{sec:FlashSigmoidHardwareAwareImplementation}).
}
\label{tab:flop-counts}
\begin{tabular}{lcccc}
\toprule
& $\mL=\mQ\mK^T$ & $\softmax\left(\mL\right)$ & $\sigmoid\left(\mL+\vb\right)$ & $\Delta$ \\ \midrule 
Expression & $ 2\,c\,n_\text{ctx}\,d_{\text{head}}$ & $ 5\,c\,n_\text{ctx}$ & $ 5\,c\,n_\text{ctx}$ & $0$\\
\bottomrule
\end{tabular}
\end{table}

\section{\textsc{FlashSigmoid}: Hardware-Aware Implementation}
\label{sec:FlashSigmoidHardwareAwareImplementation}
\begin{figure}[!thbp]
    \centering
    \begin{minipage}{0.45\textwidth}
        \footnotesize
        \centering
        \includegraphics[trim={0 0 0 0}, width=\textwidth]{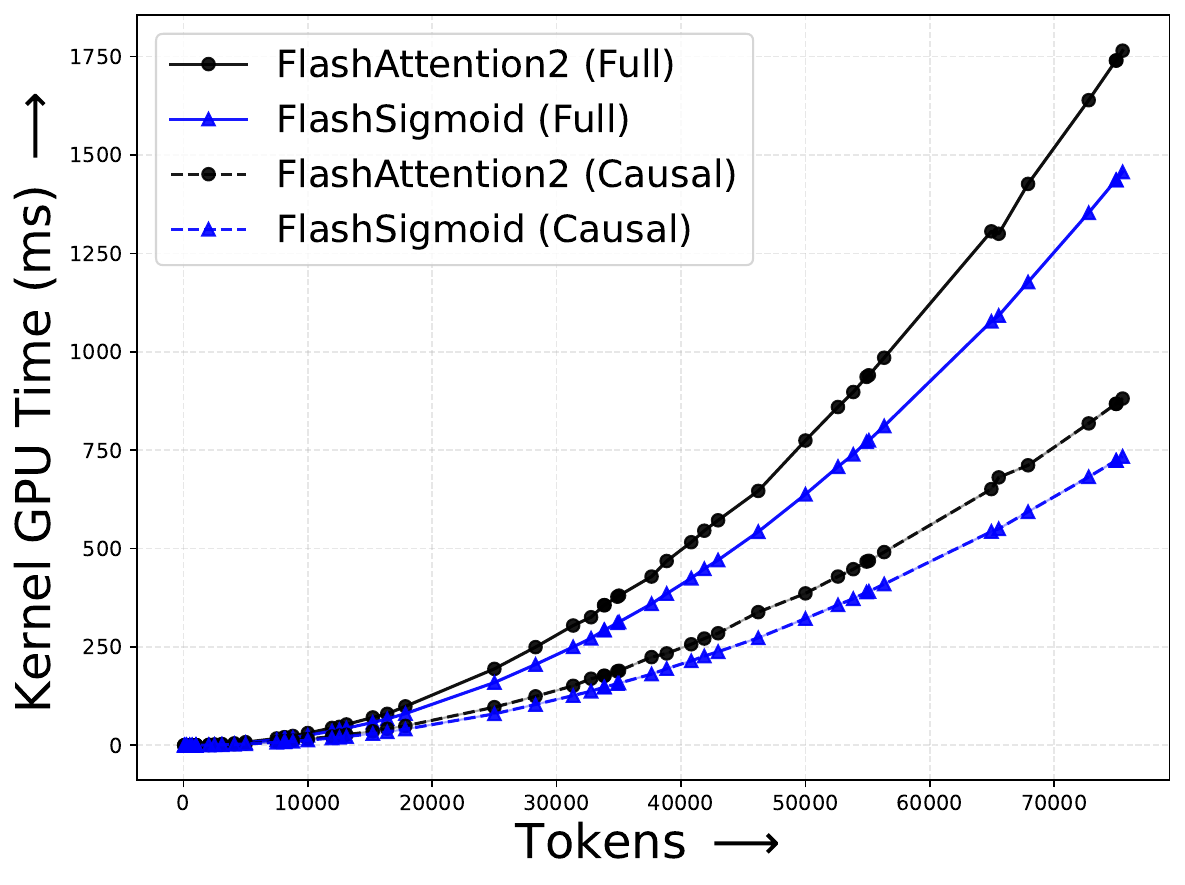}
        \captionsetup{justification=centering} 
        \caption*{(a) Inference mode kernels on H100.}
    \end{minipage}
    \hfill
    \begin{minipage}{0.45\textwidth}
        \centering        
        \includegraphics[trim={{0 0 0 0}}, width=\textwidth]{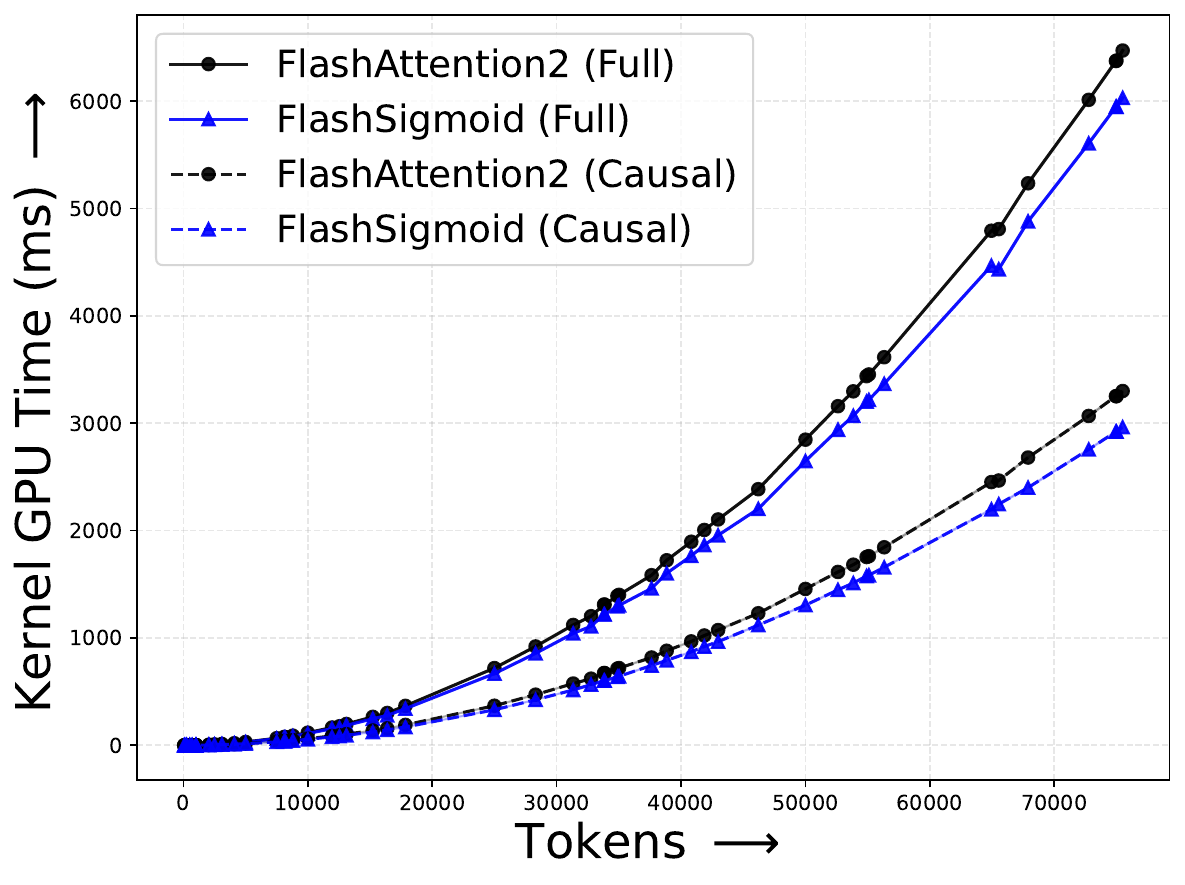}
        \captionsetup{justification=centering} 
        \caption*{(b) Training mode kernels on H100.}
    \end{minipage}
    \caption{
        Average kernel speed-up for \textsc{FlashSigmoid} over \textsc{FlashAttention2} for sequence lengths 64--78k. Inference is ${17.39}\%$ faster for self-attention and ${18.76}\%$ for causal attention. Training is $6.53\%$ faster for self-attention and $9.46\%$ for causal attention. 
    }
    \label{fig:h100-softmax-sigmoid-main-figure}
\end{figure}
Memory speed has not kept pace with recent gains in computation speed~\citep{DBLP:journals/micro/Choquette23,DBLP:conf/isca/JouppiYPPABBBBB17,mlx2023}.
Consequently, attention computations on modern architectures have been IO-bound by memory accesses \citep{DBLP:conf/mlsys/IvanovDB0H21}. 
\textsc{FlashAttention} \citep{DBLP:conf/nips/DaoFERR22} and \textsc{FlashAttention2} \citep{DBLP:journals/corr/abs-2307-08691} address these shortcomings by optimizing GPU memory hierarchy utilization to accelerate attention computations. 
Motivated by the speed boost provided by these approaches, we develop \textsc{FlashSigmoid}, a hardware-aware implementation of $\sigmoidattn$.
Like previous works, \textsc{FlashSigmoid} employs three core ideas:
\paragraph{Tiling: Divide and Conquer Approach to Attention:}\ Similar to \textsc{FlashAttention} and \textsc{FlashAttention2}, \textsc{FlashSigmoid} processes input parts in parallel to compute attention outputs in blocks, efficiently combining partial results to generate the final attention output.
\paragraph{Kernel Fusion:}\ Like \textsc{FlashAttention} and \textsc{FlashAttention2}, \textsc{FlashSigmoid} implements the computational steps of both forward and backward passes of $\sigmoidattn$ as single GPU kernels, minimizing memory accesses and improving memory efficiency by avoiding materialization of intermediate activations on High-Bandwidth Memory (HBM).
\paragraph{Activation Recomputation:}\ The backward pass of sigmoid attention requires the sigmoid activation matrix, which, if materialized on GPU HBM, results in slower implementation and memory inefficiencies. \textsc{FlashSigmoid} addresses this by retaining only query, key, and value tensors for re-computation of the sigmoid activation matrix during the backward pass. Despite increased FLOPs, this approach proves faster in wall-clock time as well as more memory-efficient than the alterantive approach of materializing and retaining the attention matrix.

The forward and backward pass algorithms of \textsc{FlashSigmoid} can be found in~\cref{sec:DetailsOfFlashSigmoidAlgorithm}.
Here, we highlight key differences between \textsc{FlashSigmoid} and \textsc{FlashAttention}/\textsc{FlashAttention2}. 
The point-wise nature of $\sigmoidattn$ results in a faster and more memory-efficient implementation by removing the need to compute the softmax normalization and materialize it to HBM. 
A reduction in the number of kernel dispatches also speeds up \textsc{FlashSigmoid}. 
Further, \textsc{FlashSigmoid} does not require accumulation and tracking of intermediate variables (row-sum and maximum of blocks) in the forward and backward passes which saves computation cost and reduces register pressure. 
We use $\textrm{sigmoid}\left(x\right) = 0.5\cdot\left(1 + \textrm{tanh}\left(0.5\cdot x\right)\right)$ to optimize the sigmoid computation on GPU. 
The speed up in \textsc{FlashSigmoid} compared to \textsc{FlashAttention} arises from optimizing hardware bottlenecks; theoretically, $\sigmoidattn$ is slower than $\softmaxattn$ (\cref{sec:parameter_and_flops}).

To measure the performance improvements of \textsc{FlashSigmoid}, we compare the timings of the kernels in its forward and backward passes against those of \textsc{FlashAttention2}.
The details of this benchmarking on H100 and A100 GPUs can be found in~\cref{sec:PerformanceAnalysisOfFlashSigmoidKernels}.
Measuring GPU computation time, we observe a $17.39\%$ speed-up during inference and a $6.53\%$ speed-up during training for attention over randomly initialized data on H100 GPU (\cref{fig:h100-softmax-sigmoid-main-figure}). 
In practice, these gains may be affected by other bottlenecks, such as movement of tensors between CPU or GPU memory, computations in other layers, and communication overhead in distributed training and inference. 
However, we demonstrate that \textsc{FlashSigmoid} speeds up training by {\textbf{$\sim$4\%} and inference by {\textbf{$\sim$8\%} in a realistic end-to-end setup. 
The details of wall-clock time improvements with~\textsc{FlashSigmoid} are in \cref{sec:SpeedBoostsOfFlashSigmoidInRealisticSettings}.
We also note that practical machine learning workflows are dominated by inference rather than training.

\section{Experiments}
\label{sec:experiments}
\begin{figure}[h]
\centering
\includegraphics[width=0.97\textwidth]{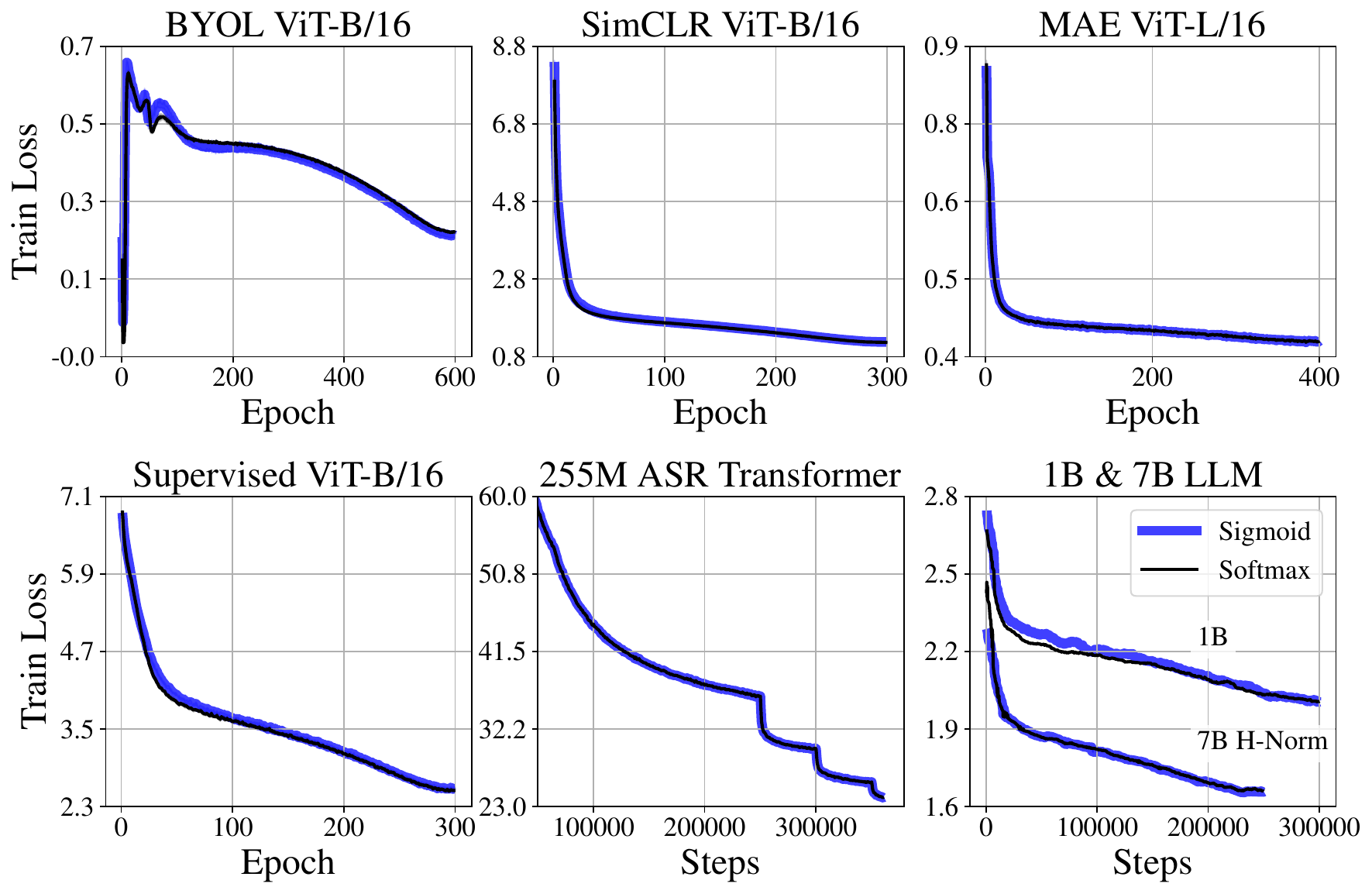}
\caption{Train losses comparing $\sigmoidattn$ with $\softmaxattn$.}
\label{fig:summary_nll}
\end{figure}
To empirically validate $\sigmoidattn$, we evaluate across several domains: supervised image classification using vision transformers \citep{DBLP:conf/iclr/DosovitskiyB0WZ21}, self-supervised image representation learning with SimCLR \citep{DBLP:conf/icml/ChenK0H20, DBLP:conf/icml/ZhaiLLBR0GS23}, Bootstrap Your Own Latent (BYOL) \citep{DBLP:conf/nips/GrillSATRBDPGAP20, DBLP:conf/nips/BusbridgeRALDCW23} and Masked AutoEncoders (MAE) \citep{DBLP:conf/cvpr/HeCXLDG22} as well as automatic speech recognition (ASR) \citep{synnaeve2019end,conformer} and auto-regressive language modeling (LM) \citep{DBLP:conf/nips/BrownMRSKDNSSAA20}. We also validate sequence length generalization on TED-LIUM v3~\citep{hernandez2018ted} for ASR and in small scale synthetic experiments in \cref{sec:a_se_pair_repeat_prob}.
Across all these domains and algorithms, we demonstrate that $\sigmoidattn$ matches the performance of $\softmaxattn$ (\cref{fig:summary_nll,fig:test_top1_results}), while offering training and inference speed-ups as highlighted in \cref{sec:FlashSigmoidHardwareAwareImplementation}. Empirically we make the following observations:
\begin{enumerate}[itemsep=0pt,leftmargin=*]
    \item $\sigmoidattn$ is effective for vision tasks without a bias (except MAE), but relies on LayerScale~\citep{touvron2021going} to match the performance of the baseline $\softmaxattn$ (\cref{fig:imagenet_top_1_ablations}-a) in a hyper-parameter free manner.\footnote{\Cref{sec:layerscale_free_sigmoid} demonstrates that supervised vision tasks using $\sigmoidattn$ without LayerScale can match baseline $\softmaxattn$ performance by relying on \emph{learnable} scalar bias and temperature: $\{b, t\} \in \mathbb{R}$.} All results presented for $\softmaxattn$ also fairly add LayerScale unless specified.
    \item LM and ASR are sensitive to the initial norm $|| \sigma(\mQ \mK^T/\sqrt{d_{qk}}) \mV ||$. Modulation is required via (a) relative positional embeddings like ALiBi \citep{DBLP:conf/iclr/PressSL22}, which reduces the initial attention norm by shifting logit mass near zero under $\sigmoidattn$, (b) appropriate initialization of $b$ to achieve the same effect -- enabling usage of any positional embedding, (c) using hybrid-norm (\Cref{app:norm_structure,sec:asr_hps,sec:practioners_guide}) at the expense of an extra normalization layer.
\end{enumerate}
\subsection{Ablations}
\label{sec:ablations}
\begin{figure}[h] %
    \centering
    \begin{minipage}{0.48\textwidth}
        \centering
        \includegraphics[width=\textwidth]{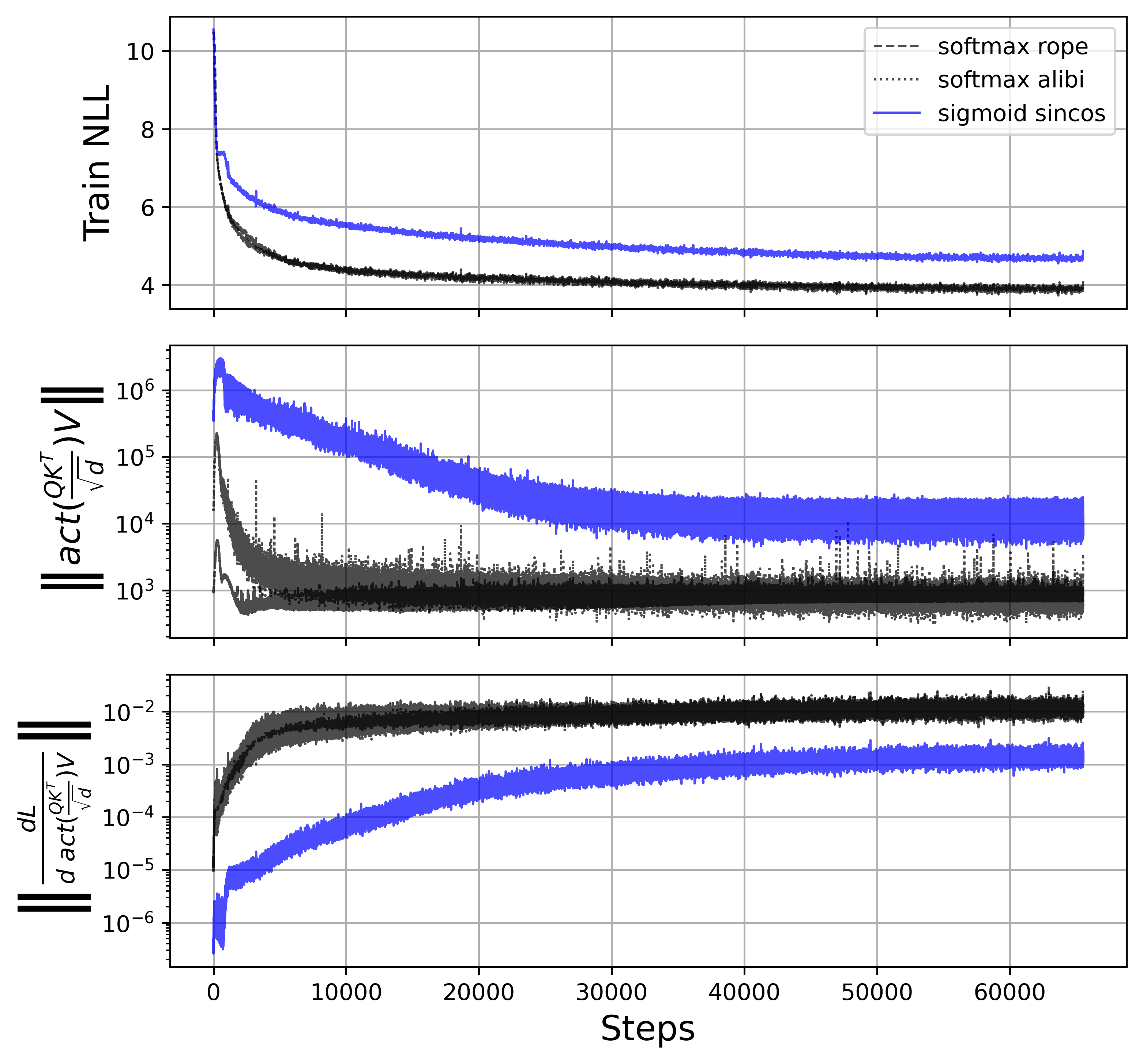}    
        \captionsetup{justification=centering}
        \caption{$\sigmoidattn$ with SinCos.}
        \label{fig:rope_vs_sincos}
    \end{minipage}\hfill
    \begin{minipage}{0.48\textwidth}
        \centering        
        \includegraphics[width=\textwidth]{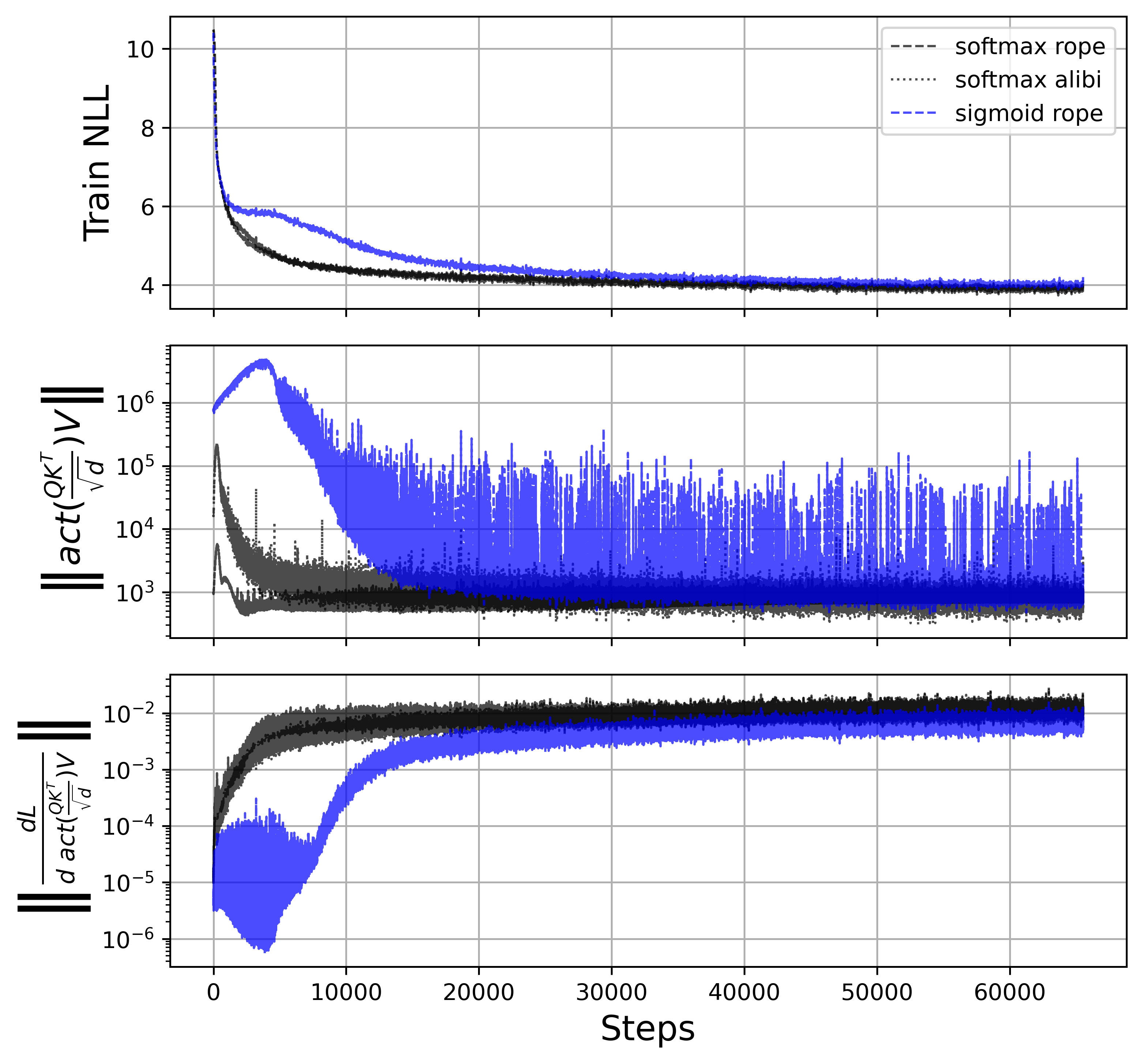}
        \captionsetup{justification=centering}
        \caption{$\sigmoidattn$ with RoPE.}
        \label{fig:rope_vs_rope}
    \end{minipage}
    \hfill
    \begin{minipage}{0.48\textwidth}
        \centering
        \includegraphics[width=\textwidth]{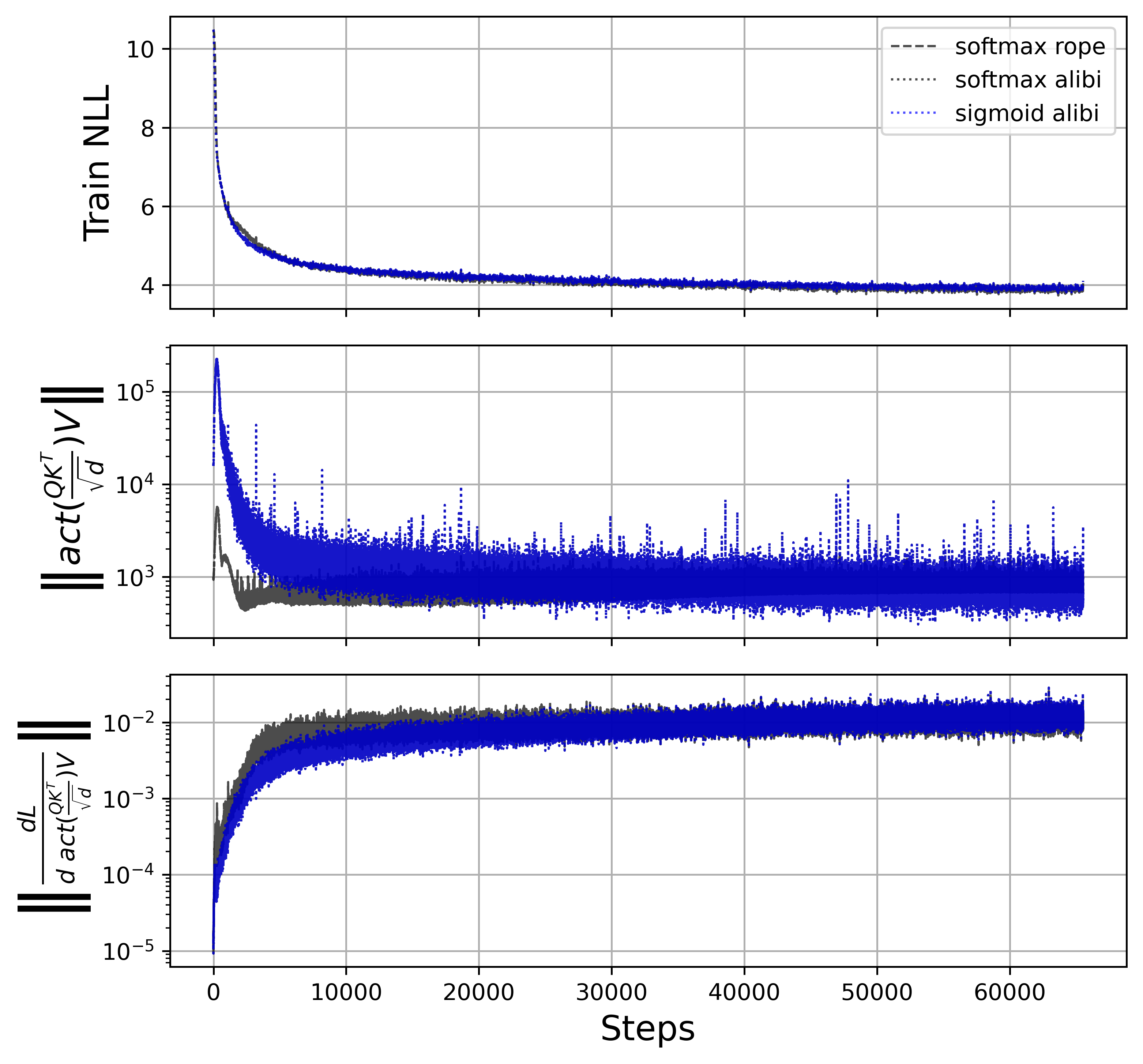}
        \captionsetup{justification=centering}        
        \caption{$\sigmoidattn$ with ALiBi.}
        \label{fig:rope_vs_alibi}
    \end{minipage}\hfill
    \begin{minipage}{0.48\textwidth}
        \centering        
        \includegraphics[width=\textwidth]{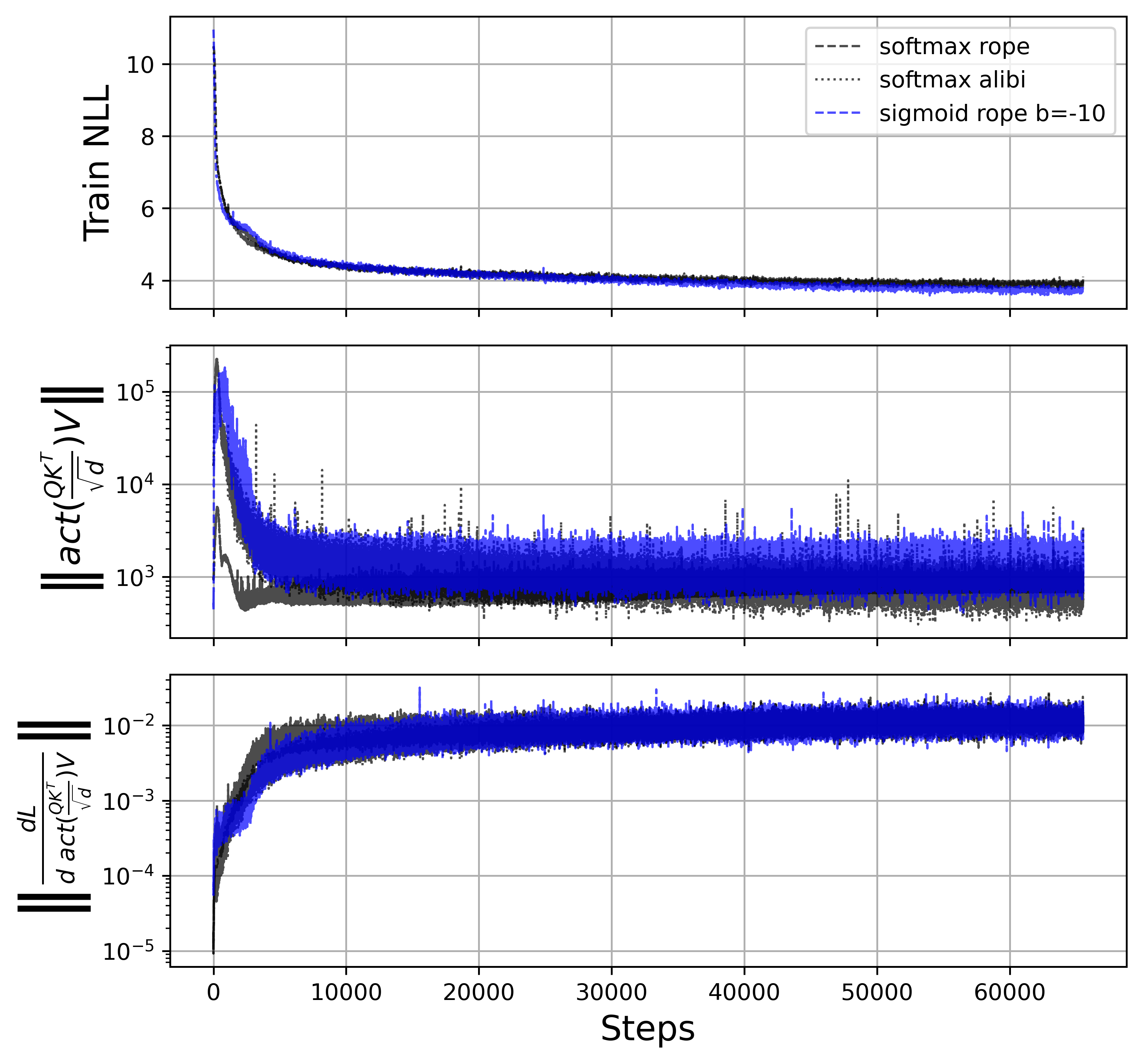}
        \captionsetup{justification=centering}
        \caption{$\sigmoidattn$ with RoPE, $b=-10$.}
        \label{fig:rope_vs_rope_b-10}
    \end{minipage}  
\end{figure}
We begin with ablations to dissect the benefits of each of our introduced components. To gain intuition about $\sigmoidattn$, we developed a research-friendly auto-regressive (AR) LM training framework to measure all components of attention and validate the effects of LayerScale, LayerNorm applied to Q and K (QK norm), different positional embedding techniques, and initialization values for $b$.
\paragraph{Mitigating Large Attention Norms} We train a single layer AR transformer block (E=3072, D\_FF=12288) on the realnews split of C4 \citep{DBLP:journals/jmlr/RaffelSRLNMZLL20}. We train for $2^{16}$ steps using a batch size of 6 and max sequence length of 4096 using a single cycle cosine learning rate (LR) schedule without weight decay. $\sigmoidattn$ initially underperformed $\softmaxattn$ when using absolute sinusoidal (SinCos) (\cref{fig:rope_vs_sincos}) or relative (\cref{fig:rope_vs_rope}) positional embeddings (PE), which we attribute to high initial attention Frobenius norms, $\lVert \sigma(\mQ \mK^T / \sqrt{d}) \mV \rVert$. A corresponding evolution of the attention distribution and sparsity can be seen in Appendix \cref{fig:attn_evolve} and \cref{fig:attn_metric_evolve} on a synthetic task.
To address these larger attention norms, we propose: (a) using ALiBi \citep{DBLP:conf/iclr/PressSL22} whose relative bias moves initial attention logit mass to the zero region under the sigmoid activation, producing equivalent train negative log-likelihoods (\cref{fig:rope_vs_alibi}); or (b) set the attention logit bias $b$ to a negative offset proportional to the sequence length, $b \propto -\ln n$ (see \cref{sec:attn_bias_ablation} for an ablation on $b$). This enables the usage of other PE techniques like RoPE~\citep{DBLP:journals/ijon/SuALPBL24} (\cref{fig:rope_vs_rope_b-10}). 
\begin{figure}[h] %
  \centering
  \includegraphics[width=\textwidth]{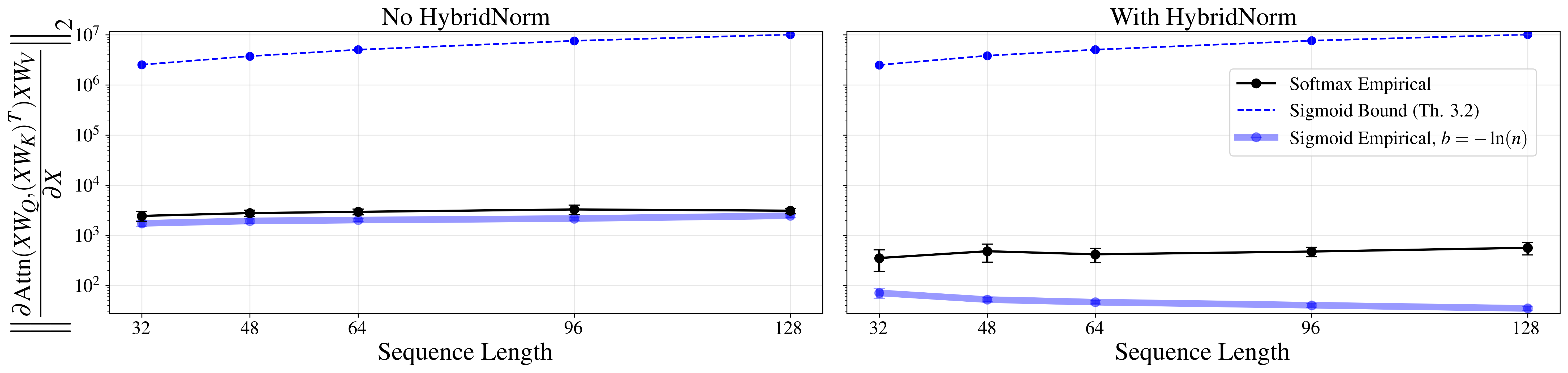}
  \captionsetup{justification=centering}
  \caption{Regularity analysis comparing $\sigmoidattn$ vs. $\softmaxattn$ (10$\times$ trials per $n$). $\softmaxattn$ theoretical bound is off scale and thus omitted.}
  \label{fig:empirical_regularity}
\end{figure}
\paragraph{Empirical Analysis of Attention Regularity} To validate our theoretical analysis (\Cref{sec:regularity}), we measure Jacobian norms of $\sigmoidattn$ and $\softmaxattn$ across sequence lengths (\Cref{fig:empirical_regularity}). Using autograd, we compute exact Jacobian norms for both mechanisms, with and without HybridNorm (\Cref{app:norm_structure,sec:practioners_guide}), comparing them to theoretical bounds ($\softmaxattn$ bound omitted as it exceeds scale). Both variants show empirical norms (solid lines) well below their theoretical bounds (dashed lines). With our proposed bias initialization ($b = -\ln(n)$), $\sigmoidattn$ achieves lower norms than $\softmaxattn$ in both settings, suggesting improved regularity. This aligns with its strong task performance (\Cref{sec:experiments}). Additionally, HybridNorm (\Cref{fig:empirical_regularity}, right) reduces norms for both mechanisms compared to baseline (left), highlighting normalization's role in attention stability at longer sequences.
\paragraph{LayerScale} To validate the need for LayerScale, we follow \citet{DBLP:journals/corr/abs-2309-14322} to quantify the impact on stability.
All models are trained with RoPE with $b \propto -\ln n$, using AdamW  \citep{loshchilov2017decoupled} on the 
realnews split of C4 
with $(\beta_1,\beta_2)=(0.9, 0.95)$, $\eps=10^{-8}$,  $wd=0$, 
batch size 24, maximum token sequence length of 512 from the T5 tokenizer \citep{DBLP:journals/jmlr/RaffelSRLNMZLL20}, cosine LR schedule of $2^{14}$ steps including a linear warmup of $2^{10}$ steps. 
Models have 
$n_{\text{heads}}=\kappa$,
$n_{\text{layers}}=2\times \kappa$,
$d_{\text{model}}=64\times \kappa$ and
$d_{\text{feed-forward}}=256\times\kappa$
for a scaling value $\kappa\in\{1,2,4,8,16\}$
leading to models with $\{2.2, 4.9,15.0,67.0,440.0\}M$ trainable non-embedding parameters.
Following \citet{DBLP:journals/corr/abs-2309-14322},
we sweep learning rates
$\eta\in \{3\times 10^{-4}, 1\times 10^{-3}, 3\times 10^{-3}, 1\times 10^{-2}, 3\times 10^{-2}, 1\times 10^{-1}, 3\times 10^{-1}\}$.
LR sensitivity is defined as 
$\mathbb E_{\eta\in[a,b]}\left[\min(\ell(\mathcal A(\eta)),\ell_0)-\ell^*\right]$
where $\ell(\mathcal A(\eta))$ is the loss achieved by the learning algorithm $\mathcal A$ with LR $\eta$,
$\ell_0$ is the loss at initialization, and
$\ell^*$ is the loss achieved by the best LR.
LayerScale is initialized at $10^{-4}$. 
Unlike vision tasks, where LayerScale \emph{improves performance} (\cref{fig:imagenet_top_1_ablations}-a), in LM, we observe that $\softmaxattn$ slightly benefits from LayerScale, while the performance of $\sigmoidattn$ remains largely unaffected.
\begin{figure}[h]
    \centering
    \begin{minipage}[t]{0.48\textwidth}
        \centering
        \includegraphics[width=\textwidth]{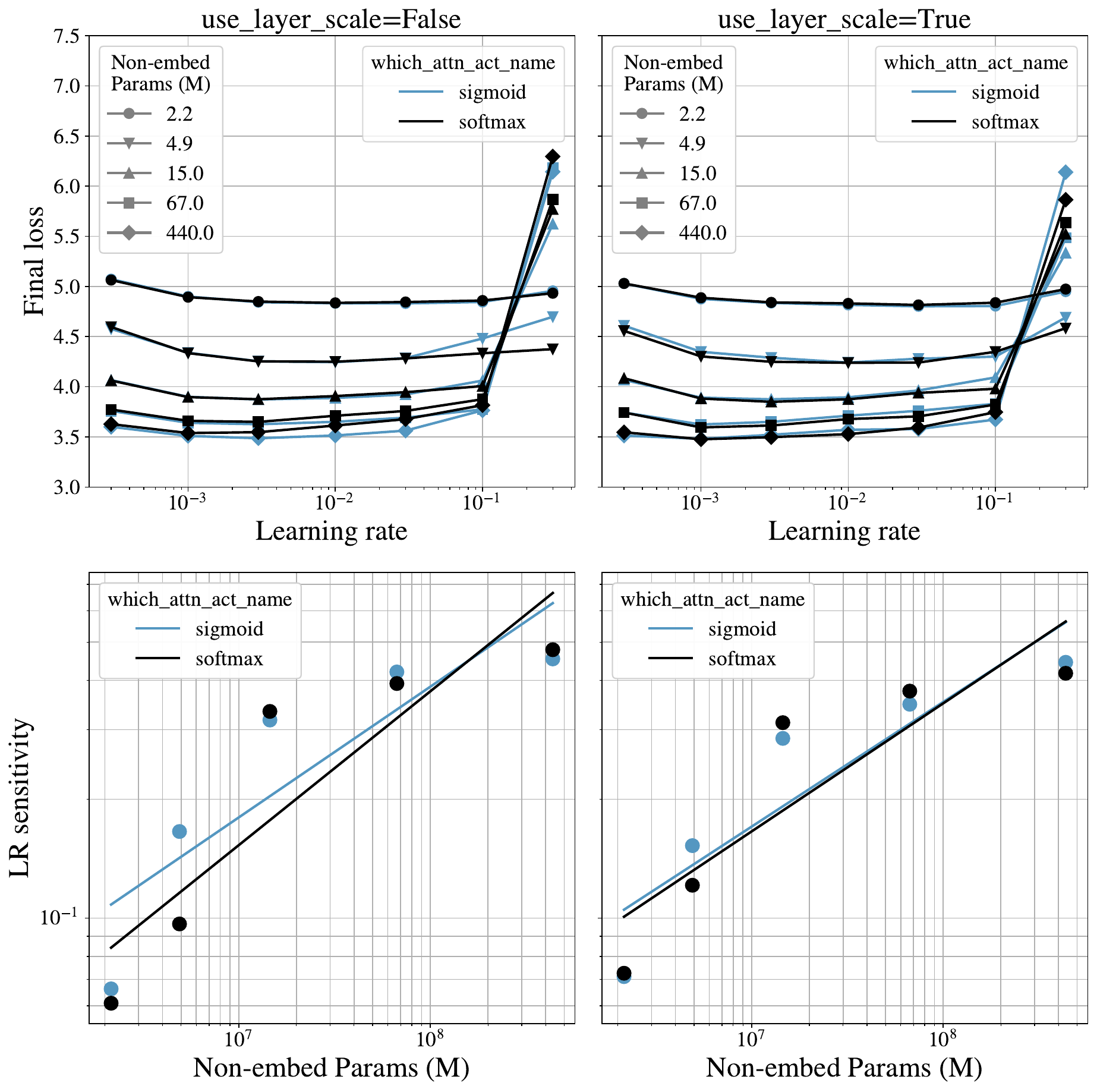} 
        \caption{LR sensitivity LayerScale ablation.}
        \label{fig:layerscale_ablation}
    \end{minipage}%
    \hfill
    \begin{minipage}[t]{0.48\textwidth}
        \centering
        \includegraphics[width=\textwidth]{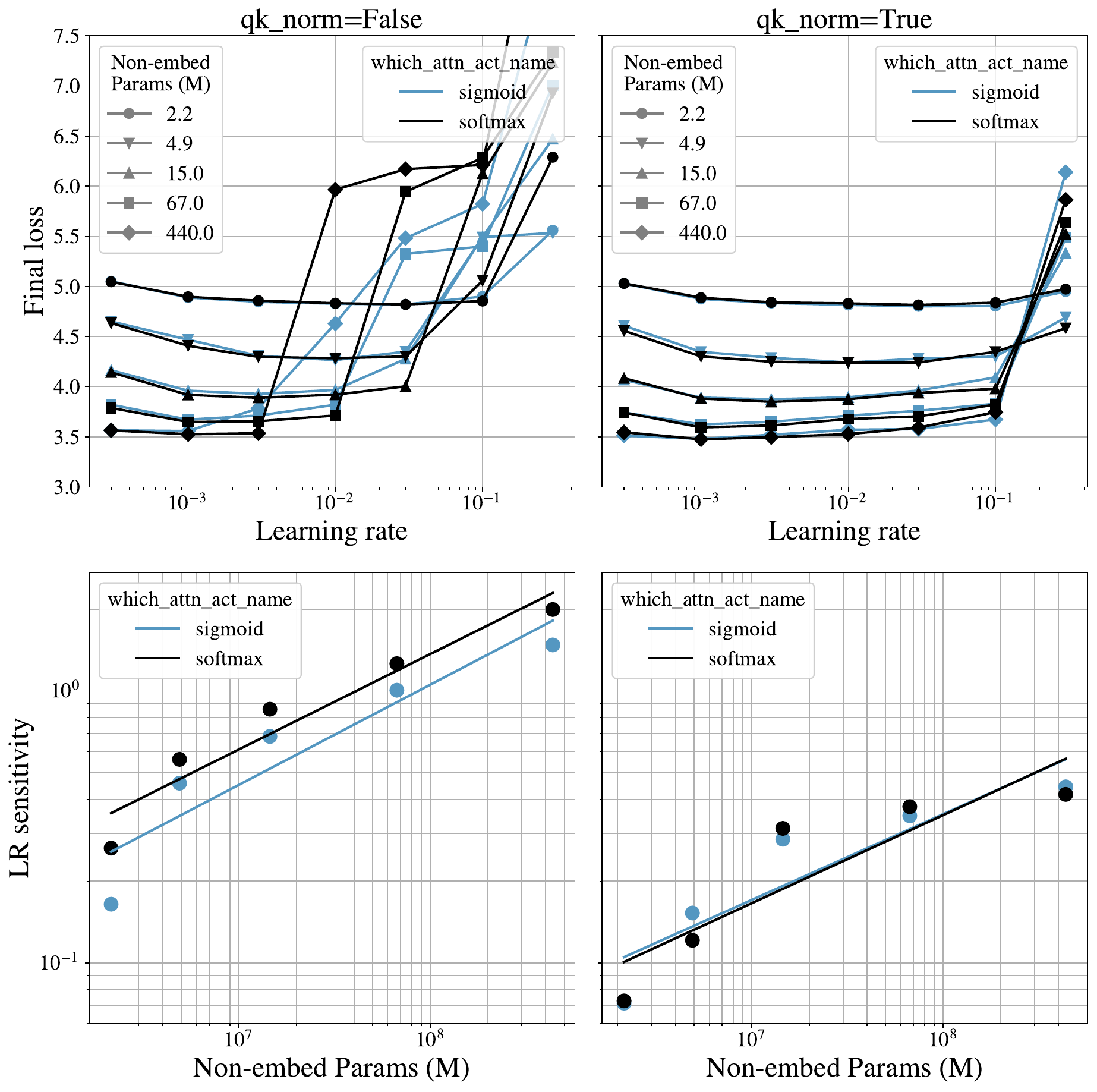}
        \caption{LR sensitivity QK norm ablation.}
        \label{fig:qk_norm_ablation}
    \end{minipage}
\end{figure}
\begin{figure}[h]
    \centering
    \includegraphics[width=\textwidth]{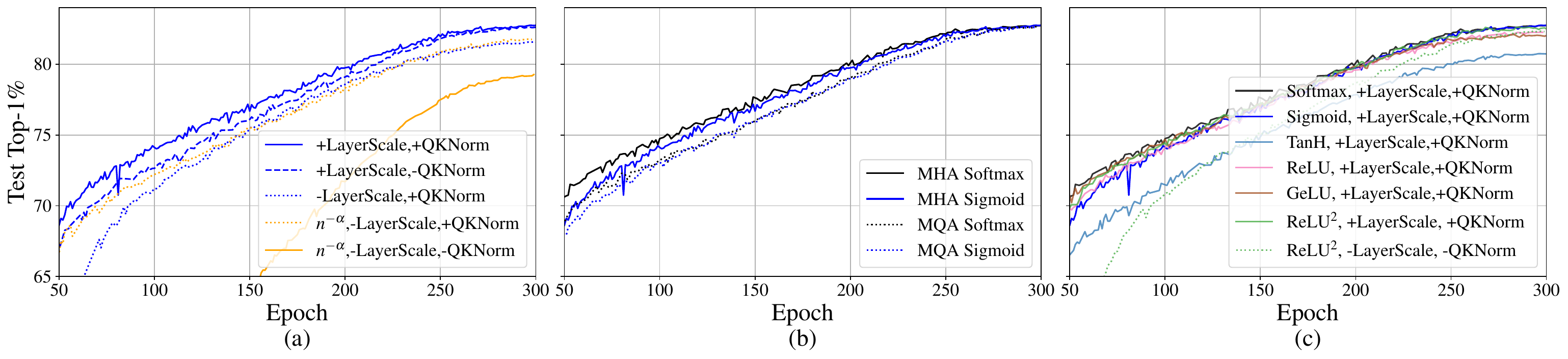}
    \caption{ImageNet1k ViT-B/16 classification. (a) $\sigmoidattn$ is robust without QK norm (+LayerScale, -QKNorm). Removing LayerScale reduces accuracy by 1.0\% (-LayerScale, +/-QKNorm). $n^{-\alpha}$ normalization \citep{wortsman2023replacing} underperforms without LayerScale. (b) $\sigmoidattn$ multi-query attention (MQA) \citep{DBLP:journals/corr/abs-1911-02150} with one head matches multi-head attention (MHA). (c) Sigmoid with LayerScale and QK norm performs comparably to other activations, except TanH. ReLU$^2$ \citep{DBLP:conf/icml/HuaDLL22} underperforms without LayerScale and QK norm.}
    \label{fig:imagenet_top_1_ablations}
\end{figure}

\paragraph{Stability with QK Norm}
To explore the stability of $\softmaxattn$ vs. $\sigmoidattn$ we repeat the analysis of \citet{DBLP:journals/corr/abs-2309-14322}, as described in the LayerScale analysis, to investigate the impact of QK norm \citep{DBLP:conf/icml/0001DMPHGSCGAJB23}. For language modeling, both $\sigmoidattn$ and $\softmaxattn$ exhibit sensitivity to learning rate changes without QK norm. However, incorporating QK norm significantly stabilizes performance (\cref{fig:qk_norm_ablation}). In vision tasks, $\sigmoidattn$ demonstrates robustness with and without QK norm (\cref{fig:imagenet_top_1_ablations}-a) and without the need for $n^{-\alpha}$ normalization from \citet{wortsman2023replacing}.\footnote{We ablate multiplicative sequence length scaling in more detail in \cref{sec:appendix_normalization}.}
\paragraph{Multi-query attention (MQA)} In \cref{fig:imagenet_top_1_ablations}-b we explore MQA \citep{DBLP:journals/corr/abs-1911-02150} for vision using only one head for $\{ \mK, \mV \}$. We find that both $\sigmoidattn$ and $\softmaxattn$ perform equally well with or without multiple heads even at the small scale of ViT-B/16.
\paragraph{Activation Function Ablations} As in \citet{wortsman2023replacing}, various activation functions, when combined with LayerScale and QK norm, perform equally well for vision tasks (\cref{fig:imagenet_top_1_ablations}-c). However, for sequence-critical tasks like ASR, activation functions such as ReLU pose instabilities and underperform. In the same figure, we also compare to the ReLU$^2$ proposal from \citet{DBLP:conf/icml/HuaDLL22} and find that it underperforms without LayerScale and QK norm.
\subsection{Supervised Image Classification}
\label{sec:supervised_image_classification}
Vision transformers \citep{DBLP:conf/iclr/DosovitskiyB0WZ21} extend transformers  \citep{DBLP:conf/nips/VaswaniSPUJGKP17} to treat $K \times K$ image grids as disparate tokens. All tokens are refined through sequential layers of self-attention, pooled using a CLS token or global average pooling layer, and optimized using the negative log likelihood, $\ln p(\vy|\vx)$. We train ViT-B/16 models using $\mathbb{R}^{224 \times 224 \times 3}$ images for 300 epochs using the recipe provided in \cref{sec:appendix_vision_hyperparams}. We use the same set of training hyper-parameters for both $\softmaxattn$ and $\sigmoidattn$, changing only the activation function between trials. The train negative log-likelihood is reported in \cref{fig:summary_nll} and the test top-1\% is reported in \cref{fig:test_top1_results}. We find that $\sigmoidattn$ matches both the training dynamics and the evaluation performance of $\softmaxattn$.
\subsection{Self-Supervised Image Representation Learning}
\label{sec:ssl}
Self-supervised representation learning (SSL) exploits vast quantities of unlabeled data to learn semantic representations based on inductive biases such as augmentation invariance (SimCLR \cite{DBLP:conf/icml/ChenK0H20}, BYOL \citep{DBLP:conf/nips/GrillSATRBDPGAP20}) or reconstruction from compressed representations (MAE \citep{DBLP:conf/cvpr/HeCXLDG22}). We employ vision transformer training recipes from \cite{DBLP:conf/icml/ZhaiLLBR0GS23} and \cite{DBLP:conf/nips/BusbridgeRALDCW23} (\cref{sec:appendix_vision_hyperparams}) for SimCLR and BYOL. As with supervised learning, we use the same set of training hyper-parameters for both $\softmaxattn$ and $\sigmoidattn$, changing only the activation function between trials. \Cref{fig:summary_nll} reports the train losses, and \cref{fig:test_top1_results} highlights the linear probe and finetuned test top-1\%. Despite the diverse training objectives in SSL, $\sigmoidattn$ matches $\softmaxattn$ while improving training and inference throughput (\cref{sec:FlashSigmoidHardwareAwareImplementation}).
\subsection{Automatic Speech Recognition (ASR)}
\label{sec:asr}
\begin{table}[t!]
\centering
\caption{Word error rate (\%) on LibriSpeech test sets and TED-LIUM v3~\citep{hernandez2018ted} (``TED'', joint validation and test sets split according to  duration) for transformer (255M params) with either $\softmaxattn$ or $\sigmoidattn$ (LayerScale and QK norm are used with $b=-\log n$) trained on LibriSpeech 960h data (mean duration is 10-15s). Hyper-parameters are in~\cref{sec:asr_hps}.}
\label{tab:asr-results}
\begin{center}
\begin{scriptsize}
\begin{sc}
\resizebox{\columnwidth}{!}{%
\begin{tabular}{lcrrrrrr}
\toprule
 attn & PE & test-clean & test-other & ted 0-10s & ted 10-20s & ted 20-30s & ted 30s+  \\
\midrule 
softmax & \multirow{7}{*}{CAPE} & 2.3 & 5.7 & 12.4 & 10.5 & 11.9 & 9.1 \\
 sigmoid &  & 2.4 & 5.5 & 12.4 & 10.3 & 12.3 & 9.7 \\
 \,\,\,\, - QK norm &  & \multicolumn{6}{c}{unstable, gradient norm and loss spikes} \\
 \,\,\,\, - LayerScale &  & 2.5 & 6.1 & 13.6 & 11.5 & 13.4 & 8.9 \\
 sigmoid ($b=-10$, learnable) &  & 2.3 & 5.5 & 12.1 & 10.5 & 13.0 & 9.3 \\
 sigmoid ($b=-5$ in $Q$, learnable) &  & 2.3 & 5.4 & 12.2 & 10.8 & 12.4 & 9.9 \\
 \,\,\,\, - QK norm &  & \multicolumn{6}{c}{unstable, gradient norm and loss spikes} \\

\midrule
softmax & \multirow{5}{*}{RoPE} & 2.2 & 5.5 & 12.7 & 10.6 & 12.8 & 9.5 \\
 sigmoid &  & 2.3 & 5.4 & 12.3 & 10.1 & 12.3 & 8.6 \\
 sigmoid ($b=-10$, learnable) &  & 2.2 & 5.2 & 12.4 & 10.5 & 12.3 & 21.8 \\
 \,\,\,\, + $\alpha=1$ &  & 2.7 & 6.6 & 14.1 & 12.0 & 14.5 & 14.9 \\
 sigmoid ($b=-5$ in $Q$, learnable) &  & \multicolumn{6}{c}{unstable, gradient norm and loss spikes} \\
\midrule
 softmax & \multirow{5}{*}{ALiBi} & 2.2 & 5.4 & 12.3 & 10.7 & 12.1 & 8.6 \\
 sigmoid &  & 2.3 & 5.1 & 12.3 & 10.5 & 12.6 & 9.1 \\
 sigmoid ($b=-10$, learnable) &  & 2.2 & 5.2 & 12.4 & 10.4 & 11.7 & 9.1 \\
 \,\, + $\alpha=1$ &  & 2.6 & 6.6 & 13.9 & 11.9 & 14.2 & 8.6 \\
 sigmoid ($b=-5$ in $Q$, learnable) &  & 2.2 & 5.2 & 12.1 & 10.4 & 12.0 & 8.2 \\
\bottomrule
\end{tabular}
}
\end{sc}
\end{scriptsize}
\end{center}
\end{table}
We benchmark ASR using LibriSpeech data \citep{DBLP:conf/icassp/PanayotovCPK15} on 100h and 960h settings of paired speech and text transcriptions. Our PyTorch implementations of encoder-based vanilla transformer~\citep{synnaeve2019end} and conformer \citep{DBLP:conf/interspeech/GulatiQCPZYHWZW20} are trained with Connectionist Temporal Classification (CTC) \citep{DBLP:conf/icml/GravesFGS06} w/ BF16 mixed precision, w/o QK norm and w/o LayerScale. After extensively tuning $\softmaxattn$ baselines, we switch to $\sigmoidattn$ per \cref{eq:sigmoid_attn} without any other changes. We investigate the effects of post/pre-LayerNorm, model depth, optimizer type, small data regime, and connection to local attention, with details in~\cref{sec:asr_hps}.

Our main findings are: i) CAPE~\citep{DBLP:conf/nips/LikhomanenkoXSC21} PE is the most unstable for $\sigmoidattn$; ii) post-LayerNorm models with $\softmaxattn$ are hard to match with stable $\sigmoidattn$; iii) w/o QK norm $\sigmoidattn$ is unstable and significant spikes happen in both gradient norms and training loss; iv) LayerScale is needed for generalization; v) learnable bias $b=-10$ gives no loss and gradient norms spikes while matching the $\softmaxattn$ (which does not benefit from the improved throughput of \textsc{FlashSigmoid}); vi) adding a learnable bias, $b=-5$, to $Q$ instead of the attention logits also solves the initial large attention norms for CAPE and ALiBi but not for RoPE; vii) $b=-\log n$ gives rare (2-5 times) marginal gradient norms spikes with smooth loss while matching $\softmaxattn$.

\Cref{tab:asr-results} shows the main result for pre-LayerNorm  transformers with CAPE, RoPE, and ALiBi, where $\sigmoidattn$ uses LayerScale, QK norm, $b=-\log n$, and no sequence normalization. The bias is ablated with learnable bias (one per layer) in attention or $Q$ with or without sequence normalization. $\sigmoidattn$ is stabilized with bias while matching $\softmaxattn$, and $b=-\log n$ works well. In most cases, bias allows generalization to longer sequences without sequence normalization, except for RoPE where it helps for longer sequences but hurts overall performance.

\subsection{Autoregressive Large Language Modeling}
\label{sec:llm}
\newcolumntype{R}[2]{%
    >{\adjustbox{angle=#1,lap=\width-(#2)}\bgroup}%
    l%
    <{\egroup}%
}
\newcommand*\rotdiag{\multicolumn{1}{R{30}{1em}}}%
\begin{table}[ht]
\centering
\caption{LLM English evaluation. All models use ALiBi. Detailed ablations in \Cref{app:norm_structure}.}
\label{tab:lm_results}
\begin{sc}
\begin{scriptsize}
\bgroup
\setlength{\tabcolsep}{.35em}
\begin{tabular}{@{}lllllllllllllll@{}}
\toprule
Model & \makecell{Size}  & \makecell{Seq.\\Len.} & \makecell{ARC\\Easy} & \makecell{ARC\\Chal.} & \makecell{Hella-\\swag} & Piqa & Sciq & \makecell{Wino-\\grande} & \makecell{Lambada\\OpenAI} & \makecell{TriviaQA\\(1-shot)} & \makecell{WebQS\\(1-shot)} & AVG & \makecell{Step\\time (s)} \\ \midrule
Softmax & 1B & 2k & 62.2       &     26.8           &    42.4       &  59.0    &   72.3   &     88.1       &     58.4           &      19.9             &    15.4            &    49.4   & 0.38   \\
Sigmoid & 1B & 2k &  62.8       &      28.8         &    42.5       &  59.7    &   70.3   &     88.6       &      59.7          &       19.1            &   13.8             &       49.5  & 0.34   \\
\midrule
Softmax & 1B & 4k & 62.6       &     27.7           &    42.4       &  58.6    &   71.1   &     88.2       &     58.6           &      18.9             &    14.7            &    49.2   & 0.84   \\
Sigmoid & 1B & 4k &  60.5       &      27.3         &    41.3       &  57.8    &   70.5   &     87.0       &      57.6          &       18.9            &   12.6             &       48.2  & 0.67   \\ 
\midrule
Soft (H-Norm) & 1B & 4k & 61.7       &      26.8         &    43.4       &  59.4    &   70.6   &     88.6       &      60.8          &       20.5            &   12.9             &       49.4  & -  \\
Sigm. (H-Norm) & 1B & 4k  & 63.5       &      28.1         &    43.5       &  60.7    &   70.8   &     88.9       &      59.0          &       20.9            &   16.0             &       50.2  & -  \\ 
\midrule
Soft (H-Norm) & 7B & 4k & 71.2       &      39.9         &    53.2       &  65.5    &   75.6   &     91.8       &      67.2          &       37.7            &   21.8             &       59.0  & 3.85  \\
Sigm. (H-Norm) & 7B & 4k  & 72.7       &      40.5         &    53.5       &  66.2    &   76.0   &     92.5       &      66.5          &       39.5            &   21.8             &       59.6  & 3.4  \\ \bottomrule
\end{tabular}
\egroup
\end{scriptsize}
\end{sc}
\end{table}

We train all models using the Llama2 recipe \citep{touvron2023llama} (with ALiBi instead of RoPE) and the RedPajama \citep{together2023redpajama} dataset in JAX without \textsc{FlashAttention} using the AXLearn framework\footnote{https://github.com/apple/axlearn} (\cref{sec:llm_appendix} for detailed hyper-parameters). Initial experiments at 85M parameters established basic stability requirements (\cref{fig:85m_4k_nll}), with attention bias $b = -\log(n)$ (n = 4096) providing effective results. At 1B n = 2048 scale with $b = -\log(n)$, $\sigmoidattn$ matches the train NLL (\cref{fig:summary_nll}) and evaluation results of $\softmaxattn$ (\cref{tab:lm_results} top row) while improving throughput by \textbf{1.12}$\mathbf{\times}$.

At the 1B n = 4096 scale, using just $b = -\log(n)$ we observe a \textbf{1.25}$\mathbf{\times}$ speedup; however, slight instabilities prevent $\sigmoidattn$ from matching the strong performance of $\softmaxattn$ (\cref{tab:lm_results} second row). We address these issues through hybrid-norm, an extra normalization layer applied on the output of the attention operation ($x + \text{norm}(\sigma(\mQ\mK^T / \sqrt{d_{qk}})\mV$, more details in \cref{sec:llm_appendix,sec:practioners_guide}). With hybrid-norm, $\sigmoidattn$ matches the train NLL and slightly outperforms $\softmaxattn$ on English evaluation results (50.2\% vs. 49.4\% -- \cref{tab:lm_results} third row). In \cref{app:norm_structure} we ablate various design choices at 1B scale, including norm structures, position embedding techniques, and attention bias configurations.

At 7B n = 4096, $\sigmoidattn$ with hybrid-norm demonstrates compelling advantages compared to $\softmaxattn$ with hybrid-norm: it matches the train NLL of $\softmaxattn$ (\cref{fig:summary_nll}), while delivering a \textbf{1.13}$\mathbf{\times}$ speedup (\cref{tab:lm_results} bottom row). The model shows marginal improvements on challenging tasks, including both reasoning (ARC-Challenge: 40.5\% vs 39.9\%) and knowledge retrieval (TriviaQA: 39.5\% vs 37.7\%), with better average performance across all benchmarks (59.6\% vs 59.0\%). These results establish SigmoidAttn with hybrid-norm as an efficient alternative for large-scale language modeling.

\section{Related Work}
\label{sec:related}
Recent studies in supervised image classification \citep{DBLP:journals/corr/abs-2110-00476} and self-supervised learning (SSL), including approaches like SigLIP \citep{DBLP:journals/corr/abs-2303-15343}, 
are shifting large-scale machine learning training from output conditional categorical distributions, traditionally parameterized by softmax functions, to richer pointwise Bernoulli conditionals parameterized 
by sigmoid functions. In this study, our focus shifts to refining the model's internal mechanics, specifically by substituting the softmax component of the attention mechanism with a pointwise sigmoid function.

Previous work has explored the replacing softmax with the ReLU activation in both practical \citep{DBLP:journals/corr/abs-2302-06461,DBLP:conf/icml/HronBSN20} and theoretical settings \citep{DBLP:conf/nips/BaiCWXM23,DBLP:conf/nips/Fu00M23}. Other works explores using the ReLU$^2$ activation \citep{DBLP:conf/icml/HuaDLL22}, exploring purely linear attention \citep{DBLP:conf/icml/KatharopoulosV020,DBLP:conf/nips/LuYZZXGXXZ21,DBLP:conf/wacv/KoohpayeganiP24} or cosine-similarity based attention \citep{DBLP:conf/icann/LuoZXWRY18,DBLP:conf/cvpr/Liu0LYXWN000WG22}. 
Our work builds upon these explorations, particularly \cite{wortsman2023replacing}, which replaces softmax with various activation functions scaled by $n^{-\alpha}$, where $n$ corresponds to the sequence length and $\alpha$, a hyper-parameter. However, we find that their formulation does not match expected performance without proper $b$ initialization and the use of LayerScale (\cref{fig:imagenet_top_1_ablations}-a, \cref{sec:appendix_normalization}). 

\section{Conclusion}
\label{sec:conclusion}
In this work, we present a comprehensive theoretical and empirical study of sigmoid attention as an alternative to softmax attention in transformers. We prove that transformers with sigmoid attention are universal function approximators with improved regularity, and identify LayerScale and prevention of large initial attention norms as key factors for successful training. We introduce \textsc{FlashSigmoid}, a memory-efficient variant providing a 17\% inference kernel speed-up. Extensive experiments across language, vision, and speech demonstrate that properly normalized sigmoid attention matches softmax attention performance on various tasks and scales. Our findings establish sigmoid attention as a viable alternative, unifying prior work and establishing best practices for its application in transformers.

\section{Acknowledgements}
\label{sec:acknowledgements}

We thank
Zakaria Aldeneh,
Samy Bengio,
Navdeep Jaitly,
David Koski,
Pau Rodriguez Lopez,
Hadi Pouransari, and
Skyler Seto
for their helpful feedback and critical discussions throughout the process of writing this paper;
Okan Akalin,
Hassan Babaie, 
Michael Brooks,
Brian Gamp,
Denise Hui,
Mubarak Seyed Ibrahim, 
Li Li, 
Rajat Phull,
Evan Samanas, 
Guillaume Seguin, 
and the wider Apple infrastructure team for assistance with developing and running scalable, fault tolerant code. 
Names are in alphabetical order by last name within group.

\bibliography{example_paper}
\bibliographystyle{iclr2025_conference}

\newpage
\appendix
\startcontents[sections]
\printcontents[sections]{l}{1}
\appendixpage
\clearpage
\section{Limitations}
\label{sec:limitations}

While our work demonstrates that $\sigmoidattn$ can serve as a viable drop-in replacement for $\softmaxattn$ in many domains and scales, there are a few key limitations to note:

\begin{enumerate}
    \item In large-scale (1B parameter, 4096 context length) language modeling, we observed some gradient norm spikes and a slight performance gap between $\sigmoidattn$ and $\softmaxattn$ (\Cref{tab:lm_results}). While runs at smaller context lengths (1B parameter, n=2048) were stable and matched $\softmaxattn$ performance, we required the use of hybrid-norm to stabilize n=4096 sequence length (and the larger 7B models). Hybrid-norm does incur a slight extra performance penalty which we quantify in \Cref{sec:practioners_guide}.
    \item Our theoretical analysis proves that transformers with $\sigmoidattn$ are universal function approximators and have improved regularity compared to $\softmaxattn$. However, the bounds we derive, while tighter than those for $\softmaxattn$, may not be maximally tight. There could be room for further theoretical refinements.
    \item We focused our empirical evaluation on standard benchmarks in language, vision, and speech domains. Performance on more niche or emerging applications remains to be validated. 
    \item In automatic speech recognition experiments, we observed that $\sigmoidattn$ can be sensitive to the choice of positional embeddings and may require careful initialization of the attention bias term to ensure stable training. Specifically, we found that the CAPE positional embedding was the most unstable for $\sigmoidattn$. Further work is needed to develop robust initialization schemes that work well across different positional embeddings. Moreover we found that w/o QK norm or with post-LayerNorm $\sigmoidattn$ is unstable and can underperforms $\softmaxattn$, thus further investigation is needed.
    \item \textsc{FlashSigmoid} demonstrates promising inference and training speed-ups by exploiting $\sigmoidattn$'s simpler kernel structure compared to $\softmaxattn$. However, realizing these gains at scale in distributed training setups may require additional engineering to optimize communication bottlenecks.
\end{enumerate}

Despite these limitations, we believe this work establishes a strong foundation for $\sigmoidattn$, unifying prior art and demonstrating its potential as a drop-in $\softmaxattn$ replacement. We hope our theoretical grounding and empirical results motivate further research into this simple yet effective architectural variation.

\section{Broader Impact}
\label{sec:broader_impact}
The development of efficient and theoretically grounded attention mechanisms has the potential for significant positive impact across a range of applications. By establishing $\sigmoidattn$ as a viable alternative to $\softmaxattn$, our work expands the toolkit of architectural choices available to researchers and practitioners.
Positive impacts of this work may include:
\begin{enumerate}
    \item Improved computational efficiency: \textsc{FlashSigmoid}'s faster kernel implementation could lead to more efficient training and inference for attention-based models, reducing energy consumption and enabling deployment on resource-constrained devices. This could democratize access to powerful models.
    \item Theoretical understanding: Our universal approximation results and tighter bounds on the regularity of $\sigmoidattn$ contribute to a deeper theoretical understanding of this key component. A stronger theoretical foundation can guide principled model design and architectural search.
    \item Application-specific benefits: Across language, vision, and speech domains, $\sigmoidattn$'s performance could translate into improved user experiences, such as more natural language interactions, enhanced image understanding, and robust speech recognition. These advancements could have positive societal impacts, such as improved accessibility tools and more effective educational technologies.
\end{enumerate}
However, as with any foundational machine learning advance, there are also risks of negative impacts that must be considered and mitigated:
\begin{enumerate}
    \item Fairness and bias considerations: As with any machine learning model, it is important to carefully evaluate $\sigmoidattn$ based models for fairness and potential biases when applied to sensitive use cases. The unique properties of $\sigmoidattn$ may have unexpected interactions with data biases. Researchers and practitioners should follow best practices for auditing and mitigating unwanted biases to ensure equitable outcomes.
    \item Environmental impact: While \textsc{FlashSigmoid} is more computationally efficient than \textsc{FlashAttention}, the overall trend of scaling up attention-based models has significant energy costs. Further efficiency improvements and the use of renewable energy sources are important to mitigate environmental harms.
\end{enumerate}
We believe that the benefits of $\sigmoidattn$ outweigh the risks, but it is crucial for the research community to actively consider and address these potential negative impacts. By doing so, we can work towards a future where the efficiency and expressivity of $\sigmoidattn$ are used for societal benefit.

\section{Universal Approximation Property for Sigmoid Attention}
\label{app:UAP_proof}
This section is dedicated to the proof for the Universal Approximation Property for attention equipped with sigmoid nonlinearity. The proof follows closely the one provided in \citet[Sec.~3]{Yun_UAP}, of which we inherit much of the notation, and we encourage the interested reader to refer to the original source for a more comprehensive understanding of its details. Here we first provide context by outlining the main steps in the original proof, before proceeding to adapt its key components to the $\sigmoidattn$ case.

The proof aims at showing that a transformer network can approximate to arbitrary accuracy any continuous, permutation-equivariant function with compact support. The proof is constructive in nature, in that it explicitly defines the architecture (and particularly, the sequence of self-attention and feed-forward layers) that can approximate a given target function. To do so, it proceeds in steps (see \citet[Sec.~3.2]{Yun_UAP}):
\begin{enumerate}
    \item\label{step::piecewise_approx} prove that any continuous function with compact support can be approximated to arbitrary accuracy by a piecewise constant function
    \item prove that an aptly-constructed \emph{modified} transformer network, (where the softmax nonlinearity is substituted with a hardmax nonlinearity), can exactly represent such piecewise constant function. This step is further divided into three sub-steps (see \citet[Sec.~4]{Yun_UAP}):
        \begin{enumerate}
            \item\label{step::quantisation} prove that a series of feed-forward layers can quantize any input to a specific discretization grid in the compact domain
            \item\label{step::context_map} prove that a series of self-attention layers can implement a \emph{contextual mapping} (see \citet[Def.~3.1]{Yun_UAP})
            \item\label{step::output_map} prove that a series of feed-forward layers can map the output of the contextual mapping to the desired output of the target piecewise-constant approximation
        \end{enumerate} 
    \item\label{step::modified_trans} prove that a (classical) transformer network can approximate such modified transformer network to arbitrary accuracy
\end{enumerate}
Fortunately, some of the steps outlined above do not rely on a specific nonlinear function being used within the attention mechanism, and can be directly reused in our proof, virtually unchanged. Notice however that \cref{step::context_map,step::modified_trans} are directly impacted by modifications to the attention layer, and hence require adaptation in our case. This is the focus of the next sections.

\subsection{Proof of \cref{step::modified_trans}: Sigmoid Transformers can Approximate Modified Sigmoid Transformers}
\label{sec::proof_modified_sigmoid}
In \cite{Yun_UAP}, to implement contextual mappings, the authors rely on a \emph{modified} version of transformers, for the sake of simplifying the analysis. In their modified version, the (row-wise) softmax operation is substituted with a (row-wise) hardmax operation. This substitution is valid because a classical transformer can still be made arbitrarily close to such modified transformer, in light of the fact that
\begin{equation}
    \text{softmax}(\lambda\bb{X})\xrightarrow[]{\lambda\to\infty}\text{hardmax}(\bb{X}).
    \label{eqn::classic2modified_softmax}
\end{equation}
In our proof, we follow a similar strategy to define our modified sigmoid transformer (and in particular, its self-attention mechanism). We have that
\begin{equation}
    \sigma(\lambda\bb{X})\xrightarrow{\lambda\to\infty}H(\bb{X}),
    \label{eqn::classic2modified_sigmoid}
\end{equation}
where $\sigma(x)=(1+e^{-x})^{-1}$ is the (elementwise) sigmoid function, while
\begin{equation}
    H(x) = \begin{cases}
        1 & x>0\\
        \frac{1}{2} & x=0\\
        0 & x<0
    \end{cases}
\end{equation}
denotes the (elementwise) Heaviside step function. This allows us to define our modified sigmoid self-attention layer, as follows.
\begin{definition}[Modified sigmoid self-attention layer]
\label{def::modified_sigmoid_attention}
    Given an input $\bb{X}\in\mathbb{R}^{d\times n}$, the action of a modified sigmoid self-attention layer with shifts and a single one-dimensional head is defined as $\bb{X}\mapsto \bb{X} + \psi(\bb{X};\bb{q},\bb{b}_q,\bb{k},\bb{b}_k,\bb{v},\bb{o})$, where
    \begin{equation}
        \psi(\bb{X};\bb{q},\bb{b}_q,\bb{k},\bb{b}_k,\bb{v},\bb{o}) = \bb{o}\left(\bb{v}^T \bb{X}\right) H\left(\left(\bb{q}^T \bb{X}-\bb{b}_q^T\right)^T\left(\bb{k}^T \bb{X}-\bb{b}_k^T\right)\right)
        \label{eqn::modified_sigmoid_attention}
    \end{equation}
    with $\bb{q},\bb{k},\bb{v}\in\mathbb{R}^{d}$ representing the query, key, and value vectors, $\bb{b}_q,\bb{b}_k\in\mathbb{R}^{n}$ the corresponding query and key bias vectors, while $\bb{o}\in\mathbb{R}^{d}$ denotes the output vector.
\end{definition}
Analogously to \cref{eqn::classic2modified_softmax}, \cref{eqn::classic2modified_sigmoid} guarantees that sigmoid attention can approximate modified sigmoid attention by simply increasing the magnitude of its inner parameters.

Here and in the following, the length of the input sequence is denoted as $n$, while $d$ represents the dimensionality of the tokens. Notice that we are considering the input tensor $\bb{X}\in\mathbb{R}^{d\times n}$, (as opposed to $\in\mathbb{R}^{n\times d}$) to better align out notation with the one used in \cite{Yun_UAP}. 

\subsection{Proof of \cref{step::context_map}: Modified Sigmoid Transformers can Implement Contextual Mappings}
\label{sec::proof_contextual_mapping_top}
The core of the proof consists in showing how, by opportunely combining the operations in \cref{eqn::modified_sigmoid_attention}, one can build an architecture capable of implementing a \emph{contextual mapping}. For completeness, we report next the definition of such a map (see also \citet[Def.~3.1]{Yun_UAP}).
\begin{definition}[Contextual mapping]
    A map $\bb{q}:\mathbb{L}\to\mathbb{R}^{n}$ from a finite set $\mathbb{L}\subset\mathbb{R}^{d\times n}$ is said to be a \emph{contextual mapping} if both the following conditions hold:
    \begin{enumerate}[label=(\roman*)]
        \item $q_i(\bb{X})\neq q_j(\bb{X})$, $\forall i\neq j$ and $\forall\bb{X}\in\mathbb{L}$
        \item $q_i(\bb{X})\neq q_j(\bb{X}')$, $\forall i,j$ and $\forall\bb{X},\bb{X}'\in\mathbb{L}$, with $\bb{X}\neq\bb{X}'$
    \end{enumerate}
    where $q_i(\bb{X})$ denotes the $i$-th component of $\bb{q}(\bb{X})$.
    \label{def::contextual_mapping}
\end{definition}
Namely, a contextual mapping is such that it transforms each token in an input sequence to a value depending \emph{uniquely} on the \emph{whole} sequence. By satisfying this property, we can ensure that any element of the quantization of the input domain (achieved by \cref{step::quantisation}) can be mapped to a unique identifying value (depending on the whole input) via a sequence of modified sigmoid self-attention layers. It is then up to the MLP (in \cref{step::output_map}) to correctly map this value to the corresponding output value in the piece-wise constant approximation.

In particular, after defining a uniform discretization (characterized by the parameter $\delta$) of the unitary hypercube $[0,1]^{d}\subset\mathbb{R}^{d}$, namely
\begin{equation}
    \mathbb{G}_\delta \coloneqq \{\bb{g}:g_i\in\{0,\delta,2\delta,\dots,1-\delta\}, \quad\forall i=1\dots d\},
\end{equation}
we consider as input a tensor $\bb{X}$ (composed of columns $\bb{X}=[\bb{x}_{i}]_{i=1}^{n}$) such that
\begin{equation}
    \bb{X}\in\mathbb{L}\coloneqq\{\bb{X}: \bb{x}_{i}\in\mathbb{G}_\delta\;\forall i=1\dots n,\quad\text{and}\quad \bb{x}_{i}\neq\bb{x}_{j}\;\forall i\neq j\}\subset\mathbb{R}^{d\times n},
    \label{eqn::quantised_input}
\end{equation}
that is, a 2D tensor whose columns are element of the discretization $\mathbb{G}_\delta$, and that all differ from each other (at least for one element).
We want to build a contextual mapping acting on $\mathbb{L}$, by stacking layers parameterized according to \cref{def::modified_sigmoid_attention}. In \cref{sec::building_blocks} we define the basic building blocks of our architecture; in \cref{sec::action_of_building_blocks} we describe how to stack them, and the effect the architecture has on a given input; finally, in \cref{sec::proof_contextual_mapping} we prove that this architecture indeed implements a contextual mapping.

\subsubsection{Basic Building Blocks of Contextual Mapping}
\label{sec::building_blocks}
The strategy we follow to assemble a contextual mapping consists in sequentially looking at each column of the input, progressively updating and storing information regarding its content in a uniquely identifiable manner, and finally broadcasting this information back to every element in the sequence. The difficulty lies in the fact that each of these updates must be carried on while relying solely on applications of the modified $\sigmoidattn$ layer in \cref{def::modified_sigmoid_attention}. In the following, we describe how we can tweak its parameters to achieve exactly this.

\paragraph{From $d$-dimensional quantized vectors to scalars}
As a first simplification, we can get rid of the $d$-dimension in the $\bb{X}$ tensor by mapping each of its columns to a corresponding identifying scalar, uniquely defined by the specific column components. This step is also performed in \citet[App.~B.5]{Yun_UAP}, and can be achieved rather straightforwardly, by defining 
\begin{equation}\bb{v}\equiv\bb{q}\equiv\bb{k}\equiv\bb{u}\coloneqq[1,\delta^{-1},\delta^{-2},\dots,\delta^{-d+1}]^T.
    \label{eqn::vector2linear_map}
\end{equation}
Notice in fact that, since each column $\bb{x}_{i}$ belongs to $\mathbb{G}_\delta$, it can equivalently be written in the form $\bb{x}_{i}=\delta\cdot[\text{id}_{0,i}, \text{id}_{1,i},\dots,\text{id}_{d-1,i}]^T$, where $\text{id}_{j,i}\in\{0,1,2,\dots,\delta^{-1}-1\}$ represents the (indexed) coordinate of the discretization along the $j$-th dimension. Scalar-multiplying $\bb{X}$ by $\bb{u}$ in \cref{eqn::vector2linear_map}, then, turns this tuple of indices into a single one, in a bijective fashion\footnote{For example, consider $d=3$ and the column defined as $\bb{x}_{i} = [3\delta, 10\delta, 2\delta]^T$, that is, the column identified by the \emph{triplet} of indices $[3,10,2]$. Multiplying by $\bb{u}$ would then give the scalar $\bb{u}^T\bb{x}_{i} = (3 + 10N + 2N^2) \delta$, where $N=\delta^{-1}$, which is uniquely identified by the \emph{single} index $(3 + 10N + 2N^2)$.}.

This allows us to equivalently consider a single vector $\bb{u}^T\bb{X}\in\mathbb{R}^{n}$, rather than the whole tensor $\bb{X}\in\mathbb{R}^{d\times n}$ in the remainder of our analysis. Analogously, choosing $\bb{o}\equiv\bb{e}_0\coloneqq[1,0,\dots,0]^T$ in \cref{eqn::modified_sigmoid_attention} constraints the effect of the layer application to impact only the first row of the tensor: the goal is then to store in this row the result of the target contextual mapping $\bb{q}$ in \cref{def::contextual_mapping}.
To slim our notation, in the following we often refer to $\bb{u}^T\bb{X}$ as the vector $\bb{l}\in\mathbb{R}^{n}$, with components $l_i$.

In light of the simplification above, we can rewrite \cref{eqn::modified_sigmoid_attention} more compactly, as follows:
\begin{equation}
    \psi(\bb{X};\bb{q}=\bb{k}=\bb{v}\equiv\bb{u},\bb{o}\equiv\bb{e}_0;\bb{b}_q,\bb{b}_k) = \bb{e}_0\bb{l}^T H\left(\left(\bb{l}-\bb{b}_q\right)\otimes\left(\bb{l}-\bb{b}_k\right)\right)
    \label{eqn::modified_attention_1}
\end{equation}

Notice that, since the elements of both $\bb{X}$ and $\bb{u}$ are always non-negative, so are those of $\bb{l}$, too. Moreover, since we are interested in permutation-equivariant functions with respect to the columns of $\bb{X}$, without loss of generality we can consider the elements of $\bb{l} = \bb{u}^T\bb{X}$ to be ordered: $0\leq l_i< l_j$, $\forall i<j$.

\paragraph{Selective shift operation for sigmoid attention} Since we aim to recover a contextual map by sequentially updating the elements of $\bb{l}$, we proceed by designing a modification of \cref{eqn::modified_attention_1} which affects only a certain selected element at a time. This is were our second simplification comes into play, and this time it pertains the roles of the bias vectors $\bb{b}_q$ and $\bb{b}_k$. Since $\bb{l}\geq 0$, these vectors have the effect of tweaking the sign of the inner arguments of the Heaviside function in \cref{eqn::modified_attention_1}, hence directly impacting when its application outputs $0$ or $1$. By aptly selecting the values of $\bb{b}_k$ and $\bb{b}_q$, then, we can explicitly decide when a specific layer triggers an update, which elements are affected by the update, and what elements to consider to compute the update itself.

More in detail, take $\bb{b}_q=\bb{1}b_q$ and $\bb{b}_v=\bb{1}b_v$, for some scalars $b_q,b_v$, and with $\bb{1}$ being the all-one vector. Plugging this into \cref{eqn::modified_attention_1}, we have
\begin{equation}
\begin{split}
    \tilde{\psi}(\bb{X};b_q,b_k) &\coloneqq \psi(\bb{X};\bb{q}=\bb{k}=\bb{v}\equiv\bb{u},\bb{o}\equiv\bb{e}_0,\bb{b}_q=\bb{1}b_q,\bb{b}_k=\bb{1}b_k)\\
    &= \bb{e}_0\bb{l}^T H\left(\left(\bb{l}-\bb{1}b_q\right)\otimes\left(\bb{l}-\bb{1}b_k\right)\right)
    =\bb{e}_0\begin{cases}
        \sum_{i:l_i<b_v} l_i &\text{ if } l_j < b_k \\
        \sum_{i:l_i>b_v} l_i &\text{ if } l_j > b_k
    \end{cases};
    \label{eqn::modified_attention_2}
\end{split}
\end{equation}
notice how $b_q$ determines what elements of $\bb{l}$ compose the update (as it impacts the indices considered in the sum), while $b_k$ defines the elements impacted by the update itself
\footnote{\label{fnt::sigmoid_attention_mat}
This can be better seen by considering independently the effects of the two parameters $b_k$, $b_q$ on the modified sigmoid attention matrix $H\left(\left(\bb{l}-\bb{1}b_q\right)\otimes\left(\bb{l}-\bb{1}b_k\right)\right)$. We have in fact, with $b_q=0$,
\begin{equation}
    \arraycolsep=2.5pt
    \begin{array}{cc}    
    &\begin{array}{rl}
        l_j<b_k & l_j>b_k
    \end{array}\\
    \begin{array}{r}
    H\left(\bb{l}\otimes\left(\bb{l}-\bb{1}b_k\right)\right)
    \end{array}=& \left[\begin{array}{ccc|ccc}
    0      & \cdots & 0      & 1      & \cdots & 1     \\[-1ex]
    \vdots & \ddots & \vdots & \vdots & \ddots & \vdots\\
    0      & \cdots & 0      & 1      & \cdots & 1     \\
    \end{array}\right].\\
    \end{array}
\end{equation}
This shows how, by modifying $b_k$, one can decide which columns will receive an update: namely, all those with index $l_j>b_k$. By combining two such operators with $b_k=\left(i-\frac{1}{2}\right)\delta$ and $b_k=\left(i+\frac{1}{2}\right)\delta$, we then recover
\begin{equation}
    \arraycolsep=2.5pt
    \begin{array}{cc}
    &l_j=i\delta\\
    \begin{array}{r}
    H\left(\bb{l}\otimes\left(\bb{l}-\bb{1}\left(i-\frac{1}{2}\right)\delta\right)\right)\\
    -H\left(\bb{l}\otimes\left(\bb{l}-\bb{1}\left(i+\frac{1}{2}\right)\delta\right)\right) 
    \end{array}=&
    \left[\begin{array}{ccc|c|ccc}
    0      & \cdots & 0      & 1      & 0      & \cdots & 0     \\[-1ex]
    \vdots & \ddots & \vdots & \vdots & \vdots & \ddots & \vdots\\
    0      & \cdots & 0      & 1      & 0      & \cdots & 0     \\
    \end{array}\right],
    \end{array}
    \label{eqn::multihead_bq0}
\end{equation}
which allows us to limit the update to only one specific column: the one with index $l_j=i\delta$.

The parameter $b_q$ acts analogously, but varies the output of the Heaviside function as we move down the rows, rather than the columns. The same operator as in \cref{eqn::multihead_bq0}, but with $b_q=\left(i+\frac{1}{2}\right)\delta$ gives us in fact:
\begin{equation}
    \arraycolsep=2.5pt
    \begin{array}{ccc}
    &l_j=i\delta&\\
    \begin{array}{r}
        H\left(\left(\bb{l}-\bb{1}\left(i+\frac{1}{2}\right)\delta\right)\otimes\left(\bb{l}-\bb{1}\left(i-\frac{1}{2}\right)\delta\right)\right)\\
        -H\left(\left(\bb{l}-\bb{1}\left(i+\frac{1}{2}\right)\delta\right)\otimes\left(\bb{l}-\bb{1}\left(i+\frac{1}{2}\right)\delta\right)\right) \\
    \end{array}=&
    \left[\begin{array}{ccc|c|ccc}
    0      & \cdots & 0      & -1     & 0      & \cdots & 0     \\[-1ex]
    \vdots & \ddots & \vdots & \vdots & \vdots & \ddots & \vdots\\
    0      & \cdots & 0      & -1     & 0      & \cdots & 0     \\\hline
    0      & \cdots & 0      & 1      & 0      & \cdots & 0     \\[-1ex]
    \vdots & \ddots & \vdots & \vdots & \vdots & \ddots & \vdots\\
    0      & \cdots & 0      & 1      & 0      & \cdots & 0     \\
    \end{array}\right]&
    \begin{array}{c}
    l_j<i\delta\\
    \\
    \\
    l_j>i\delta\\
    \end{array}
    \end{array}.
    \label{eqn::multihead_bq}
\end{equation}
Finally, \cref{eqn::selective_shift} can be recovered by combining \cref{eqn::multihead_bq0,eqn::multihead_bq}: this has the effect of removing the $-1$'s in \cref{,eqn::multihead_bq}.}.
If we opportunely combine \emph{four} modified sigmoid self-attention heads $\tilde{\psi}(\bb{X};b_q,b_k)$, we recover, for a given index $i=0\dots\delta^{-d}-1$,
\begin{equation}
\begin{split}
    \Psi^{(i)}(\bb{X})\coloneqq& \bb{X}
    + \frac{1}{2}c \left(\begin{array}{l}
    \tilde{\psi}\left(\bb{X};b_q=0,b_k=\left(i -\frac{1}{2}\right)\delta\right)\\
    - \tilde{\psi}\left(\bb{X};b_q=0,b_k=\left(i +\frac{1}{2}\right)\delta\right)\\
    - \tilde{\psi}\left(\bb{X};b_q=b_k=\left(i +\frac{1}{2}\right)\delta\right)\\
    + \tilde{\psi}\left(\bb{X};b_q=\left(i +\frac{1}{2}\right),b_k=\left(i -\frac{1}{2}\right)\delta\right)\\
    \end{array}\right)\\
    = & \bb{X}
    + \frac{1}{2}c \bb{e}_0\bb{l}^T\left(\begin{array}{l}
    H\left(\bb{l}\otimes\left(\bb{l}-\left(i -\frac{1}{2}\right)\delta\right)\right)\\
    - H\left(\bb{l}\otimes\left(\bb{l}-\left(i +\frac{1}{2}\right)\delta\right)\right)\\
    - H\left(\left(\bb{l}-\left(i +\frac{1}{2}\right)\delta\right)\otimes\left(\bb{l}-\left(i +\frac{1}{2}\right)\delta\right)\right)\\
    + H\left(\left(\bb{l}-\left(i +\frac{1}{2}\right)\delta\right)\otimes\left(\bb{l}-\left(i -\frac{1}{2}\right)\delta\right)\right)\\
    \end{array}\right)\\
    \Longrightarrow &\Psi^{(i)}_{1,j}(\bb{X})=\bb{X}_{1,j}+c\begin{cases}
        \sum_{k:l_k> i\delta} l_k &\text{ if }  l_j = i\delta \\
        0 &\text{ otherwise}
    \end{cases}\\
    \Longrightarrow &\Psi^{(i)}_{k>1,j}(\bb{X})=\bb{X}_{k,j},
    \end{split}
    \label{eqn::selective_shift}
\end{equation}
where $c\equiv c(\delta,d,n)$ is a multiplicative constant which will be chosen later. 

The operator assembled in \cref{eqn::selective_shift} defines the basic layer of the architecture that we use in our proof. Notice $\Psi^{(i)}(\bb{X})$ has the effect of modifying only the column $\bb{x}_j$ which has index $l_j=\bb{u}^T\bb{x}_j=i\delta$ (if at all present in the input $\bb{X}$). This layer covers a similar role to the \emph{selective shift operation} introduced in \citet[App.~B.5]{Yun_UAP}, but it has been adapted to account for the presence of a sigmoid nonlinearity: notice this required us to use $4$-headed attention, while in \cite{Yun_UAP} a $2$-headed version is sufficient.

\subsubsection{Result of Applying a Sequence of Selective Shifts}
\label{sec::action_of_building_blocks}
Ultimately we want to show how, by stacking a sequence of selective shift layers \cref{eqn::selective_shift} for increasing $i=0\dots\delta^{-d}-1$ and one additional global shift, we can build an architecture capable of representing a contextual mapping. As a preliminary step, in this section we provide an explicit formula for the result of applying such an architecture. Once again, we are proceeding analogously to \citet[App.~B.5.1]{Yun_UAP}.

\paragraph{After the first selective shift application} Consider a quantized input sequence $\bb{X}\in\mathbb{L}$ as defined in \cref{eqn::quantised_input}, with its columns ordered according to their scalar indices $\bb{l}=\bb{u}^T\bb{X}$. The sequence of selective shift layers $\Psi^{(0)},\Psi^{(1)},\dots$ initially has no effect on the input itself, and it leaves it unchanged until we hit the layer corresponding to the index of the first column in the input, $\Psi^{(\hat{i})}$, where $l_1=\bb{u}^T\bb{x}_1=\hat{i}\delta$. At this point, following \cref{eqn::selective_shift}, the first column of the input is modified into
\begin{equation}
    \bb{x}_1\quad\mapsto\quad\Psi^{(\hat{i})}_{|,1}(\bb{X})
    =\bb{x}_1+ c\bb{e}_0\sum_{k:l_k>l_1}l_k
    =\bb{x}_1+ c\bb{e}_0\left(\sum_{k=1}^{n}l_k - l_1\right)
    \label{eqn::first_shift}
\end{equation}
while the other columns are still left untouched. In the following, we compactly refer to the quantities $\sum_{k=1}^{n}l_k - l_i$ as $s_i$:
\begin{equation}
    \bb{s} = [s_1,s_2,\dots,s_n]^T \coloneqq \left[\sum_{k=1}^{n}l_k - l_1, \sum_{k=1}^{n}l_k - l_2,\dots,\sum_{k=1}^{n}l_k - l_n\right]^T.
    \label{eqn::partial_sum}
\end{equation}
According to \cref{eqn::first_shift}, the index $l_1$ of column $\bb{x}_1$ is then analogously mapped to
\begin{equation}
    l_1=\bb{u}^T\bb{x}_1\quad\mapsto\quad\tilde{l}_1\coloneqq\bb{u}^T\Psi^{(\hat{i})}_{|,1}(\bb{X})
    =\bb{u}^T\bb{x}_1+ cs_1 = l_1 + cs_1.
    \label{eqn::tildel1}
\end{equation}
Notice that, by choosing $c>1$, we can ensure
\begin{equation}
    c>1\quad\Longrightarrow\quad\tilde{l}_1>\cancel{l_1} + \sum_{k=1}^{n}l_k-\cancel{l_1} > \sum_{k=1}^{n} > l_i\quad\forall i,
    \label{eqn::tildel1_is_larger}
\end{equation}
and particularly $\tilde{l}_1>l_2$, implying that at the next (effective) application of the selective shift operation, this term, too, will contribute to the update.

\paragraph{Subsequent selective shift applications}
Following similar considerations, the next effective update will be applied by the layer $\Psi^{(\hat{i})}$ with $l_2=\bb{u}^T\bb{x}_2=\hat{i}\delta$. At this point, the second column index is updated as follows:
\begin{equation}
\begin{split}
    l_2=\bb{u}^T\bb{x}_2\qquad\mapsto\qquad\tilde{l}_2\coloneqq&\bb{u}^T\Psi^{(\hat{i})}_{|,2}(\bb{X})
    =\bb{u}^T\bb{x}_2+ c\left(\sum_{k:l_k>l_2}l_k + \tilde{l}_1\right) \\
    =& l_2 + c \left(\sum_{k=1}^{n}l_k - l_2 - \cancel{l_1} + \cancel{l_1} + cs_1\right) = l_2 + c s_2 + c^2 s_1
\end{split}
\end{equation}
where $\tilde{l}_1$ is also included in light of \cref{eqn::tildel1_is_larger}, and we used the definitions \cref{eqn::tildel1,eqn::partial_sum}. Continuing to apply $\Psi^{(i)}(\bb{X})$, for increasing $i$, and unrolling the recursion, we recover
\begin{equation}
\begin{split}
    \tilde{l}_3 &= l_3 + c\left(\sum_{k=1}^{n}l_k - l_1 - l_2 -l_3 + \tilde{l}_1 + \tilde{l}_2\right)
        = l_3 + cs_3 + c^{2}(s_2+s_1) + c^{3}s_1\\
    \tilde{l}_4 &= l_4 + c\left(\sum_{k=1}^{n}l_k - l_1 - l_2 - l_3 - l_4 + \tilde{l}_1 + \tilde{l}_2 + \tilde{l}_3\right)\\
        &= l_4 + cs_4 + c^{2}(s_3+s_2+s_1) + c^{3}(s_2+2s_1) + c^{4}s_1\\
    \tilde{l}_5 &= l_5 + c\left(\sum_{k=1}^{n}l_k - l_1 - l_2 - l_3 - l_4 - l_5 + \tilde{l}_1 + \tilde{l}_2 + \tilde{l}_3 + \tilde{l}_4\right)\\
        &= l_5 + cs_5 + c^{2}(s_4+s_3+s_2+s_1) + c^{3}(s_3+2s_2+3s_1) + c^{4}(s_2+3s_1) + c^{5}s_1\\
        &\vdots
\end{split}
\label{eqn::ltilde_unroll}
\end{equation}
which eventually allows us to write the general formula
\footnote{From \cref{eqn::ltilde_unroll}, we can notice that, for a given $\tilde{l}_k$, the coefficients $a_{i,j}^{(k)}$ appearing in front of the various $s_{k-i}$ for each of the $c^{j}$ terms, are first given by a list of ones, $a_{i,1}^{(k)} = 1$, then a list of increasing numbers $a_{i,2}^{(k)}=i \Longrightarrow a_{-,2}^{(k)} = \text{cumsum}(a_{-,1}^{(k)})$, then a list of triangular numbers $a_{i,3}^{(k)}=i(i+1)/2\Longrightarrow a_{-,3}^{(k)} = \text{cumsum}(a_{-,2}^{(k)})$, and so on: $a_{-,j}^{(k)}= \text{cumsum}(a_{-,j-1}^{(k)})$. The result of iterated applications of cumsum, starting from an all-one vector, can be compactly described via the binomial coefficient: we have in fact
$$a_{i,j} = [\text{cumsum}^j([1,1,\dots])]_i=\binom{i+j-2}{j-1}.$$ The actual formula \cref{eqn::ltilde} can be recovered after a few algebraic steps, by rearranging the summation indices.}
\begin{equation}
\begin{split}
    \tilde{l}_j &\coloneqq l_j + cs_j +\sum_{i=0}^{j-2}c^{i+2}\sum_{k=i}^{j-2}\binom{k}{i}s_{k-i+1}, \qquad\qquad j=1\dots n.
    \label{eqn::ltilde}
\end{split}
\end{equation}

\subsubsection{Result of Applying One Last \emph{Global Shift} Layer}
After the last selective shift layer, the original input $\bb{X}$ has been mapped to a modified one $\tilde{\bb{X}}$ whereby each column $\tilde{\bb{x}}_j$ is characterized by the index $\tilde{l}_j=\bb{u}^T\tilde{\bb{x}}_j$ given in \cref{eqn::ltilde}. Remember our goal is to recover a contextual mapping, but notice that these $\tilde{l}_j$ indices are \emph{not} uniquely defined by the input\footnote{To convince ourselves of this, it suffices to look at the formula for \cref{eqn::tildel1}: two sequences with different elements $\bb{l}\neq\bb{l}'$, but such that $l_1=l_1'$ and $s_1=s_1'$ (that is, with $\sum_{i=1}^n l_i=\sum_{i=1}^n l_i'$) would map to the same $\tilde{l}_1=\tilde{l}_1'$.}; in other words, they do not satisfy property (2) in \cref{def::contextual_mapping}. The only exception to this is the last index $\tilde{l}_n$, as (loosely speaking) it has ``seen'' all the previous updates - and indeed in \cref{sec::proof_contextual_mapping} we prove this rigorously, under some assumption on the yet-undefined coefficient $c(\delta,d,n)$.

A straightforward way to recover a one-to-one mapping for the whole sequence, then, is to update every index $\tilde{l}_j$ via a quantity directly depending on $\tilde{l}_n$. This is precisely what the last \emph{global shift} layer $\bar{\Psi}(\bb{X})$ aims to accomplish. This last layer is also defined starting from the simplified modified sigmoid attention \cref{eqn::modified_attention_2}, by picking $b_k=0$ and $b_q=\left(c(\delta,d,n)^n+\frac{1}{2}\right)\delta$: if, for any input, we can guarantee that
\begin{equation}
    \tilde{l}_j\leq c(\delta,d,n)^n\delta\quad j<n\qquad\text{and}\qquad \tilde{l}_n>c(\delta,d,n)^n\delta,
    \label{eqn::last_layer_condition}
\end{equation}
then the application of the global shift layer would result in\footnote{As in \cref{fnt::sigmoid_attention_mat}, this is also better seen by considering the resulting modified sigmoid attention matrix. With $b_k=0$ and $b_q=\left(c(\delta,d,n)^n+\frac{1}{2}\right)\delta$, in fact, if condition \cref{eqn::last_layer_condition} is verified, this matrix is given by
\begin{equation}
    \arraycolsep=2.5pt
    \begin{array}{rcl}
    \begin{array}{c}
        H\left(\left(\tilde{\bb{l}}-\bb{1}\left(c^n+\frac{1}{2}\right)\delta\right)\otimes\tilde{\bb{l}}\right)
    \end{array}=&
    \left[\begin{array}{ccc}
    0      & \cdots & 0      \\[-1ex]
    \vdots & \ddots & \vdots \\
    0      & \cdots & 0      \\\hline
    1      & \cdots & 1      \\
    \end{array}\right]&
    \begin{array}{l}
    \\
    \\
    \tilde{l}_j,\,j<n\\
    \\
    \tilde{l}_n\\
    \end{array}
    \end{array}.
\end{equation}
}:
\begin{equation}
\begin{split}
    \bar{\Psi}(\tilde{\bb{X}})\coloneqq&\tilde{\bb{X}} + c^{n+1}\tilde{\psi}\left(\tilde{\bb{X}};b_q=\left(c^n+\frac{1}{2}\right)\delta,b_k=0\right)\\
    \Longrightarrow &\bar{\Psi}_{1,j}(\tilde{\bb{X}}) = \tilde{\bb{X}}_{1,j} + c^{n+1}\tilde{l}_n\\
    \Longrightarrow &\bar{\Psi}_{k>1,j}(\tilde{\bb{X}})=\tilde{\bb{X}}_{k,j}.
\end{split}
\label{eqn::global_shift}
\end{equation}

The global shift \cref{eqn::global_shift} is the last layer we need to define our candidate contextual mapping. Collecting the results from this section together, our architecture is defined by sequentially composing the selective shift layers with the global shift one,
\begin{equation}
    \Psi(\bb{X}) \coloneqq \bar{\Psi}\circ\Psi^{(\delta^{-d}-1)}\circ\cdots\circ\Psi^{(2)}\circ\Psi^{(1)}(\bb{X}).
\end{equation}
After being scalar-multiplied by $\bb{u}$, this results in a sequence
\begin{equation}
    \bb{q}(\bb{X}) \coloneqq \bb{u}^T\Psi(\bb{X}) = \tilde{\bb{l}} + c^{n+1} \bb{1}\tilde{l}_n
    \label{eqn::final_q}
\end{equation}
which we aim to prove is a contextual mapping. This is shown in the next section.

\subsubsection{A Sequence of Selective Shifts Followed by a Global Shift Produces a Contextual Mapping}
\label{sec::proof_contextual_mapping}
To complete the proof, it remains to show that the recovered sequence \cref{eqn::final_q} represents a contextual mapping and, in particular, that it is \emph{(i)} one-to-one in $\mathbb{L}$, and that \emph{(ii)} all of its elements are distinct for different inputs. To do so, we need a few preparatory lemmas. The first few are needed to show that each of the basic components of \cref{eqn::final_q} is indeed a one-to-one map.

\begin{lemma}
\label{thm::S_1to1}
The map $\bb{l}\mapsto\bb{s}$ in \cref{eqn::partial_sum} is one-to-one.
\end{lemma}
\begin{proof}
The target map can be compactly represented as a linear operator $S$:
\begin{equation}
    \bb{l}\mapsto \bb{s}\coloneqq \bb{1}\sum_{k=1}^n l_k - \bb{l} = (\bb{1}\otimes\bb{1} - I)\bb{l}\eqqcolon S\bb{l}
    \label{eqn::s_as_linear_op}
\end{equation}
which is invertible\footnote{Indeed its inverse can be explicitly recovered by directly applying Sherman-Morrison formula.}, denoting that $\bb{l}\mapsto\bb{s}$ is bijective.
\end{proof}

\begin{lemma}
\label{thm::ln_1to1}
The map $\bb{l}\mapsto\tilde l_n$ in \cref{eqn::ltilde} is one-to-one, under the condition 
\begin{equation}
    c(\delta,d,n)>(n-1)(\delta^{-d}-1)\binom{n-1}{\left\lceil\frac{n-1}{2}\right\rceil}.
    \label{eqn::condition_on_c_final}
\end{equation}
\end{lemma}
\begin{proof}
Consider two vectors of column indices $\bb{l},\bb{l}'$ differing for at least one element. We have by definition \cref{eqn::ltilde} that
\begin{equation}
    \tilde{l}_n-\tilde{l}_n' = 
        (l_n-l_n') + c(s_n-s_n')
        +\sum_{i=0}^{n-2}c^{i+2}\sum_{k=i}^{n-2}\binom{k}{i}(s_{k-i+1}-s_{k-i+1}')
\end{equation}
By absurd, assume $\tilde{l}_n-\tilde{l}_n'=0$ even though $\exists i:l_i\neq l_i'$. We have then that it must hold
\begin{equation}
\begin{split}
    (l_n'-l_n) &= c(s_n-s_n')
    +\sum_{i=0}^{n-2}c^{i+2}\sum_{k=i}^{n-2}\binom{k}{i}(s_{k-i+1}-s_{k-i+1}')\\
    &=c\left((s_n-s_n')
    +\sum_{i=0}^{n-2}c^{i+1}\sum_{k=i}^{n-2}\binom{k}{i}(s_{k-i+1}-s_{k-i+1}')\right)
\end{split}
\label{eqn::ln'-ln}
\end{equation}
Notice that, for $c(\delta,d,n)$ large enough, the right-hand side does not have enough \emph{granularity} to counter the left-hand side: in fact, since $l_n\in\{0,\delta,2\delta,\dots,\delta^{-d+1}-\delta\}$, the left-hand side can attain values
\begin{equation}
    l_n'-l_n\in\{0,\pm\delta,\pm2\delta,\dots,\pm(\delta^{-d+1}-\delta)\}
\end{equation}
while the former, in light of the presence of the $c(\delta,d,n)$ factor, can only attain values $\in\{0,\pm c\delta,\pm2c\delta,\dots\}$. Picking $c>\delta^{-d}-1$, then, ensures that equality between the two sides of \cref{eqn::ln'-ln} can only be achieved if they are both $0$. In this case, we need to impose
\begin{equation}
\begin{split}
    &c(s_n'-s_n) = 
    \sum_{i=0}^{n-2}c^{i+1}\sum_{k=i}^{n-2}\binom{k}{i}(s_{k-i+1}-s_{k-i+1}')\\
    \quad\Longleftrightarrow\quad& s_n'-s_n %
    =c\left(\sum_{i=0}^{n-2}c^i\sum_{k=i}^{n-2}\binom{k}{i}(s_{k-i+1}-s_{k-i+1}')\right).
\end{split}
\label{eqn::condition_on_sn_for_1to1}
\end{equation}
Similarly, notice that\footnote{This is a direct consequence of the definition of operator $S$ in \cref{eqn::s_as_linear_op}: since it has $1$'s everywhere but on its diagonal, its $\infty$-norm is simply $n-1$.}, $\forall i$,
\begin{equation}
    |s_i-s_i'|=\left|\sum_{k=1}^{n} (l_k-l_k') - (l_i-l_i')\right|=\left|\sum_{k=1,k\neq i}^{n} (l_k-l_k')\right|<(n-1)(\delta^{-d+1}-\delta),
\end{equation}
implying that $s_n'-s_n\in\{0,\pm\delta,\pm2\delta,\dots,\pm(n-1)(\delta^{-d}-1)\delta\}$. Again, by picking $c(\delta,d,n)>(n-1)(\delta^{-d}-1)$ we ensure that the right-hand side does not have enough granularity, and hence
\begin{equation}
    c(\delta,d,n)>(n-1)(\delta^{-d}-1) \qquad\Longrightarrow\qquad s_n'-s_n=0,  
    \label{eqn::conditions_on_c_for_1to1_0}
\end{equation}
implying
\begin{equation}
\begin{split}
    &c\left(\sum_{i=0}^{n-2}c^i\sum_{k=i}^{n-2}\binom{k}{i}(s_{k-i+1}-s_{k-i+1}')\right)=0\\
    \Longleftrightarrow\quad& \sum_{k=0}^{n-2}\binom{k}{0}(s_{k+1}'-s_{k+1}) = c\left(\sum_{i=1}^{n-2}c^{i-1}\sum_{k=i}^{n-2}\binom{k}{i}(s_{k-i+1}-s_{k-i+1}')\right)\\
    \Longleftrightarrow\quad& \sum_{k=0}^{n-2}(s_{k+1}'-s_{k+1}) = c\left(\sum_{i=1}^{n-2}c^{i-1}\sum_{k=i}^{n-2}\binom{k}{i}(s_{k-i+1}-s_{k-i+1}')\right).\\
\end{split}
\end{equation}
Following a similar reasoning as the one applied above shows us that picking
\begin{equation}
    c(\delta,d,n)>(n-1)^2(\delta^{-d}-1) \qquad\Longrightarrow\qquad \sum_{k=0}^{n-2}(s_{k+1}-s_{k+1}')=0,
\end{equation}
and requires us to satisfy
\begin{equation}
\begin{split}
    & c\left(\sum_{i=1}^{n-2}c^{i-1}\sum_{k=i}^{n-2}\binom{k}{i}(s_{k-i+1}-s_{k-i+1}')\right)=0\\
    \Longleftrightarrow\quad& \sum_{k=1}^{n-2}\binom{k}{1}(s_{k}'-s_{k}) = c\left(\sum_{i=2}^{n-2}c^{i-2}\sum_{k=i}^{n-2}\binom{k}{i}(s_{k-i+1}-s_{k-i+1}')\right)\\
    \Longleftrightarrow\quad& \sum_{k=1}^{n-2}k(s_{k}'-s_{k}) = c\left(\sum_{i=2}^{n-2}c^{i-2}\sum_{k=i}^{n-2}\binom{k}{i}(s_{k-i+1}-s_{k-i+1}')\right).\\
\end{split}
    \label{}
\end{equation}
Once again, then, by choosing 
\begin{equation}
    c(\delta,d,n)>\frac{(n-2)(n-1)^2}{2}(\delta^{-d}-1)\qquad\Longrightarrow\qquad\sum_{k=1}^{n-2}k(s_{k}-s_{k}')=0.
\end{equation}
This reasoning can be repeated recursively: at each step $i$ of the recursion, by imposing a stricter and stricter bound on $c(\delta,d,n)$ we gain more and more conditions that the quantity $\bb{s}'-\bb{s}$ needs to satisfy:
\begin{equation}
    c(\delta,d,n)>(n-1)(\delta^{-d}-1)\sum_{k=i}^{n-2}\binom{k}{i}\qquad\Longrightarrow\qquad\sum_{k=i}^{n-2}\binom{k}{i}(s_{k-i+1}-s_{k-1+1}') = 0.
    \label{eqn::conditions_on_c_for_1to1_i}
\end{equation}
Notice that, every time we increase $i=0\dots n-2$, these conditions involve one less term $s_{k-i+1}-s_{k-i+1}'$, $k=i\dots n-2$: if we were to collect all these conditions within a single linear system, the system would have an upper-triangular structure, and hence be non-singular. This implies that for the set of $n$ independent conditions on $\bb{s}-\bb{s}'$ to hold (we have $n-1$ in \cref{eqn::conditions_on_c_for_1to1_i}, plus one more in \cref{eqn::conditions_on_c_for_1to1_0}), the only possibility is that $\bb{s}\equiv\bb{s}'$. Because of \cref{thm::S_1to1}, though, this also implies $\bb{l}\equiv\bb{l}'$: we have finally reached a contradiction, and proven that indeed $\bb{l}\mapsto\tilde{l}_n$ is one-to-one, under an opportune condition on $c(\delta,d,n)$.
Such condition can be promptly recovered\footnote{This is a consequence of some useful properties of the binomial coefficient, namely the Hockey stick identity \cite{HockeyStick}, and the symmetry of $\binom{k}{i}$ with respect to $i$.} by \cref{eqn::conditions_on_c_for_1to1_i}:
\begin{equation}
    \max_{i=0 \dots n-2}\sum_{k=i}^{n-2}\binom{k}{i} =\max_{i=0 \dots n-2}\binom{n-1}{i+1}=\binom{n-1}{\left\lceil\frac{n-1}{2}\right\rceil}.
    \label{eqn::max_binomial}
\end{equation}
Substituting this in \cref{eqn::conditions_on_c_for_1to1_i}, we recover that it suffices to impose
\begin{equation}
    c(\delta,d,n)>(n-1)(\delta^{-d}-1)\binom{n-1}{\left\lceil\frac{n-1}{2}\right\rceil}.
\end{equation}
\end{proof}

The next few lemmas are needed to bound the elements in the $\tilde{l}_j$ sequence, which in turn are used to prove property \emph{(ii)} in \cref{def::contextual_mapping}.

\begin{lemma}
\label{thm::ltilde_increasing}
    $\tilde{l}_j$ in \cref{eqn::ltilde} is an increasing sequence.
\end{lemma}
\begin{proof}
This can be proven directly: we have in fact, by definition \cref{eqn::ltilde},
    \begin{equation}
    \begin{split}
        \tilde{l}_j>\tilde{l}_{j-1} \quad\Longleftrightarrow\quad&
            l_j + cs_j +\sum_{i=0}^{j-2}c^{i+2}\sum_{k=i}^{j-2}\binom{k}{i}s_{k-i+1}\\
            &>l_{j-1} + cs_{j-1} +\sum_{i=0}^{j-3}c^{i+2}\sum_{k=i}^{j-3}\binom{k}{i}s_{k-i+1}\\
            \prescript{\text{\tiny combine sums}}{}{\quad\Longleftrightarrow\quad}&
            (l_j-l_{j-1})(1-c) +\sum_{i=0}^{j-2}c^{i+2}\binom{j-2}{i}s_{j-1-i}>0\\
            \prescript{\text{\tiny $\binom{j-2}{i}\geq1,c^{i+2}\geq c^2$}}{}{\quad\Longleftarrow\quad}&
            (l_j-l_{j-1})(1-c) +c^2\sum_{i=0}^{j-2}s_{j-1-i}>0\\
            \prescript{\text{\tiny \cref{eqn::partial_sum}}}{}{\quad\Longleftrightarrow\quad}&
            (l_j-l_{j-1})(1-c) +c^2\sum_{i=0}^{j-2}\left(\sum_{k=1}^n l_k - l_{j-1-i}\right)>0\\
            \quad\Longleftrightarrow\quad&(l_j-l_{j-1})(1-c) +c^2\left((j-1)\sum_{k=1}^n l_k - \sum_{k=1}^{j-1}l_{k}\right)>0\\
            \quad\Longleftrightarrow\quad&(1-c)l_j+(c-1)l_{j-1} +c^2(j-2)\sum_{k=1}^n l_k + c^2\sum_{k=j}^{n}l_{k}>0\\
            \quad\Longleftrightarrow\quad&(c^2-c+1)l_j+(c-1)l_{j-1}
            +c^2(j-2)\sum_{k=1}^n l_k + c^2\sum_{k=j+1}^{n}l_{k}>0\\
    \end{split}
    \end{equation}
    Already with $c>1$, all the coefficients are positive (and at least one is non-zero), implying that the condition above is always satisfied and that indeed $\tilde{l}_j$ is an increasing sequence.
\end{proof}

\begin{lemma}
\label{thm::ltilde_bounds}
    Under constraint \cref{eqn::condition_on_c_final}, each term $\tilde{l}_j$, $j>1$ in \cref{eqn::ltilde} is bounded from below by
    $$\tilde{l}_j>c^j\delta,$$
    and each term $\tilde{l}_j$, $1<j<n$ is bounded from above by
    $$\tilde{l}_j<c^{j+1}\delta.$$
\end{lemma}
\begin{proof}
    We start by proving the lower bound. By definition \cref{eqn::ltilde}, we have
    \begin{equation}
        \tilde{l}_j = l_j+cs_j +\sum_{i=0}^{j-2}c^{i+2}\sum_{k=i}^{j-2}\binom{k}{i}s_{k-i+1} 
                    = l_j+cs_j +c^j s_{1} + \sum_{i=0}^{j-3}c^{i+2}\sum_{k=i}^{j-2}\binom{k}{i}s_{k-i+1}.
    \label{eqn::tildelj_simp}
    \end{equation}
    Since by assumption $l_j$ is an ordered sequence without repetitions, for $j>1$ we necessarily have $l_j>l_1\geq0$, and hence $l_j\geq\delta$. All the other terms in \cref{eqn::tildelj_simp} are non-negative, so we can safely claim that
    \begin{equation}
        \tilde{l}_j \geq \delta + c^j\delta > c^j\delta \qquad\forall j>1,
    \end{equation}
    which confirms the lower bound.
    
    For the upper bound, we start again from the definition of $\tilde{l}_j$:
    \begin{equation}
    \begin{split}
        \tilde{l}_{j} &= l_{j} + cs_{j} +\sum_{i=0}^{j-2}c^{i+2}\sum_{k=i}^{j-2}\binom{k}{i}s_{k-i+1}\\
            &<    (\delta^{-d}-1)\delta + c(n-1)(\delta^{-d}-1)\delta + s_1\sum_{i=0}^{j-2}c^{i+2}\binom{j-1}{i+1}\\
            &\leq (n-1)(\delta^{-d}-1)\binom{j-1}{\left\lceil\frac{j-1}{2}\right\rceil}\delta \sum_{i=0}^{j}c^{i}
            = (n-1)(\delta^{-d}-1)\binom{j-1}{\left\lceil\frac{j-1}{2}\right\rceil}\delta \frac{1-c^{j+1}}{1-c},
    \end{split}
    \label{eqn::tildelj_simp2}
    \end{equation}   
    where we used relationship \cref{eqn::max_binomial} and collected all $c$ terms within the sum. Notice that, for a given $a>1$ we have that
    \begin{equation}
        \frac{1-c^{j+1}}{1-c}\leq ac^{j},
        \label{eqn::ratio_of_cs}
    \end{equation}
    provided that $c\geq\frac{a}{a-1}$. In fact,
    \begin{equation}
    \begin{split}
        \frac{1-c^{j+1}}{1-c} \leq ac^{j} \quad\Longleftrightarrow\quad& \frac{1-c^{j+1}-ac^{j}+ac^{j+1}}{1-c} \leq 0 \\
        \quad\Longleftarrow\quad& \frac{1}{a-1}+\left(c-\frac{a}{a-1}\right)c^{j} \geq 0
        \quad\Longleftarrow\quad \frac{1}{a-1} \geq 0\\
    \end{split}
    \end{equation}
    which is always satisfied. After substituting \cref{eqn::ratio_of_cs} in \cref{eqn::tildelj_simp2}, this allows us to write
    \begin{equation}
        \tilde{l}_{j} < a(n-1)(\delta^{-d}-1)\binom{j-1}{\left\lceil\frac{j-1}{2}\right\rceil}\delta c^{j}.
    \end{equation}   
    To prove that $\tilde{l}_{j}<\delta c^{j+1}$, then, it remains to show that
    \begin{equation}
        c \geq a(n-1)(\delta^{-d}-1)\binom{j-1}{\left\lceil\frac{j-1}{2}\right\rceil}\qquad\forall 1<j<n.
    \end{equation}   
    Substituting condition \cref{eqn::condition_on_c_final} in the inequality above, we are left with proving
    \begin{equation}
        \binom{n-1}{\left\lceil\frac{n-1}{2}\right\rceil} \geq \max_{j=2\dots n-1}a\binom{j-1}{\left\lceil\frac{j-1}{2}\right\rceil}=a\binom{n-2}{\left\lceil\frac{n-2}{2}\right\rceil}.
    \end{equation}   
    The outcome depends on the parity of $n$. For $n$ odd, we have
    \begin{equation}
        \binom{n-1}{\left\lceil\frac{n-1}{2}\right\rceil} \geq a\binom{n-2}{\left\lceil\frac{n-2}{2}\right\rceil}
        \qquad\Longleftrightarrow\qquad 2\frac{n-1}{n-1}\geq a,
    \end{equation}   
    to satisfy which it suffices to pick $a=2$. This requires having $c\geq\frac{a}{a-1}=2$, which is automatically satisfied. For $n$ even, on the other hand, the binomial coefficients simplify to
    \begin{equation}
        \binom{n-1}{\left\lceil\frac{n-1}{2}\right\rceil} \geq a\binom{n-2}{\left\lceil\frac{n-2}{2}\right\rceil}
        \qquad\Longleftrightarrow\qquad 2\frac{n-1}{n}\geq a.
    \end{equation}   
    To satisfy this, we need to pick $a=2\frac{n-1}{n}$, which requires $c\geq\frac{a}{a-1}=2\frac{n-1}{n-2}$; however, this too is automatically satisfied by \cref{eqn::condition_on_c_final} provided $n\geq4$. This completes the proof.
\end{proof}

\begin{lemma}
    Under the constraint \cref{eqn::condition_on_c_final}, condition \cref{eqn::last_layer_condition} holds.
\end{lemma}
\begin{proof}
    We remind that condition \cref{eqn::last_layer_condition} is necessary for the correct ``functioning'' of the global shift layer, and it composes of two parts.
    The first part requires that $\tilde{l}_{j}<c^n\delta$ $\forall j<n$. Thanks to \cref{thm::ltilde_increasing}, it suffices to show that $\tilde{l}_{n-1}<c^n\delta$, but this is already granted by the upper bound in \cref{thm::ltilde_bounds}. Analogously, for the second part, we need to show that $\tilde{l}_n>c^n\delta$: for this too we can use the lower bound in \cref{thm::ltilde_bounds}.    
\end{proof}

We finally have all the ingredients to prove the main theorem of this section:
\begin{theorem}
    The map in \cref{eqn::final_q}, given by
    $$\bb{X}\mapsto\bb{q}(\bb{X}) = \bb{u}^T\Psi(\bb{X})$$
    represents a contextual mapping.
\end{theorem}
\begin{proof}
    As defined in \cref{def::contextual_mapping}, a contextual mapping must satisfy two conditions. The first one is that
    \begin{equation}
        q_i(\bb{X})\neq q_j(\bb{X}), \quad\forall i\neq j\qquad\text{and}\qquad\forall\bb{X}\in\mathbb{L}.
    \end{equation}
    This is directly proven by considering \cref{thm::ltilde_increasing}: since $\tilde{l}_j$ is a (strictly) increasing sequence, all its elements are already distinct. The action of the last global shift layer merely translates all these elements by a same quantity, but they remain distinct nonetheless.
    
    The second condition for a contextual mapping is given by
    \begin{equation}
        q_i(\bb{X})\neq q_j(\bb{X}'),\quad\forall i,j\quad\text{and}\quad\forall\bb{X},\bb{X}'\in\mathbb{L},\quad \text{with}\quad\bb{X}\neq\bb{X}'.
    \end{equation}
    We prove that this holds for \cref{eqn::final_q} by directly considering the difference between two components $i,j$ for different inputs:
    \begin{equation}
        q_i(\bb{X})- q_j(\bb{X}') = \tilde{l}_i-\tilde{l}_j' + c^{n+1}\left(\tilde{l}_n-\tilde{l}_n'\right)=0
        \quad\Longleftrightarrow\quad \tilde{l}_i-\tilde{l}_j' = c^{n+1}\left(\tilde{l}_n'-\tilde{l}_n\right).
    \end{equation}
    Notice that, due to \cref{thm::ln_1to1}, we have $\tilde{l}_n-\tilde{l}_n'\neq0$ and particularly, $|\tilde{l}_n-\tilde{l}_n'|\geq\delta$. On the other hand, in light of the bounds in \cref{thm::ltilde_bounds}, we have that the left-hand side $|\tilde{l}_j-\tilde{l}_i|<c^{n}\delta$. Consequently, the two sides can never cancel each other out, and the proof is complete.
\end{proof}

\section{Lipschitzness of Sigmoid Attention}
\label{app:lipschitz_proof}
In the following, we report the proof for the recovering the Lipschitzness constant associated with $\sigmoidattn$, as stated in \cref{thm:regularity}.

Letting $A = W_q^T W_k$, and calling $\sigma_{ij} = \sigma(\langle W_qx_i, W_kx_j\rangle)$ and $\sigma'_{ij} = \sigma'(\langle W_qx_i, W_kx_j\rangle)$, we find that the Jacobian of $\phi$ in the direction $(\delta_1, \dots, \delta_n)$ for the sample $x_i$ is given by:
\begin{equation}
    \mathrm{Jac}_i = \left(\sum_{j=1}^n\sigma'_{ij}x_jx_j^TA^T\right)\delta_i +  \sum_{j=1}^n\left(\sigma'_{ij}x_jx_i^TA + \sigma_{ij}I_p\right)\delta_j,
\end{equation}
We see that this Jacobian is the sum of two terms. To control its norm, we can control each norm individually. 

The first term, $\left(\sum_{j=1}^n\sigma'_{ij}x_jx_j^TA^T\right)\delta_i$ is of the form $U_i\delta_i$ with $U_i$ a matrix. 
Its squared-norm is therefore:
\begin{equation}
    \sum_{i=1}^n \|U_i\delta_i\|^2 \leq \max_{i}\|U_i\|_2^2 \|\delta\|_F .
\end{equation}
Hence, its squared spectral norm is bounded by $\max_{i}\|U_i\|_2^2$.

We now let $\sigma'_{\infty}$ be a bound on $n\times |\sigma'|$; 
We have: 
\begin{align}
\|U_i\|_2 & \leq \sum_{j=1}^n\|\sigma'_{ij}x_jx_j^{\top}A\|_2 \\
& \leq \sigma'_{\infty}\|A\|_2\frac1n \sum_{j=1}^n \|x_j\|^2 \\
&\leq \sigma'_{\infty}\|A\|_2 \mathbb{E}[\|x_j\|^2].
\label{eqn:lip_bound_first}
\end{align}
We see that if the points $x_i$ have norm $\leq R$, then the Jacobian grows at most like $R^2$, because it is ``quadratic'' in $x$.
However, we see that the quadratic term is likely to be mitigated by the $\sigma'(a_{ij})$ term that goes to $0$ if $a_{ij}$ is large.

The second term, $\sum_{j=1}^n\left(\sigma'_{ij}x_jx_i^TA + \sigma_{ij}I_p\right)\delta_j$, is the sum of two terms. Here, too, we use the triangular inequality to control their norm individually. We get:
\begin{align}
    \|\sum_{j=1}^n\sigma_{ij}\delta_j\|^2 &=\|\delta^T \sigma_i\|^2 \\
    &\leq \|\delta\|_F^2 \|\sigma_i\|^2,
\end{align}
where $\sigma_i\in\mathbb{R}^p$ is the $i$-th column of $\sigma_{ij}$, and $\delta\in \mathbb{R}^{n\times p}$.
and by summing, letting $\sigma_{\infty}$ an upper bound on $n\times |\sigma(x)|$:
\begin{equation}
    \sum_{i=1}^n\|\sum_{j=1}^n\sigma_{ij}\delta_j\|^2 \leq \sigma_{\infty}^2 \|\delta\|_F^2 .
\end{equation}
So that $\sigma_{\infty}$ upper bounds the spectral norm of the last term.

For the final term, $\sum_{j=1}^n\sigma'_{ij}x_jx_i^TA\delta_j$, define $\hat{\delta} = \delta A^T$. We get:
\begin{equation}
    \sum_{j=1}^n\sigma'_{ij}x_jx_i^TA\delta_j = \sum_{j=1}^n\sigma'_{ij}\langle x_i,\hat{\delta}_j\rangle x_j .
\end{equation}
Hence, letting $M$ the matrix of entries $M_{ij} = \sigma'_{ij}\langle x_i,\hat{\delta}_j\rangle$, we see that the previous term is simply $x^TM_i^T$, so that we get the upper bound on the norm of the term:
\begin{equation}
    \sum_{i=1}^n \|x^TM_i^T\|^2 \leq \|x\|_F^2 \|M\|_F^2 
\end{equation}
and $\|M\|_F^2 = \sum_{ij}(\sigma'_{ij})^2\langle x_i,\hat{\delta}_j\rangle^2 \leq \frac{1}{n^2} \sigma'_{\infty}\|x\|_F^2\|A\|_2^2\|\delta\|_F^2$, giving overall:
\begin{equation}
    \sqrt{\sum_{i=1}^n \|x^TM_i^T\|^2} \leq \sigma'_{\infty} \|A\|_2\mathbb{E}[\|x_j\|^2]\|\delta\|_F .
\end{equation}
Notice how this quantity matches the one in \cref{eqn:lip_bound_first}.

Finally, summing all together gives:
\begin{equation}
    \|\mathrm{Jac}\|_2\leq 2\sigma'_{\infty} \|A\|_2\mathbb{E}[\|x_j\|^2] + \sigma_{\infty},
\end{equation}
which completes the proof.

\textbf{Remark}: The previous upper bound might not be tight. Indeed, intuitively, if the $x_i$ are large, then the term $\sigma'_{ij}$ should be exponentially small (provided, of course, that $W_qx_i$ and $W_kx_j$ are not orthogonal), which would even remove the dependency on the variance in the sigmoid attention.

\section{The Bias Term of Sigmoid Attention}
\label{app:sigmoid_bias}

One of the differences between $\sigmoidattn$ and $\softmaxattn$ is the normalization constant. In $\sigmoidattn$, one way to emulate the effect of a normalization constant (which links all the elements of the input together and defines a distribution over them), is to include a bias term in the definition as proposed in $\cref{eq:sigmoid_attn}$.

For an input vector $\vz\in\mathbb{R}^n$, the output of the sigmoid with bias $b$ is $$
\sigma^b(\vz)_i := \frac{\exp(z_i)}{\exp(z_i) + \exp(-b)}
$$
Contrary to the softmax, this output cannot always sum to one because there is no normalization.
We therefore seek a value for $b$ that \emph{approximately} normalizes $\sigma^b(\vz)$, i.e., such that $\sum_{i=1}^n \sigma^b(\vz)_i \simeq 1$.
We have
\begin{proposition}
    Let $\vz \in\mathbb{R}^n$, and take $m, M \in\mathbb{R}$ such that for all $i$, it holds $m\leq z_i\leq M$. Then, the equation $\sum_{i=1}^n \sigma^b(\vz)_i = 1$ with variable $b$ has a single solution $b^*$ with
    $$-\log(n-1) - M\leq b^*\leq -\log(n - 1) - m\enspace.$$
\end{proposition}
\begin{proof}
    The function $\phi: b\to \sum_{i=1}^n \sigma^b(\vz)_i$ is smooth and monotonically increasing, and we have $\phi(-\log(n-1)  - M) \leq 1$ and $\phi(-\log(n-1) - m)\geq1$. This shows the existence of $b^*$ as well as the advertised bound on $b^*$. 
\end{proof}
This suggests using a $b$ of the order of $-\log(n)$; in practice we use $b=-\log(n)$.

We can also look for a bias term $b$, which helps to approximate the softmax function by the sigmoid function.

We assume that softmax provides us with the true distribution $p^\star$, where $p_i^\star = \frac{e^{z_i}}{e^{z_i} + \sum_{j \neq i} e^{z_j}}$. The goal is to find the bias term $b$ such that sigmoid function with weights over all elements denoted by $p$, where $p_i = \sigma^b(\vz)_i$, approximates $p^\star$. Note that, as mentioned before, $p$ is not necessarily a distribution, i.e.\ $\sum_{i = 1}^n p_i$ is not always equal to one.

In technical terms, we aim to estimate the normalizing factor $\mZ = \sum_{i = 1}^n e^{z_i}$. The existing approaches for estimating $\mZ$ is compute-expensive for high dimensions and requires resampling methods.
Also, the optimal value of $b$ would depend on the exact values of $\vz$, which is unknown beforehand. Therefore, we propose a more intuitive way to estimate the order of bias but possibly with larger disparity.
To distribute the independent masses in $\sigmoidattn$, we assume that each element has uniform weight for the model apriori, which means that none of the elements of the input vector $\vz$ has any known importance over the others. In the simplest case when softmax is a uniform distribution, we ideally want to have the same order of values for sigmoid as of softmax, which should be $\frac{1}{n}$. Therefore, we can write down the following:
\begin{align}
    \forall i \qquad p_i = \frac{1}{1 + e^{-(z_i + b)}} \simeq \frac{1}{n} = p_i^\star
\end{align}
Ideally, we would like to have $1 + e^{-(z_i + b)} \simeq n$. Requiring that $p=p^*$ in the case where all the $z_i$ are $0$ gives $\exp(-b) = n - 1$, i.e. $b \simeq -\log(n)$ for large $n$. In the case that all the $z_i$ are bounded, $|z_i| \leq M < \infty$ for some constant $M$, then $b \simeq -(M + \log(n)) \approx -\max\{M, \log(n)\}$. However, in most cases we do not know $M$. When the sequence length $n$ is large enough, the constant $M$ loses its importance while in short sequence length, it impacts distributing the weights over elements more. To resolve this issue, we assume that $z_i$ are sampled from a standard Gaussian distribution, i.e. $z_i \sim \mathcal{N}(0, \sigma^2)$ where $\sigma = 1$. Note that this assumption comes from the fact that $z_i$ in our problem is one of the elements of $\mQ\mK^T / \sqrt{d_{qk}}$, which is the sum of $d_{qk}$ random variables. Using Central Limit Theorem, we can assume that $z_i$ is sampled from a Gaussian distribution. The idea is to estimate $M$, such that with high probability, $|z_i| \leq M$, i.e. $\mathbb{P}\left(|z_i| > M\right) \leq \epsilon$ for a desired $\epsilon$. Therefore, we have
\begin{align}
    \mathbb{P}\left(|z_i| > M\right)  = \mathbb{P}\left(|z_i| > \frac{M}{\sigma}\sigma\right) \leq \frac{1}{(\frac{M}{\sigma})^2} = \frac{\sigma^2}{M^2} \leq \epsilon,
\end{align}
where the inequality is resulted from Chebychev's inequality. Setting $\sigma = 1$, we have $M \simeq \sqrt{1/\epsilon}$. Therefore, the order-optimal value would be $b \simeq -\max\{\sqrt{1/\epsilon}, \log(n)\}$, and for long sequence length, $b \simeq -\log(n)$. For example, if we want $90\%$ accuracy in our estimation, $M \approx 3\sigma = 3$, which means $b \simeq -\max\{3, \log(n)\}$.
Note that this approximation also follows the intuition that as $n$ grows, we expect the $\sigmoidattn$ without bias term overestimate the mass on each point, so we need to normalize the mass according to $n$ at each point as well. 

On another side, one may be more interested in the gradients of $p^\star$ and $p$ with respect to $z_i$ to behave similarly. We show that $b \simeq -\log(n)$ is still a good choice in this scenario. Let us derive the derivative of $\sigmoidattn$ and $\softmaxattn$ with respect to the input. We note that for any $i$, both functions can be written as $\frac{e^{z_i}}{e^{z_i} + \mZ_{-i}}$ where $\mZ_{-i}$ is the share of normalization factor except element $i$ of $\vz$. For $\softmaxattn$, $\mZ_{-i} = \sum_{j \neq i} e^{z_j}$ and for $\sigmoidattn$, $\mZ_{-i} = e^{-b}$. Now, we have
\begin{align}
     \frac{\partial}{\partial z_i} \frac{e^{z_i}}{e^{z_i} + \mZ_{-i}} = \frac{e^{z_i}\mZ_{-i}}{\left(e^{z_i} + \mZ_{-i}\right)^2}.
\end{align}
Therefore, we have the following
\begin{align}
    \frac{\partial p_i^\star}{\partial z_i} &= p_i^\star (1-p_i^\star)\\
    \frac{\partial p_i}{\partial z_i} &= p_i (1-p_i).
\end{align}
We can see that if $p_i \simeq p_i^\star$, then $\frac{\partial p_i}{\partial z_i} \simeq \frac{\partial p_i^\star}{\partial z_i}$. So, the previous choice of bias term $b \simeq -\log(n)$ approximates the order of gradients as well. In fact, this is the only valid choice even though we have a quadratic term.
\begin{align}
    \frac{\partial p_i}{\partial z_i} \simeq \frac{\partial p_i^\star}{\partial z_i} &\quad\Longleftrightarrow\quad p_i^\star (1-p_i^\star) = p_i (1-p_i)\\
    &\quad\Longleftrightarrow\quad \left(p_i - p_i^\star\right)\left(p_i - (1-p_i^\star)\right) = 0.
\end{align}
Which means either $p_i \simeq p_i^\star$ or $p_i \simeq 1- p_i^\star$. The first one provides us with $b \simeq -\log(n)$ while the second one cannot happen since the nominator of $p_i$ is dependent on $z_i$ while the nominator of $1-p_i^\star$ is independent of $z_i$.

\section{Details of \textsc{FlashSigmoid}}
\label{sec:DetailsOfFlashSigmoid}
\noindent This appendix provides details of the \textsc{FlashSigmoid} algorithm.
We begin by discussing the implementation details of \textsc{FlashSigmoid}, which we build as an extension of \textsc{FlashAttention2}, followed by a benchmark of the performance of the involved kernels.
We show that the kernels of \textsc{FlashSigmoid} provide a considerable performance boost in model inference over those of \textsc{FlashAttention2} and a modest performance boost for model training. 
Further, we demonstrate that the kernel speed boosts also reflect in a considerable performance gain in realistic end-to-end experiments, with an example of training vision transformers~\citep{DBLP:conf/iclr/DosovitskiyB0WZ21} on the ImageNet dataset~\citep{DBLP:conf/cvpr/DengDSLL009}. 
Finally, we also provide kernel benchmarking details of \textsc{FlashSigmoid} implementation by taking into account ALiBi slopes~\citep{DBLP:conf/iclr/PressSL22}, which is one of the important components of $\sigmoidattn$ as seen in the main text of the paper. 

\subsection{Details of \textsc{FlashSigmoid} Algorithm}
\label{sec:DetailsOfFlashSigmoidAlgorithm}
\paragraph{Softmax vs. Sigmoid Attention:} In this subsection, we discuss the implementation details of \textsc{FlashSigmoid} algorithm, which is a hardware-aware implementation of $\sigmoidattn$ approach. 
We begin with the expressions of the forward and backward passes of softmax and sigmoid attention mechanisms.
Let $\mQ$, $\mK$, and $\mV$ represent the query, key, and value tensors. 
Then, the desired forward and backward pass expressions are reported in \cref{table:ComparisonOfForwardBackward}.
\begin{table}[!h]
    \centering
    \begin{sc}
    \resizebox{\columnwidth}{!}{%
    \begin{tabular}{@{\extracolsep{4pt}}cccc}
        \toprule
            \multicolumn{2}{c}{\textsc{Softmax}}
            &
            \multicolumn{2}{c}{\textsc{Sigmoid}}
        \\
        \cmidrule{1-2} 
        \cmidrule{3-4} 
            \textsc{Forward}
            &
            \textsc{Backward}
            &
            \textsc{Forward}
            &
            \textsc{Backward}
        \\
        \toprule
            $\displaystyle \mS = \frac{\mQ\cdot\mK^\top}{\sqrt{d}}$
            &
            $\displaystyle\mathbf{d}\mV = \mP^\top \cdot \mathbf{d}\mO$ 
            &
            $\displaystyle\mS = \frac{\mQ\cdot\mK^\top}{\sqrt{d}}$
            &
            $\displaystyle\mathbf{d}\mV = \mP^\top \cdot \mathbf{d}\mO$ 
        \\
            {\textcolor{orange}{$\displaystyle\mP = \textrm{softmax}\left(\mS\right)$}}
            &
            $\displaystyle\mathbf{d}\mP = \mathbf{d}\mO\cdot \mV^\top$
            &
            {\textcolor{orange}{$\displaystyle\mP = \sigma\left(\mS\right)$}}
            &
            $\displaystyle\mathbf{d}\mP = \mathbf{d}\mO\cdot \mV^\top$
        \\
            $\displaystyle\mO = \mP\cdot \mV$
            &
            {\textcolor{purple}{$\displaystyle\mathbf{d}\mS = \mP\odot \left(\mathbf{d}\mP - \textrm{rowsum}\left(\mathbf{d}\mO\odot\mO\right)\right)$}}
            &
            $\mO = \mP\cdot \mV$
            &
            {\textcolor{purple}{$\displaystyle\mathbf{d}\mS = \mP\odot \left(1 - \mP\right)\odot \mathbf{d}\mP$}}
        \\
            {}
            &
            $\displaystyle\mathbf{d}\mQ = \sqrt{d}\cdot \mathbf{d}\mS\cdot \mK$
            &
            {}
            &
            $\displaystyle\mathbf{d}\mQ = \sqrt{d}\cdot \mathbf{d}\mS\cdot \mK$
        \\
            {}
            &
            $\displaystyle\mathbf{d}\mK = \sqrt{d}\cdot \mathbf{d}\mS^\top\cdot \mQ$
            &
            {}
            &
            $\displaystyle\mathbf{d}\mK = \sqrt{d}\cdot \mathbf{d}\mS^\top\cdot \mQ$
        \\
        \bottomrule
        \\
    \end{tabular}
    }
    \end{sc}
    \caption{
        Description of the forward and backward passes of softmax and sigmoid attention. With $\odot$, we denote Hadamard (element-wise) multiplication.
    }
    \label{table:ComparisonOfForwardBackward}
\end{table}
\begin{algorithm}[!h]
    \caption{\textsc{FlashSigmoid} Forward Pass}
    \label{alg:FlashSigmoidForward}
    \begin{algorithmic}[1]
    \Procedure{Forward}{
        $\mQ, \mK, \mV, B_r, B_c$
    }:
        \State {\textcolor{gray}{"""}}
        \State {\textcolor{gray}{\textbf{inputs:}\ Matrices $\mQ, \mK, \mV\in\mathbb{R}^{n\times d}$ are on HBM of the GPU.}}
        \State {\textcolor{gray}{\textbf{inputs:}\ Integers $B_r$ and $B_c$ are the block size for queries and key-values respectively.}}
        \State
        \State {\textcolor{gray}{\textbf{outputs:}\ Matrix $\mO\in\mathbb{R}^{n\times d}$ on HBM of the GPU.}}
        \State \qquad {\textcolor{blue}{\#\ No need to output logsumexp vector $\mL\in\mathbb{R}^n$ on HBM.}}
        \State """
        \State Divide $\mQ$ into $T_r := \ceil{\frac{n}{B_r}}$ blocks: $\mQ_1, \cdots, \mQ_{T_r}$ with $\mQ_i\in\mathbb{R}^{B_r\times d}$. 
        \State Divide $\mK$ into $T_c := \ceil{\frac{n}{B_c}}$ blocks: $\mK_1, \cdots, \mK_{T_c}$ with $\mK_i\in\mathbb{R}^{B_c\times d}$. 
        \State Divide $\mV$ into $T_c$ blocks: $\mV_1, \cdots, \mV_{T_c}$ with $\mV_i\in\mathbb{R}^{B_c\times d}$. 
        \State Divide $\mO$ into $T_r$ blocks: $\mO_1, \cdots, \mO_{T_r}$ with $\mO_i\in\mathbb{R}^{B_r\times d}$. 
        \For {$i = 1, \cdots, T_r$}
            \State Load block $\mQ_i$ from HBM to SRAM of the GPU.
            \State On chip, initialize $\mO_i$ with zeros: $\mO_i \leftarrow \mathbf{0}^{B_r\times d}$.
            \State\qquad{} {\textcolor{blue}{\#\ No allocation of either row-sum $\ell_i\in\mathbb{R}^{B_r}$ or row-max $m_i\in\mathbb{R}^{B_r}$ on chip.}}
            \For {$j = 1\cdots T_c$}
                \State Load blocks $\mK_j, \mV_j$ from HBM to SRAM of the GPU.
                \State On chip, evaluate pre-activations: $\mS_{ij} \leftarrow \mQ_i\cdot \mK_j^\top / \sqrt{d}\in \mathbb{R}^{B_r\times B_c}$.
                \State {\textcolor{orange}{On chip, evaluate sigmoid attention: $\mP_{ij} \leftarrow \sigma\left(\mS_{ij}\right)$}}.
                \State On chip, update output block: $\mO_i \leftarrow \mO_i + \mP_{ij}\cdot \mV_j$.
                \State \qquad{} {\textcolor{blue}{\#\ No need to update and track $\ell_i$ and $m_i$ vectors.}}
            \EndFor
            \State Store $\mO_i$ from chip to HBM as the $i-$th block of $\mO$ matrix. 
            \State \qquad{} {\textcolor{blue}{\#\ No post-processing of $\mO_i$ or $\mL_i$ blocks on chip.}}
            \State \qquad{} {\textcolor{blue}{\#\ No movement of $\mL_i$ block from chip to HBM.}}
        \EndFor
        \State \textbf{return} matrix $\mO$. 
    \EndProcedure
    \end{algorithmic}
\end{algorithm}
\begin{algorithm}[!h]
    \caption{\textsc{FlashSigmoid} Backward Pass}
    \label{alg:FlashSigmoidBackward}
    \begin{algorithmic}[1]
    \Procedure{Backward}{
        $\mQ, \mK, \mV, \mathbf{d}\mO, B_r, B_c$
    }:
        \State {\textcolor{gray}{"""}}
        \State {\textcolor{gray}{\textbf{inputs:}\ Matrices $\mQ, \mK, \mV, \mathbf{d}\mO\in\mathbb{R}^{n\times d}$ are on HBM of the GPU.}}
        \State {\textcolor{gray}{\textbf{inputs:}\ Integers $B_r$ and $B_c$ are the block size for queries and key-values respectively.}}
        \State \qquad{} {\textcolor{blue}{\#\ No need of logsumexp vector $\mL\in\mathbb{R}^n$ to be saved for the backward pass.}}
        \State
        \State {\textcolor{gray}{\textbf{outputs:}\ Matrices $\mathbf{d}\mQ, \mathbf{d}\mK, \mathbf{d}\mV\in\mathbb{R}^{n\times d}$ on HBM of the GPU.}}
        \State """
        \State Divide $\mQ$ into $T_r := \ceil{\frac{n}{B_r}}$ blocks: $\mQ_1, \cdots, \mQ_{T_r}$ with $\mQ_i\in\mathbb{R}^{B_r\times d}$. 
        \State Divide $\mK$ into $T_c := \ceil{\frac{n}{B_c}}$ blocks: $\mK_1, \cdots, \mK_{T_c}$ with $\mK_i\in\mathbb{R}^{B_c\times d}$. 
        \State Divide $\mV$ into $T_c$ blocks: $\mV_1, \cdots, \mV_{T_c}$ with $\mV_i\in\mathbb{R}^{B_c\times d}$. 
        \State Divide $\mO$ into $T_r$ blocks: $\mO_1, \cdots, \mO_{T_r}$ with $\mV_i\in\mathbb{R}^{B_r\times d}$. 
        \State Divide $\mathbf{d}\mO$ into $T_r := \ceil{\frac{n}{B_r}}$ blocks: $\mathbf{d}\mO_1, \cdots, \mathbf{d}\mO_{T_r}$ with $\mathbf{d}\mO_i\in\mathbb{R}^{B_r\times d}$. 
        \State Allocate $\mathbf{d}\mQ$ on HBM and divide into $T_r$ blocks: $\mathbf{d}\mQ_1, \cdots, \mathbf{d}\mQ_{T_r}$ with $\mathbf{d}\mQ_i\in\mathbb{R}^{B_r\times d}$. 
        \State Allocate $\mathbf{d}\mK$ on HBM and divide into $T_c$ blocks: $\mathbf{d}\mK_1, \cdots, \mathbf{d}\mK_{T_c}$ with $\mathbf{d}\mK_i\in\mathbb{R}^{B_c\times d}$. 
        \State Allocate $\mathbf{d}\mV$ on HBM and divide into $T_c$ blocks: $\mathbf{d}\mV_1, \cdots, \mathbf{d}\mV_{T_c}$ with $\mathbf{d}\mV_i\in\mathbb{R}^{B_c\times d}$. 
        \State {\textcolor{blue}{\qquad{}  \#\ No need to compute $\textrm{rowsum}\left(\mathbf{d}\mO\odot \mO\right)$ as sigmoid and its gradients are pointwise.}}
        \For {$j = 1, \cdots, T_c$}
            \State Load blocks $\mK_j, \mV_j$ from HBM to SRAM of the GPU.
            \State On chip, initialize $\mathbf{d}\mK_j, \mathbf{d}\mV_j$ with zeros: $\mathbf{d}\mK_j \leftarrow \mathbf{0}^{B_c\times d}; \mathbf{d}\mV_j \leftarrow \mathbf{0}^{B_c\times d}$.
            \For {$i = 1\cdots T_r$}
                \State Load blocks $\mQ_i, \mathbf{d}\mO_i, \mathbf{d}\mQ_i$ from HBM to SRAM of the GPU.
                \State \qquad{} {\textcolor{blue}{\#\ No need of movement of blocks $\textrm{rowsum}\left(\mathbf{d}\mO\odot \mO\right)_i$ and logsumexp $\mL_i$.}}
                \State On chip, evaluate pre-activations: $\mS_{ij} \leftarrow \mQ_i\cdot \mK_j^\top / \sqrt{d}\in \mathbb{R}^{B_r\times B_c}$.
                \State On chip, evaluate sigmoid attention: $\mP_{ij} \leftarrow \sigma\left(\mS_{ij}\right)$.
                \State On chip, update gradient of values: $\mathbf{d}\mV_i \leftarrow \mathbf{d}\mV_i + \mP_{ij}^\top\cdot \mathbf{d}\mO_j$.
                \State On chip, compute gradients of attention matrix: $\mathbf{d}\mP_{ij} \leftarrow \mathbf{d}\mO_i\cdot \mV_i^\top\in\mathbb{R}^{B_r\times B_c}$.
                \State {\textcolor{purple}{On chip, compute gradients of pre-activations: $\mathbf{d}\mS_{ij} \leftarrow \mP_{ij}\odot\left(1 - \mP_{ij}\right)\odot \mathbf{d}\mP_{ij}$}}.
                \State Load query gradient block $\mathbf{d}\mQ_i$ from HBM to SRAM, and then on to chip.
                \State Update query gradient block on chip: $\mathbf{d}\mQ_i \leftarrow \mathbf{d}\mQ_i + \sqrt{d}\cdot \mathbf{d}\mS_{ij}\cdot \mK_j$.
                \State Store query gradient block $\mathbf{d}\mQ_i$ from chip back to HBM.
                \State On chip, update key gradient block: $\mathbf{d}\mK_j \leftarrow \mathbf{d}\mK_j + \sqrt{d}\cdot \mathbf{d}\mS_{ij}^\top\cdot \mQ_i$. 
            \EndFor
            \State Store $\mathbf{d}\mK_j, \mathbf{d}\mV_j$ from chip to HBM as the $j-$th blocks of $\mathbf{d}\mK, \mathbf{d}\mV$ matrices respectively. 
        \EndFor
        \State \textbf{return} matrices $\mathbf{d}\mQ, \mathbf{d}\mK, \mathbf{d}\mV$. 
    \EndProcedure
    \end{algorithmic}
\end{algorithm}
\noindent The application of sigmoid and softmax activation functions, as highlighted in {\textcolor{orange}{orange color}} in \cref{table:ComparisonOfForwardBackward}, is the only implementation difference in the forward passes.
Similarly, the expressions for the gradients of the preactivation ($\mathbf{d}\mS$), as highlighted in {\textcolor{purple}{purple color}} in the table above, is the only implementation difference in the backward passes. 
In light of this, we implement the \textsc{FlashSigmoid} algorithm as an extension of the \textsc{FlashAttention2}~\citep{DBLP:journals/corr/abs-2307-08691} algorithm, which is a highly optimized hardware-aware implementation of $\softmaxattn$. 
\paragraph{Flash Attention in Brief:}\ As pointed at in the main text, the \textsc{FlashAttention}~\citep{DBLP:conf/nips/DaoFERR22} and \textsc{FlashAttention2}~\citep{DBLP:journals/corr/abs-2307-08691} algorithms provide hardware-aware implementations of exact attention mechanism by optimizing for bottlenecks of modern accelerators~\citep{DBLP:journals/micro/ChoquetteGGSK21,DBLP:journals/micro/Choquette23}. 
These GPUs possess massive amounts (e.g., $\sim80$GB) of High-Bandwidth Memory (HBM), which stores large tensors but is slow in moving the data to the accelerators.
On the other hand, they have smaller amounts (e.g., $\sim 20$ MB) of SRAM, which is often more than an order magnitude faster for carrying out actual computations using the registers/tensor cores of the GPU.
This trade-off between memory size and computation speed across hierarchies results in the attention mechanism computation being bottlenecked by memory accesses between the HBM and the SRAM~\citep{DBLP:conf/mlsys/IvanovDB0H21}.
Consequently, flash algorithms optimize for memory accesses across the hierarchy of GPU memory types in order to accelerate computation of attention mechanism and its gradients. \textsc{FlashSigmoid} is no exception to this approach.

\Cref{alg:FlashSigmoidForward} describes the forward pass and \cref{alg:FlashSigmoidBackward} describes the backward pass of the \textsc{FlashSigmoid} algorithm. 
We highlight in {\textcolor{orange}{orange color}} the steps in the forward pass of \textsc{FlashSigmoid} that differ from those in \textsc{FlashAttention2} by virtue of sigmoid activation. Similarly, we highlight in {\textcolor{purple}{purple color}} the differences in the backward pass. Finally, we highlight in {\textcolor{blue}{blue color}} the salient points of \textsc{FlashSigmoid} that further help minimize bottlenecking factors on modern accelerators.
\paragraph{Fewer Tensor Allocations, Fewer Memory Accesses, Fast-\textrm{Tanh}:}\ 
In \textsc{FlashAttention} and \textsc{FlashAttention2}, the attention mechanism is computed by splitting the attention matrix into blocks. 
Since softmax activation requires a row-wise reduction to compute its normalization factor (i.e., the denominator), one needs to properly compute and track such factor across blocks. Moreover, in \textsc{FlashAttention} this normalization factor is stored after being computed in the forward pass, to have it easily accessible to further speed-up the backward pass.
By contrast, substituting sigmoid to softmax eliminates the need to allocate and move across the GPU memory hierarchy the tensors related to the normalization factor (i.e., moving the logsumexp tensor $\mL\in\mathbb{R}^n$ on HBM in the forward and backward passes).
In addition, applying softmax in a stable manner requires tracking the row-max variable $m_i$ on chip, which instead is not needed for sigmoid activation.
This further helps reducing some on-chip operations and lowering register pressure in \textsc{FlashSigmoid}. 

Moving on to the backward pass~(described in \cref{alg:FlashSigmoidBackward}), \textsc{FlashAttention2} requires computing $\textrm{rowsum}\left(\mathbf{d}\mO\odot \mO\right)$, which is needed to backpropagate the gradients of softmax attention outputs to the preactivations.
However, since sigmoid activation is applied element-wise, its gradients also backpropagate across sigmoid element-wise, eliminating the need of the row-sum variable and the movement of its blocks across the memory hierarchy. 
Another optimization of \textsc{FlashAttention} and \textsc{FlashAttention2} consists of partially re-computing the forward pass of attention mechanism in the backward pass to avoid bottlenecks and speed-up the implementation.
To keep the backward pass implementation fast, they require the logsumexp variable to be available and transferred between HBM and SRAM in the backward pass.
\textsc{FlashSigmoid}, being an element-wise activation, eliminates the need of this variable from the backward pass, and consequently, from the entire algorithm. 
Finally, a major component in our implementation is the usage of GPU-based implementation of the \textrm{tanh} activation. 
Sigmoid activation is related to Tanh activation via the following relation: $\sigma\left(x\right) = 0.5\cdot \left(1 + \tanh\left(0.5\cdot x\right)\right)$. 
We utilize the fast GPU-implementation of Tanh activation, which trades off some precision for better speed, in order to compute sigmoid activation in both the forward and the backward pass. 
This provides a considerable speed-boost in both the forward and backward passes of \textsc{FlashSigmoid}, while maintaining parity in performance with a na\"ive implementation of sigmoid attention. 
Based on these points of modification, we extend \textsc{FlashAttention2} to obtain \textsc{FlashSigmoid}, a hardware-aware implementation of $\sigmoidattn$. 
\subsection{Benchmarking of \textsc{FlashSigmoid} Kernels}
\label{sec:PerformanceAnalysisOfFlashSigmoidKernels}
\paragraph{Benchmarking Setup:} Having seen the details of the \textsc{FlashSigmoid} algorithm, we next consider the benchmarking of its kernels.
For this, we create a small model in PyTorch~\citep{DBLP:conf/nips/PaszkeGMLBCKLGA19} that inputs query, key, and value tensors (all of shape $[\textrm{batch}, \textrm{tokens}, \textrm{heads}, \textrm{features}]$) and passes these through a number of attention layers. 
Mimicking the design of vision transformers (ViTB-16/224)~\citep{DBLP:conf/iclr/DosovitskiyB0WZ21}, we set the number of heads and per-head features as $12$ and $64$, respectively.
We set a batch size of $32$, and consider a $10$-layer architecture.
Then, for the number of tokens sampled from a wide range of $[64, 78\mathrm{k}]$, we compute the forward and backward passes of this model.
For these computations, we measure the kernel GPU time using PyTorch's profiler. 
We carry out our experiments on both H100~\citep{DBLP:journals/micro/Choquette23} and A100~\citep{DBLP:journals/micro/ChoquetteGGSK21} GPUs. 
\begin{figure}[!htbp]
    \centering
    \begin{minipage}{0.46\textwidth}
        \footnotesize
        \centering
        \includegraphics[trim={0 0 0 0}, width=\textwidth]{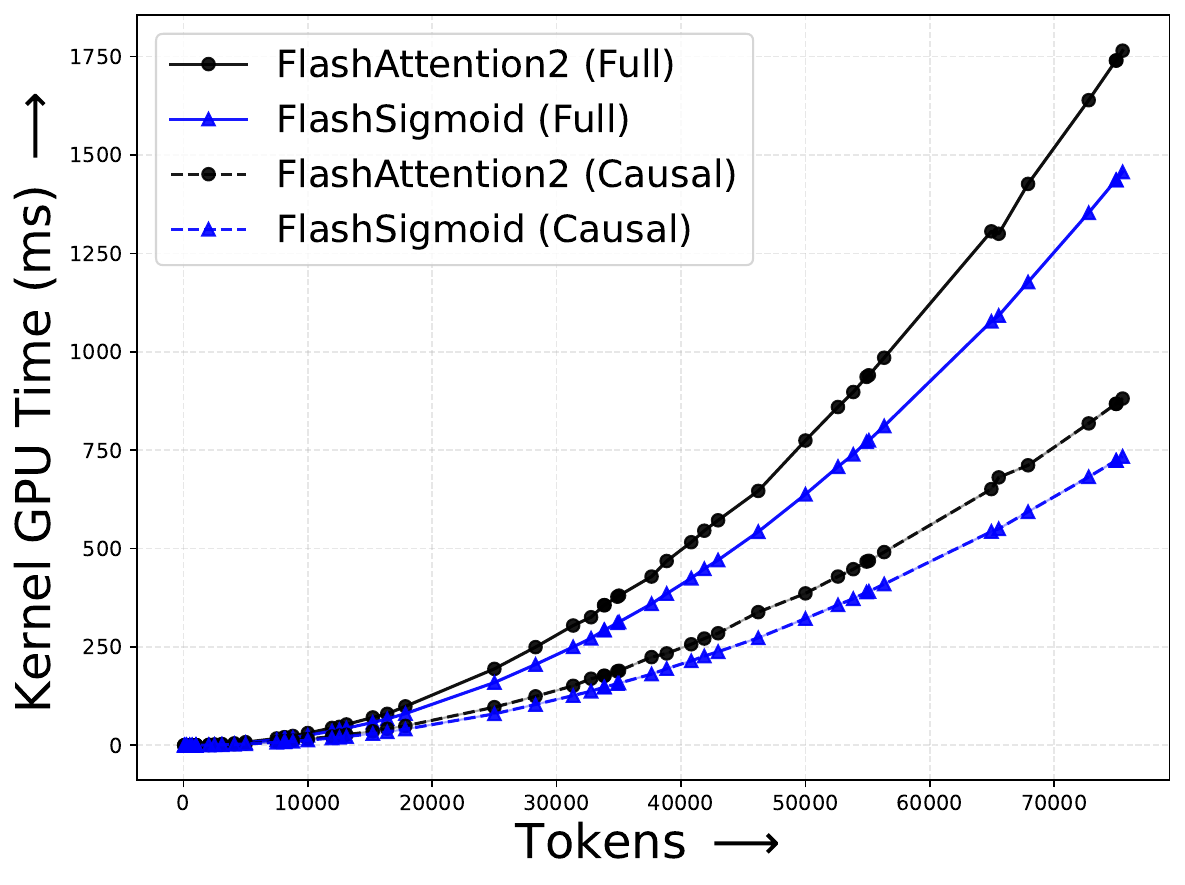}
        \captionsetup{justification=centering}
        \caption*{
            (a) Inference mode kernels on H100.
        }
    \end{minipage}
    \hfill
    \begin{minipage}{0.46\textwidth}
        \centering        
        \includegraphics[trim={0 0 0 0}, width=\textwidth]{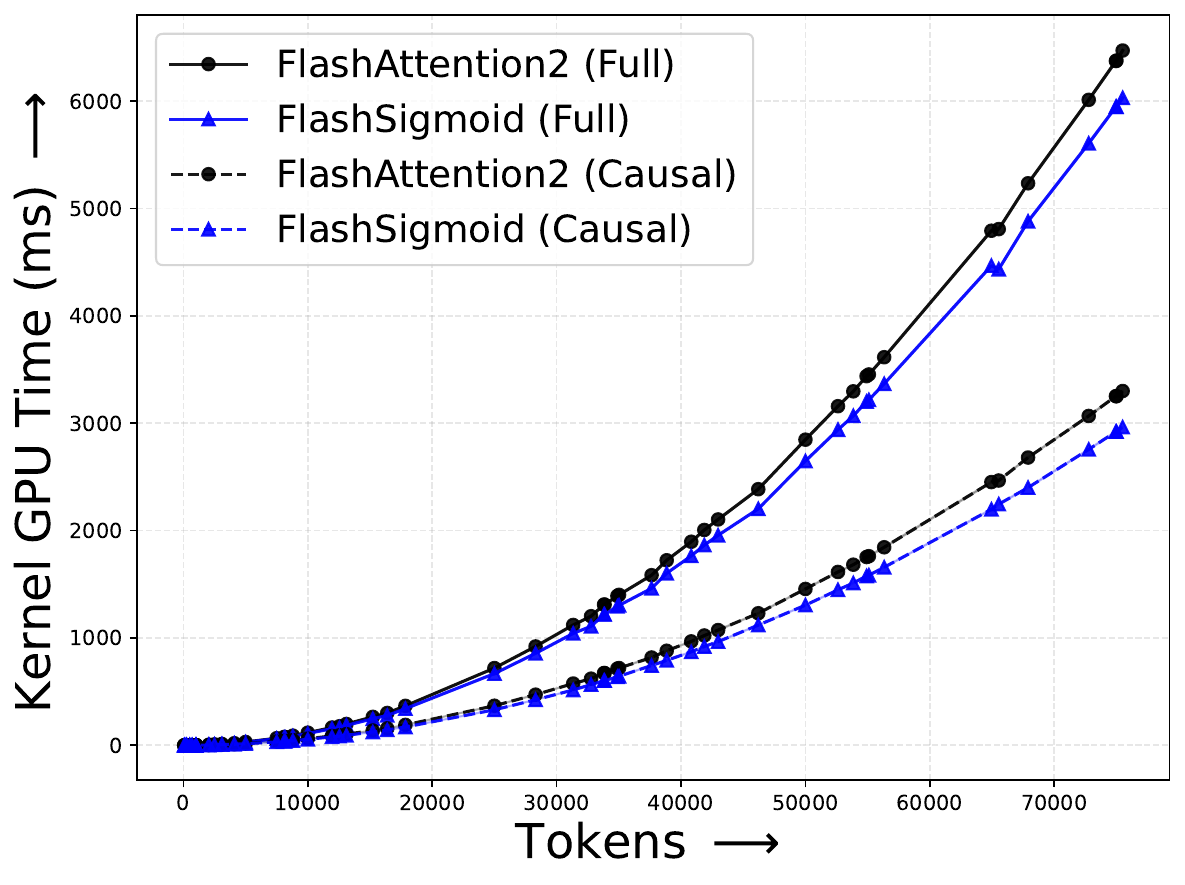}
        \captionsetup{justification=centering} 
        \caption*{
            (b) Training mode kernels on H100.
        }
    \end{minipage}
    \caption{
        On average, for sequence lengths between $[64, 78\mathrm{k}]$, the inference mode kernel of \textsc{FlashSigmoid} is ${17.39}\%$ faster than \textsc{FlashAttention2} for self-attention and ${18.76}\%$ for causal attention.
        The training mode kernels of \textsc{FlashSigmoid} are ${6.53}\%$ faster than \textsc{FlashAttention2} for self-attention and ${9.46}\%$ for causal attention.
        Note that inference involves only the forward pass of the model and training involves both the forward and the backward pass of the model. 
    }
    \label{fig:h100-softmax-sigmoid-fwd-bwd}
\end{figure}
\begin{figure}[!htbp]
    \centering
    \begin{minipage}{0.46\textwidth}
        \footnotesize
        \centering
        \includegraphics[trim={0 0 0 0}, width=\textwidth]{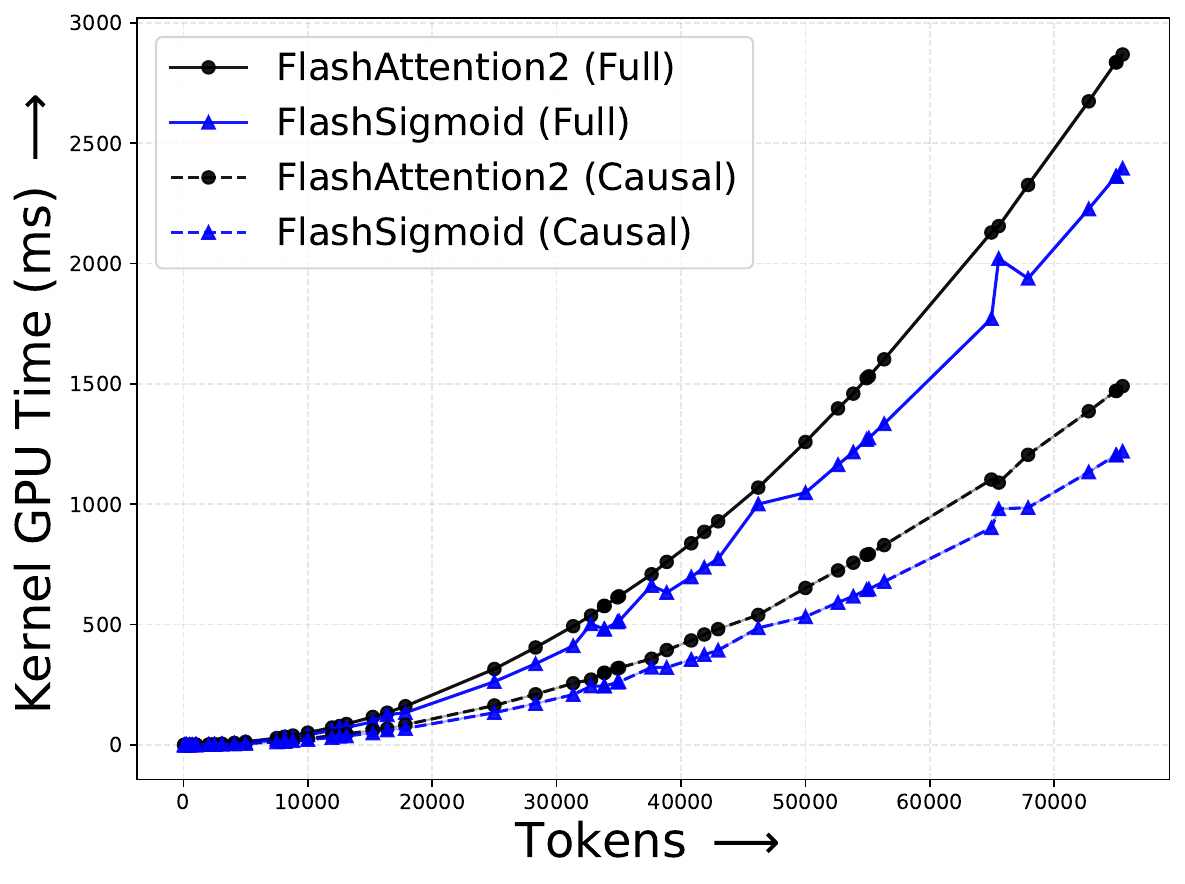}
        \captionsetup{justification=centering}
        \caption*{
            (a) Inference mode kernels on A100.
        }
    \end{minipage}
    \hfill
    \begin{minipage}{0.46\textwidth}
        \centering        
        \includegraphics[trim={0 0 0 0}, width=\textwidth]{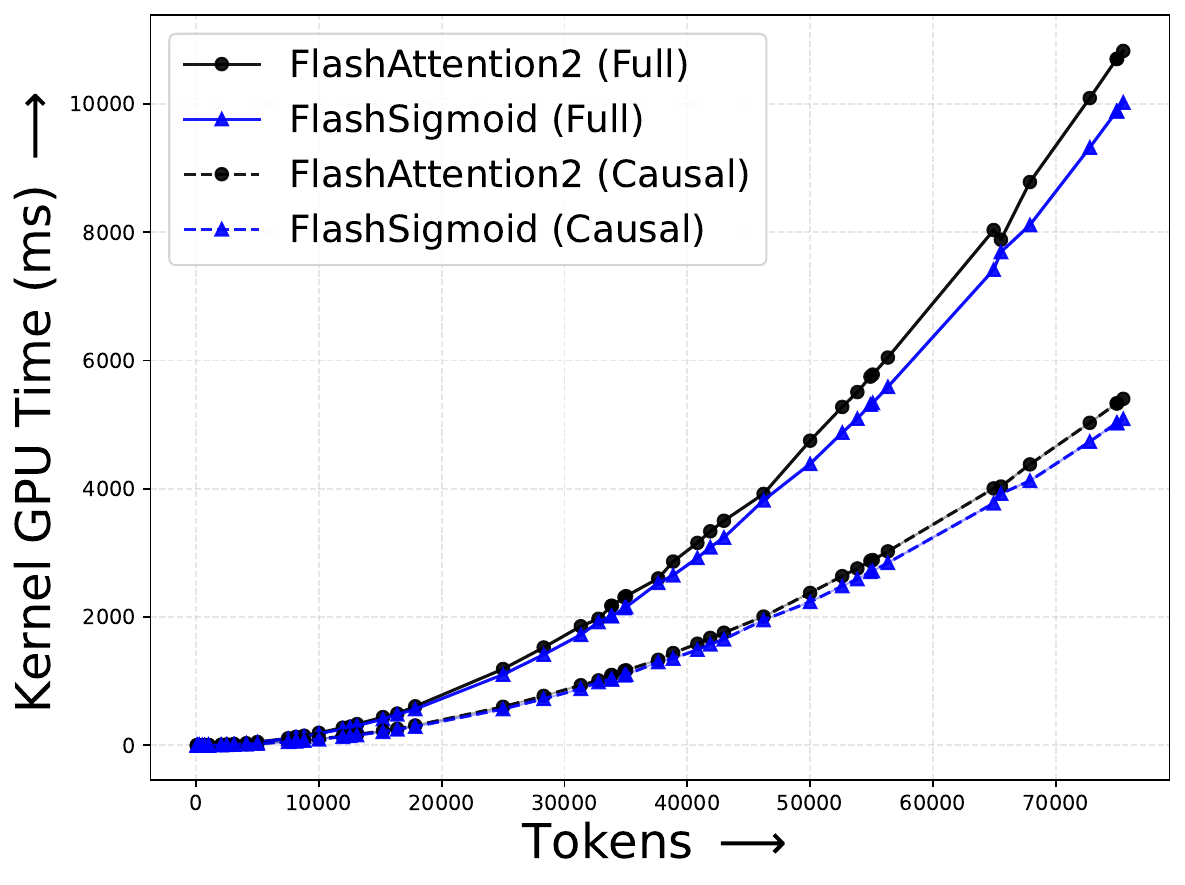}
        \captionsetup{justification=centering} 
        \caption*{
            (b) Training mode kernels on A100.
        }
    \end{minipage}
    \caption{
        On average, for sequence lengths between $[64, 78\mathrm{k}]$, the inference mode kernel of \textsc{FlashSigmoid} is ${14.33}\%$ faster than \textsc{FlashAttention2} for self-attention and ${16.92}\%$ for causal attention.
        The training mode kernels of \textsc{FlashSigmoid} are ${6.02}\%$ faster than \textsc{FlashAttention2} for self-attention and ${5.27}\%$ for causal attention.
        Note that inference involves only the forward pass of the model and training involves both the forward and the backward pass of the model. 
    }
    \label{fig:a100-softmax-sigmoid-fwd-bwd}
\end{figure}
\vspace{-0.1in}
\paragraph{Results:} \Cref{fig:h100-softmax-sigmoid-fwd-bwd,fig:a100-softmax-sigmoid-fwd-bwd} show the GPU time comparisons of kernels in inference mode and training mode of \textsc{FlashSigmoid} and \textsc{FlashAttention2} respectively.
We observe that we obtain a large average speed-boost for inference and a modest average speed-boost for training. 
Note that the speed-ups in all the subsequent figures are obtained by averaging the performances for tokens sampled in the range of $[64, 78\mathrm{k}]$.
\paragraph{Details of Individual Kernels:} Next, we also show the performance of individual flash kernels of \textsc{FlashSigmoid} and \textsc{FlashAttention2}.
Note that inference mode involves only the forward pas of the model, while training mode involves both the forward and the backward pass of the model.
The forward pass of both these approaches involves one kernel, which we term \textrm{flash\char`_fwd\char`_kernel}, and the backward pass of both these approaches is made up of three kernels, which we term \textrm{bwd\char`_dq\char`_dk\char`_dv}, \textrm{bwd\char`_dot\char`_do\char`_o}, and \textrm{bwd\char`_convert\char`_dq}.
In code, the real names of these kernels are as follows.
\begin{equation}
    \label{eq:KernelRealNames}
    \begin{aligned}
    \textrm{fwd}
    &
    := 
    \textrm{flash\char`_fwd\char`_kernel}
    \\
    \textrm{bwd\char`_dq\char`_dk\char`_dv}
    &
    :=    \textrm{flash\char`_bwd\char`_dq\char`_dk\char`_dv\char`_loop\char`_seqk\char`_parallel\char`_kernel}
    \\
    \textrm{bwd\char`_dot\char`_do\char`_o}
    &
    :=
    \textrm{flash\char`_bwd\char`_dot\char`_do\char`_o\char`_kernel}
    \\
    \textrm{bwd\char`_convert\char`_dq}
    &
    :=
    \textrm{flash\char`_bwd\char`_convert\char`_dq\char`_kernel}
    \end{aligned}    
\end{equation} 
\noindent Here, we first provide a brief description of the tasks performed by each of these kernels; for a detailed explanation, we refer the reader to \textsc{FlashAttention2}~\citep{DBLP:journals/corr/abs-2307-08691} paper and code.
The \textrm{fwd} kernel computes the full forward pass of the model as shown in \cref{table:ComparisonOfForwardBackward}. 
The bulk of computations of the backward pass happen in the \textrm{bwd\char`_dq\char`_dk\char`_dv} kernel, which performs re-computation of attention matrix and reduction of key and value gradient tensors ($\mathbf{d}\mK$, $\mathbf{d}\mV$). 
Again, the exact steps carried out in the backward pass can be checked from \cref{table:ComparisonOfForwardBackward}. 
The \textrm{bwd\char`_convert\char`_dq} kernel performs the reduction of query gradient tensor ($\mathbf{d}\mQ$).
Finally, note that the \textrm{bwd\char`_dot\char`_do\char`_o} kernel in \textsc{FlashAttention2} performs the task of computing the $\textrm{rowsum}(\mathbf{d}\mO\odot \mO)$ tensor along with clearing of the accumulators of query gradients ($\mathbf{d}\mQ$).
Although \textsc{FlashSigmoid} does not require this row-sum tensor, the clearing of accumulators of query gradients is still needed. 
For this reason, \textrm{bwd\char`_dot\char`_do\char`_o} kernel also appears in the profiling of \textsc{FlashSigmoid}.
\paragraph{Performance of Individual Kernels:} ~\Cref{fig:h100-softmax-sigmoid-kernels,fig:a100-softmax-sigmoid-kernels} show the performance comparison of each flash kernel in \textsc{FlashSigmoid} with the corresponding kernel in \textsc{FlashAttention2} when tested on an H100 GPU and an A100 GPU respectively.
We observe that on both the H100 and A100 GPU architectures, the \textrm{fwd} kernel of \textsc{FlashSigmoid} is significantly faster than that of \textsc{FlashAttention2} and the \textrm{bwd\char`_dq\char`_dk\char`_dv} kernel of \textsc{FlashSigmoid} has a modest average speed boost over \textsc{FlashAttention2}.
The \textrm{bwd\char`_dot\char`_do\char`_o} kernel in \textsc{FlashSigmoid} is significantly faster on A100 GPUs.
Note that even though the \textrm{bwd\char`_dot\char`_do\char`_o} kernel of \textsc{FlashSigmoid} appears to be slower on average on H100 GPUs, the kernel time of \textrm{bwd\char`_dot\char`_do\char`_o} ($\sim 5$ms) is negligible compared to that of the main \textrm{bwd\char`_dq\char`_dk\char`_dv} kernel ($\sim 5000$ms). 
Thus, the combined backward pass kernel in \textsc{FlashSigmoid} time does not suffer from this slowdown. 
Finally, note that for \textrm{bwd\char`_convert\char`_dq}, \textsc{FlashSigmoid} and \textsc{FlashAttention2} have identical performance.
This is expected, since the task of this kernel is to reduce the gradient of the queries $\mathbf{d}\mQ$, which is a common step in both the approaches and is not modified in \textsc{FlashSigmoid}. 
\begin{figure}[!htbp]
    \centering
    \begin{minipage}{0.24\textwidth}
        \footnotesize
        \centering
        \includegraphics[trim={0 0 0 0}, width=\textwidth]{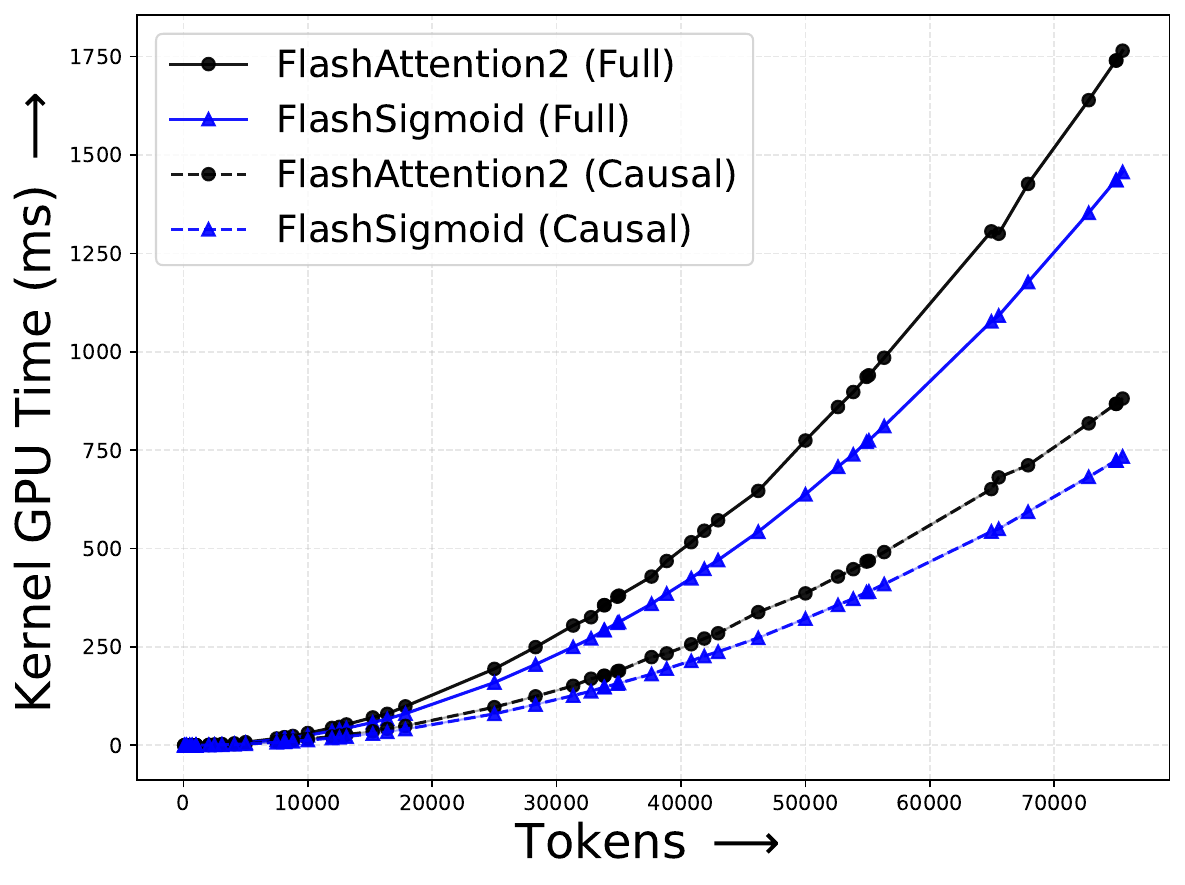}
        \captionsetup{justification=centering}
        \caption*{
            fwd:\\$17.39\%$ faster for self-attention and $18.76\%$ for causal.
        } 
    \end{minipage}
    \hfill
    \begin{minipage}{0.24\textwidth}
        \centering        
        \includegraphics[trim={0 0 0 0}, width=\textwidth]{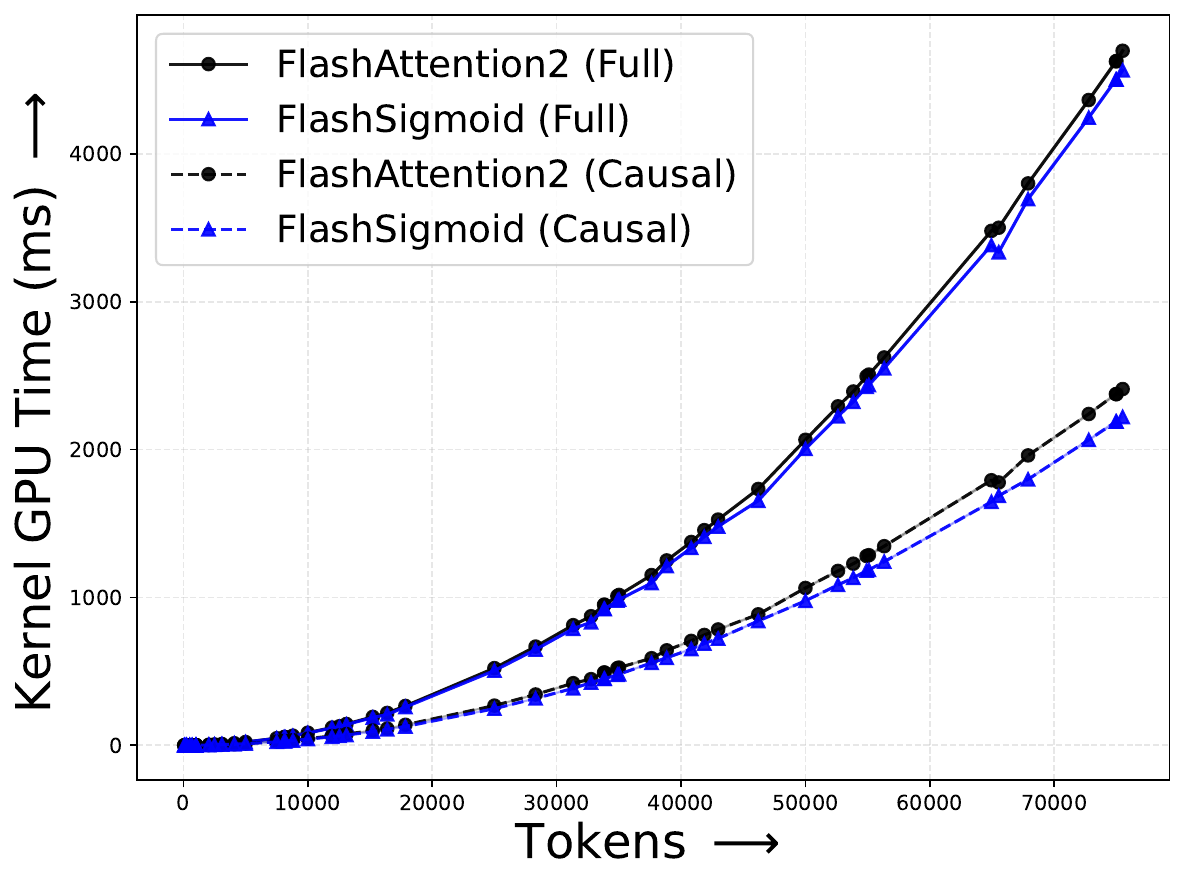}
        \captionsetup{justification=centering} 
        \caption*{
            \textrm{bwd\char`_dq\char`_dk\char`_dv}:\\$3.29\%$ faster for self-attention and $6.97\%$ for causal.
        }
    \end{minipage}
    \hfill
    \begin{minipage}{0.24\textwidth}
        \centering        
        \includegraphics[trim={0 0 0 0}, width=\textwidth]{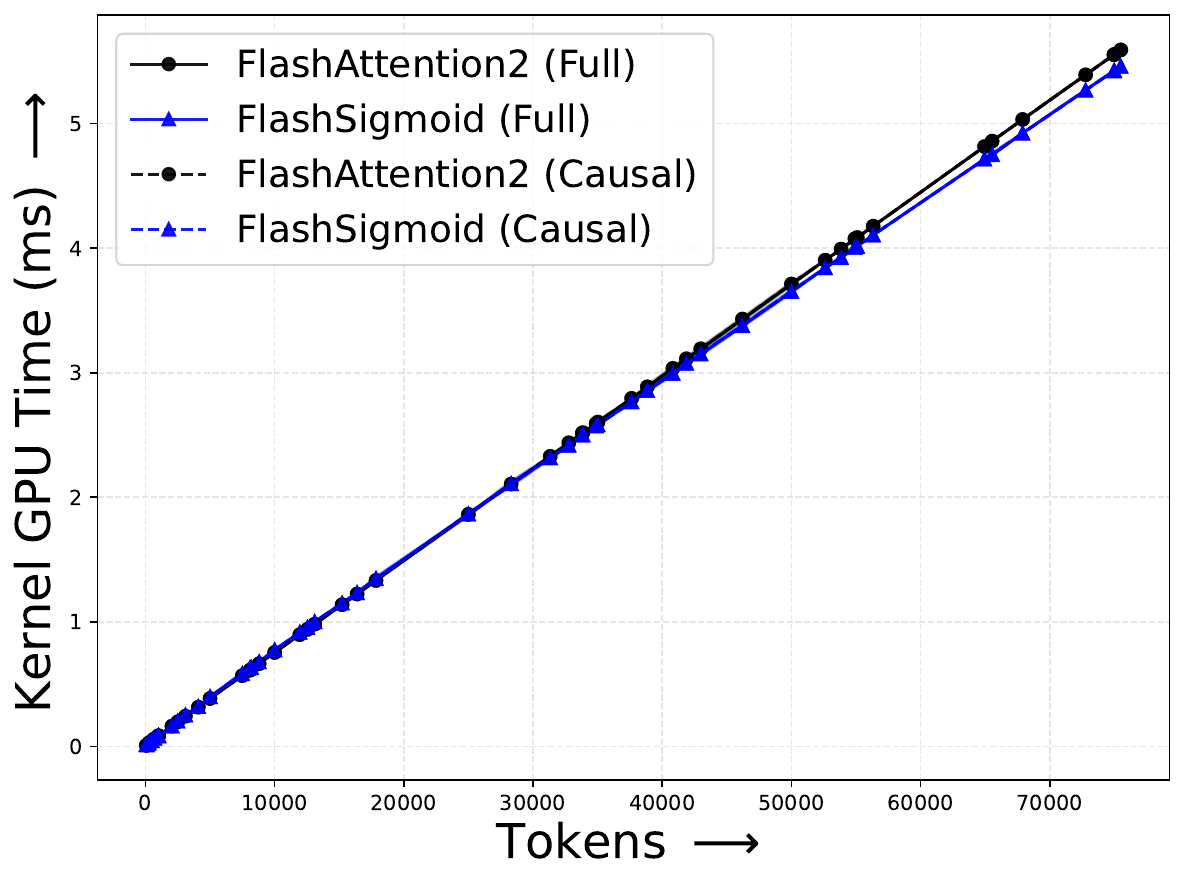}
        \captionsetup{justification=centering} 
        \caption*{
            \textrm{bwd\char`_dot\char`_do\char`_o}:\\
            $2.24\%$ slower for self-attention and $2.17\%$ for causal.  
        }
    \end{minipage}
    \hfill
    \begin{minipage}{0.24\textwidth}
        \centering        
        \includegraphics[trim={0 0 0 0}, width=\textwidth]{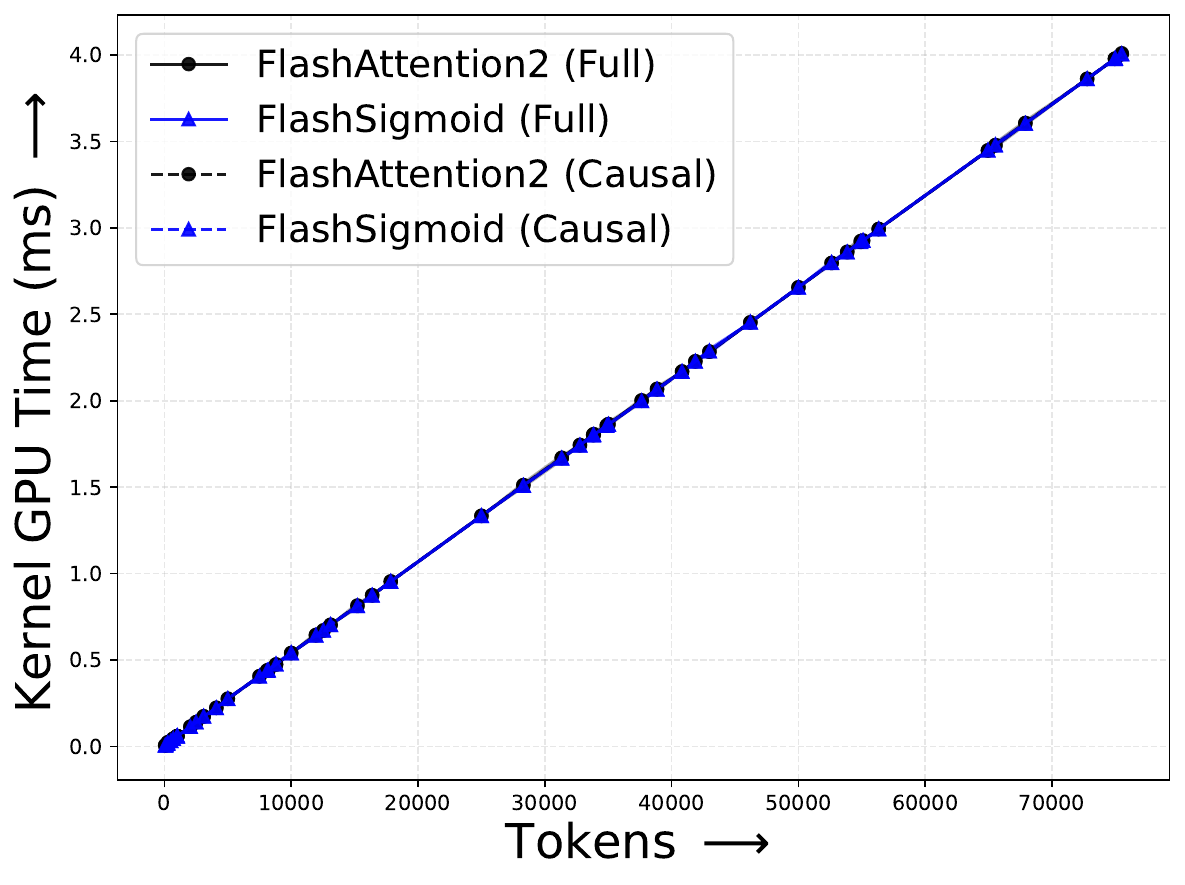}
        \captionsetup{justification=centering} 
        \caption*{
            \textrm{bwd\char`_convert\char`_dq}:\\
            $0.03\%$ faster for self-attention, $0.02\%$ slower for causal. 
        }
    \end{minipage}
    \caption{
        \textsc{FlashSigmoid} and \textsc{FlashAttention2} kernel comparison on H100 GPUs. 
    }
    \label{fig:h100-softmax-sigmoid-kernels}
\end{figure}
\begin{figure}[!htbp]
    \centering
    \begin{minipage}{0.24\textwidth}
        \footnotesize
        \centering
        \includegraphics[trim={0 0 0 0}, width=\textwidth]{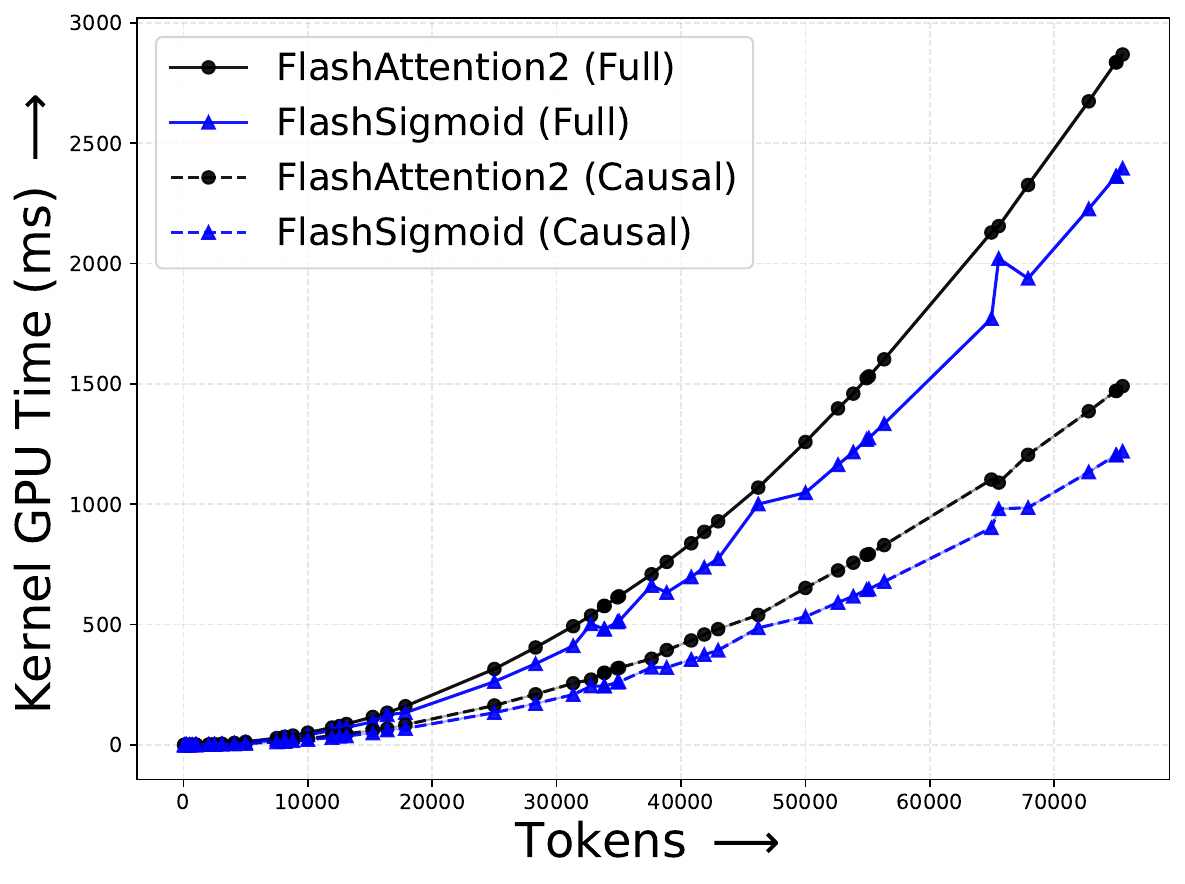}
        \captionsetup{justification=centering}
        \caption*{
            fwd:\\$14.33\%$ faster for self-attention and $16.92\%$ for causal. 
        } 
    \end{minipage}
    \hfill
    \begin{minipage}{0.24\textwidth}
        \centering        
        \includegraphics[trim={0 0 0 0}, width=\textwidth]{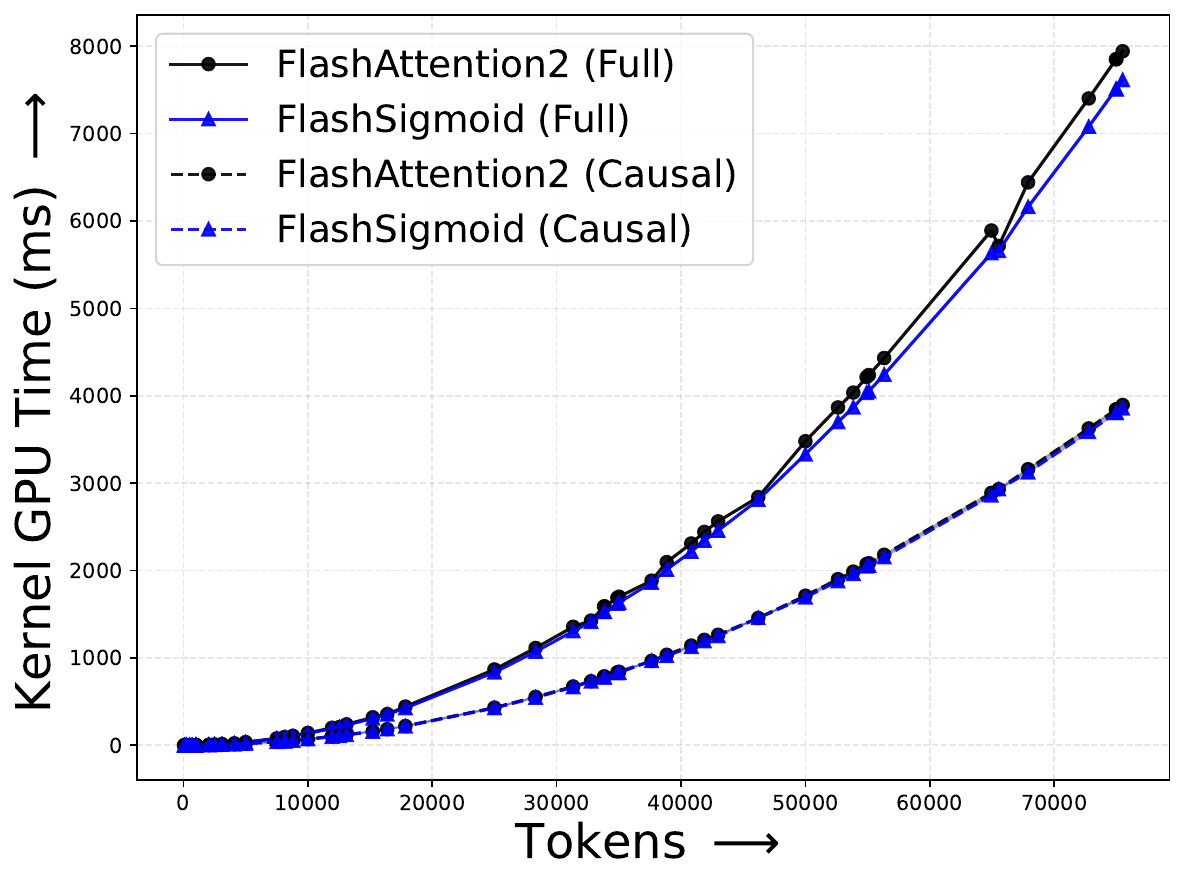}
        \captionsetup{justification=centering} 
        \caption*{
            \textrm{bwd\char`_dq\char`_dk\char`_dv}:\\$3.50\%$ faster for self-attention and $1.39\%$ for causal.
        }
    \end{minipage}
    \hfill
    \begin{minipage}{0.24\textwidth}
        \centering        
        \includegraphics[trim={0 0 0 0}, width=\textwidth]{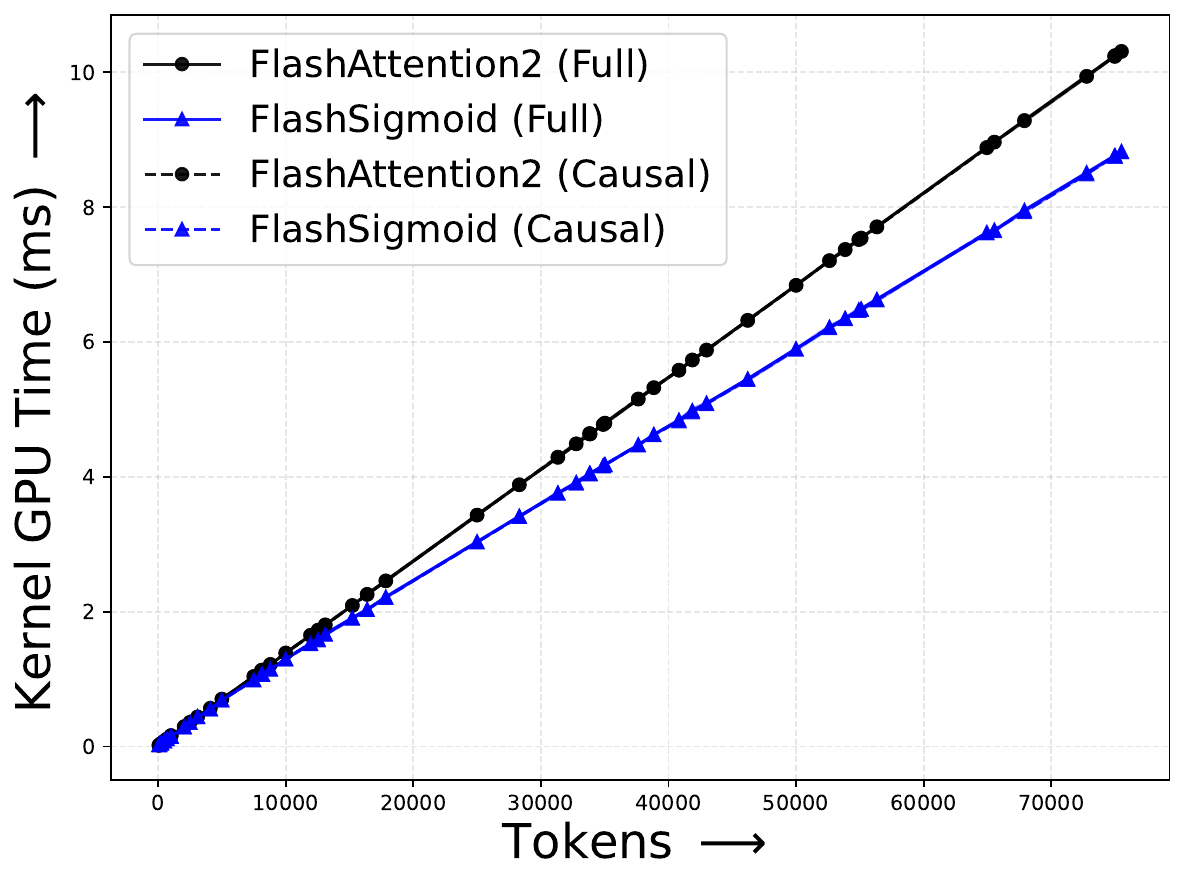}
        \captionsetup{justification=centering} 
        \caption*{
            \textrm{bwd\char`_dot\char`_do\char`_o}:\\
            $7.95\%$ faster for self-attention and $8.00\%$ for causal.  
        }
    \end{minipage}
    \hfill
    \begin{minipage}{0.24\textwidth}
        \centering        
        \includegraphics[trim={0 0 0 0}, width=\textwidth]{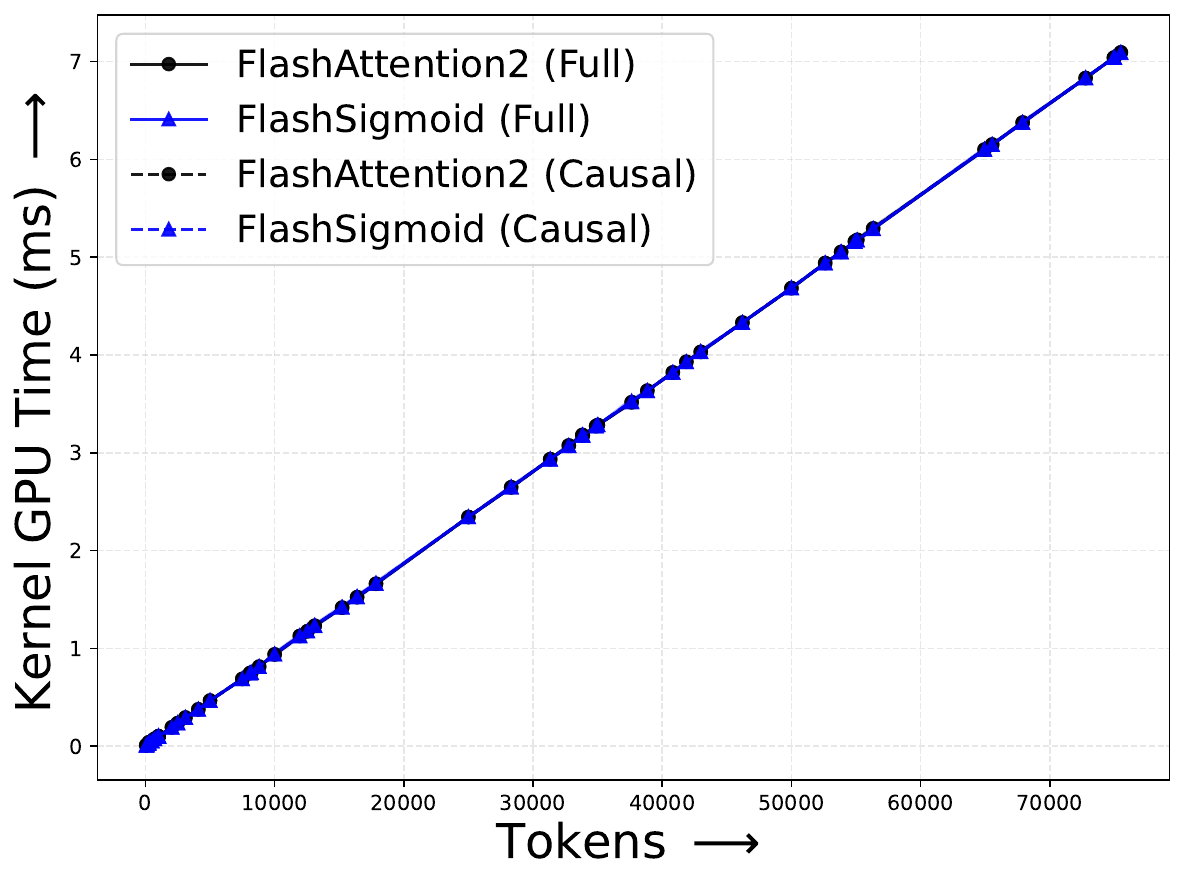}
        \captionsetup{justification=centering} 
        \caption*{
            \textrm{bwd\char`_convert\char`_dq}:\\
            $0.01\%$ faster for self-attention, $0.03\%$ slower for causal. 
        }
    \end{minipage}
    \caption{
        \textsc{FlashSigmoid} and \textsc{FlashAttention2} kernel comparison on A100 GPUs. 
    }
    \label{fig:a100-softmax-sigmoid-kernels}
\end{figure}
\subsection{Speed Boosts of \textsc{FlashSigmoid} in Realistic Settings}
\label{sec:SpeedBoostsOfFlashSigmoidInRealisticSettings}
\noindent In this section, we demonstrate how the performance boosts measured in \cref{sec:PerformanceAnalysisOfFlashSigmoidKernels} for the individual kernels of \textsc{FlashSigmoid} contributes to speeding-up realistic runs with end-to-end training. 

\noindent\textbf{Setup:}\ As a target experiment, we consider training a vision transformer~\citep{DBLP:conf/iclr/DosovitskiyB0WZ21} on the ImageNet dataset~\citep{DBLP:conf/cvpr/DengDSLL009}. 
We create two vision transformer model variants-- one with \textsc{FlashAttention2} attention and the other with \textsc{FlashSigmoid} attention. 
We carry out the training of these models with a distributed data-parallel (DDP) setup using PyTorch~\citep{DBLP:conf/nips/PaszkeGMLBCKLGA19}. 
We perform two sets of experiments-- \emph{i.} the first performs DDP training on four nodes of H100 GPUs with eight GPUs per node and EFA/RDMA interconnect for the nodes, and \emph{ii.} the second performs DDP training on four nodes of A100 GPUs with eight GPUs per node. 
In each set of experiments, we use three different image sizes ($64\times 64$, $90\times 90$, and $100\times 100$), along with patch size of $1$ to result in different number of tokens for the underlying attention mechanism in the vision transformer model ($64\times 64 = 4096$, $90\times 90 = 8100$, and $100\times 100 = 10000$ tokens).
For each of these configurations, we select batch sizes so that the GPU memory utilization would be greater than $80\%$.
These considerations are in order to minimize, if not eliminate, other confounders that can unfairly affect estimation speed-ups in realistic runs. For instance, a low GPU utilization would lead to a larger number of updates, which in turn would incur unnecessary delays, variations, and slow-downs due to across-nodes communications.

\noindent\textbf{Results:}\ The results of the runs on H100 nodes and A100 nodes are shown in \cref{fig:ComparisonsOfFlashesOnH100,fig:ComparisonsOfFlashesOnA100} respectively.
There, we show how the kernel GPU times for forward and backward passes vary according to the number of tokens considered, and include the wall-clock time of the end-to-end runs as explained above. 
We observe that the kernel speed-up reflects significantly in the speed-up of inference of the models (during testing) and modestly in the training of the models.
We observe $\sim8\%$ speed-up in wall-clock time of inference and $\sim4\%$ speed-up in wall-clock time of training. 
\begin{table}[htbp]
    \tiny
    \centering
    \begin{sc}
    \resizebox{\columnwidth}{!}{%
    \begin{tabular}{@{\extracolsep{4pt}}ccccccc}
        \toprule
            \multirow{2}{*}{Tokens}
            &
            \multicolumn{3}{c}{Kernel GPU Time Comparison}
            &
            \multicolumn{3}{c}{Full Run Wall-Clock Time Comparison} 
        \\
        \cmidrule{2-4} 
        \cmidrule{5-7} 
            & 
            Kernels
            & 
            \textsc{FlashAttention2} (ms)
            & 
            \textsc{FlashSigmoid} (ms)
            &  
            Mode
            & 
            \textsc{FlashAttention2} (s)
            & 
            \textsc{FlashSigmoid} (s) 
        \\ 
        \toprule
            \multirow{2}{*}{4096} 
            & 
            $\textrm{fwd}$ 
            & 
            $\reading{4.98}{0.01}$ 
            & 
            $\reading{4.17}{0.01}\ \mathbf{\left(-16.31\%\right)}$
            &   
            $\textrm{Inference}$
            & 
            $\reading{11.17}{0.18}$
            & 
            $\reading{10.68}{0.18}\ \mathbf{\left(-4.42\%\right)}$
        \\
        \cmidrule{2-4} 
        \cmidrule{5-7}  
            & 
            $\textrm{fwd} + \textrm{bwd}$  
            &  
            $\reading{19.58}{0.06}$
            & 
            $\reading{18.12}{0.04}\ \mathbf{\left(-7.45\%\right)}$
            &  
            $\textrm{Training}$
            & 
            $\reading{1563.39}{1.30}$
            & 
            $\reading{1521.68}{2.27}\ \mathbf{\left(-2.67\%\right)}$
        \\
        \toprule
            \multirow{2}{*}{8100} 
            & 
            $\textrm{fwd}$ 
            & 
            $\reading{20.46}{0.05}$
            & 
            $\reading{16.73}{0.05}\ \mathbf{\left(-18.22\%\right)}$
            &  
            $\textrm{Inference}$
            & 
            $\reading{28.21}{0.18}$
            & 
            $\reading{25.93}{0.17}\ \mathbf{\left(-8.06\%\right)}$
        \\
        \cmidrule{2-4} 
        \cmidrule{5-7}  
            & 
            $\textrm{fwd} + \textrm{bwd}$  
            & 
            $\reading{77.63}{0.13}$
            & 
            $\reading{72.70}{0.12}\ \mathbf{\left(-6.35\%\right)}$
            & 
            $\textrm{Training}$ 
            & 
            $\reading{4282.75}{2.14}$
            & 
            $\reading{4129.25}{4.14}\ \mathbf{\left(-3.58\%\right)}$
        \\
        \toprule
            \multirow{2}{*}{10000} 
            & 
            $\textrm{fwd}$ 
            & 
            $\reading{31.17}{0.07}$
            & 
            $\reading{25.49}{0.05}\ \mathbf{\left(-18.20\%\right)}$
            &   
            $\textrm{Inference}$
            & 
            $\reading{38.71}{0.19}$
            & 
            $\reading{35.37}{0.17}\ \mathbf{\left(-8.62\%\right)}$
        \\
        \cmidrule{2-4} 
        \cmidrule{5-7}  
            & 
            $\textrm{fwd} + \textrm{bwd}$  
            & 
            $\reading{117.53}{0.13}$
            & 
            $\reading{109.87}{0.12}\ \mathbf{\left(-6.52\%\right)}$
            &   
            $\textrm{Training}$
            & 
            $\reading{5990.72}{2.21}$
            & 
            $\reading{5751.43}{5.77}\ \mathbf{\left(-3.99\%\right)}$
        \\
        \bottomrule
        \\
    \end{tabular}
    }
    \caption{
        \textsc{FlashSigmoid} vs. \textsc{FlashAttention2} on H100 nodes.
        The kernel GPU time for both the approaches is reported in milliseconds and wall-clock times is reported in seconds per epoch. 
    } 
    \label{fig:ComparisonsOfFlashesOnH100}
    \end{sc}
\end{table}
\begin{table}[htbp]
    \tiny
    \centering
    \begin{sc}
    \resizebox{\columnwidth}{!}{%
    \begin{tabular}{@{\extracolsep{4pt}}ccccccc}
        \toprule
            \multirow{2}{*}{Tokens}
            &
            \multicolumn{3}{c}{Kernel GPU Time Comparison}
            &
            \multicolumn{3}{c}{Full Run Wall-Clock Time Comparison} 
        \\
        \cmidrule{2-4} 
        \cmidrule{5-7} 
            & 
            Kernels
            & 
            \textsc{FlashAttention2} (ms)
            & 
            \textsc{FlashSigmoid} (ms)
            &  
            Mode
            & 
            \textsc{FlashAttention2} (s)
            & 
            \textsc{FlashSigmoid} (s) 
        \\ 
        \toprule
            \multirow{2}{*}{4096} 
            & 
            $\textrm{fwd}$ 
            & 
            $\reading{8.32}{0.02}$
            & 
            $\reading{7.84}{0.03}\ \mathbf{\left(-5.79\%\right)}$
            &   
            $\textrm{Inference}$
            & 
            $\reading{19.05}{0.22}$
            & 
            $\reading{18.74}{0.19}\ \mathbf{\left(-1.65\%\right)}$
        \\
        \cmidrule{2-4} 
        \cmidrule{5-7}  
            & 
            $\textrm{fwd} + \textrm{bwd}$  
            &  
            $\reading{31.81}{0.08}$
            & 
            $\reading{31.11}{0.08}\ \mathbf{\left(-2.19\%\right)}$
            &  
            $\textrm{Training}$
            & 
            $\reading{2795.03}{2.35}$
            & 
            $\reading{2769.44}{5.10}\ \mathbf{\left(-0.92\%\right)}$
        \\
        \toprule
            \multirow{2}{*}{8100} 
            & 
            $\textrm{fwd}$ 
            & 
            $\reading{33.65}{0.09}$
            & 
            $\reading{27.92}{0.07}\ \mathbf{\left(-17.04\%\right)}$
            &  
            $\textrm{Inference}$
            & 
            $\reading{47.35}{0.20}$
            & 
            $\reading{44.05}{0.17}\ \mathbf{\left(-6.96\%\right)}$
        \\
        \cmidrule{2-4} 
        \cmidrule{5-7}  
            & 
            $\textrm{fwd} + \textrm{bwd}$  
            & 
            $\reading{128.18}{0.13}$
            & 
            $\reading{119.04}{0.12}\ \mathbf{\left(-7.13\%\right)}$
            & 
            $\textrm{Training}$ 
            & 
            $\reading{7519.64}{4.21}$
            & 
            $\reading{7254.84}{12.64}\ \mathbf{\left(-3.52\%\right)}$
        \\
        \toprule
            \multirow{2}{*}{10000} 
            & 
            $\textrm{fwd}$ 
            & 
            $\reading{51.17}{0.07}$
            & 
            $\reading{42.49}{0.06}\ \mathbf{\left(-16.96\%\right)}$
            &   
            $\textrm{Inference}$
            & 
            $\reading{64.61}{0.32}$
            & 
            $\reading{59.55}{0.18}\ \mathbf{\left(-7.82\%\right)}$
        \\
        \cmidrule{2-4} 
        \cmidrule{5-7}  
            & 
            $\textrm{fwd} + \textrm{bwd}$  
            & 
            $\reading{194.54}{0.14}$
            & 
            $\reading{180.59}{0.15}\ \mathbf{\left(-7.17\%\right)}$
            &   
            $\textrm{Training}$
            & 
            $\reading{10455.64}{8.85}$
            & 
            $\reading{10052.04}{18.87}\ \mathbf{\left(-3.86\%\right)}$
        \\
        \bottomrule
        \\
    \end{tabular}
    }
    \caption{
        \textsc{FlashSigmoid} vs. \textsc{FlashAttention2} on A100 nodes.
        The kernel GPU time for both the approaches is reported in milliseconds and wall-clock times is reported in seconds per epoch. 
    } 
    \label{fig:ComparisonsOfFlashesOnA100}
    \end{sc}
\end{table}

\noindent\textbf{Connection of Wall-Clock Time Speed-Up and Kernel Speed-Up:}\ 
From \cref{fig:ComparisonsOfFlashesOnA100,fig:ComparisonsOfFlashesOnH100}, it is clear that the speed-up in kernels is larger than that in the wall-clock times of the full runs.
In fact, the speed-up in kernels is the upper bound for the speed-up that we would see in wall-clock times.
To see why, let us denote by $\tau_{\textrm{sm}}$ and $\tau_{\sigma}$ the total kernel GPU time for softmax attention and sigmoid attention respectively. 
Then, the kernel speed-up is given by $s_{\textrm{kernel}} := 1 -  \frac{\tau_{\sigma}}{\tau_{\textrm{sm}}}$. 
However, in a full run, the total wall clock time also incorporates the time required to load data, time taken by other layers of the underlying models, time required to communicate gradients and other data across GPUs and across nodes, and so on.
For our corresponding sigmoid and softmax runs, these extra factors are designed to add, upon expectation, in the same extra time $\tau$.
Thus, the wall-clock time speed-up of a full run with end-to-end training is $s_{\textrm{wall-clock}} := 1 - \frac{\tau_{\sigma} + \tau}{\tau_{\textrm{sm}} + \tau}$. 
Since we have faster sigmoid kernels, we have $\tau_\sigma < \tau_{\textrm{sm}}$, which in turn shows that $s_{\textrm{wall-clock}} = 1 - \frac{\tau_{\sigma} + \tau}{\tau_{\textrm{sm}} + \tau} <  1 -  \frac{\tau_{\sigma}}{\tau_{\textrm{sm}}} = s_{\textrm{kernel}}$. 
This explains the speed boost trends in kernel time versus full run wall-clock time for each setting in \cref{fig:ComparisonsOfFlashesOnA100,fig:ComparisonsOfFlashesOnH100}. 
However, in particular, if a model performs attention mechanism over large number of tokens, the attention mechanism, and hence the corresponding kernel time, starts to dominate the other computations in the network. 
In that case, we see that the wall-clock time speed-boost is closer to the kernel speed-boost.
Mathematically, if $\tau_{\sigma}, \tau_{\textrm{sm}} >\!\!> \tau$, we have: $\tau_{\sigma} + \tau\approx \tau_{\sigma}$, $\tau_{\textrm{sm}} + \tau\approx \tau_{\textrm{sm}}$. 
Thus, $s_{\textrm{kernel}}\approx s_{\textrm{wall-clock}}$, thereby making ${s_{\textrm{wall-clock}}}/{s_{\textrm{kernel}}}\rightarrow 1$.

\begin{figure}[!htbp]
    \centering
    \begin{minipage}{0.46\textwidth}
        \footnotesize
        \centering
        \includegraphics[trim={0 0 0 0}, width=\textwidth]{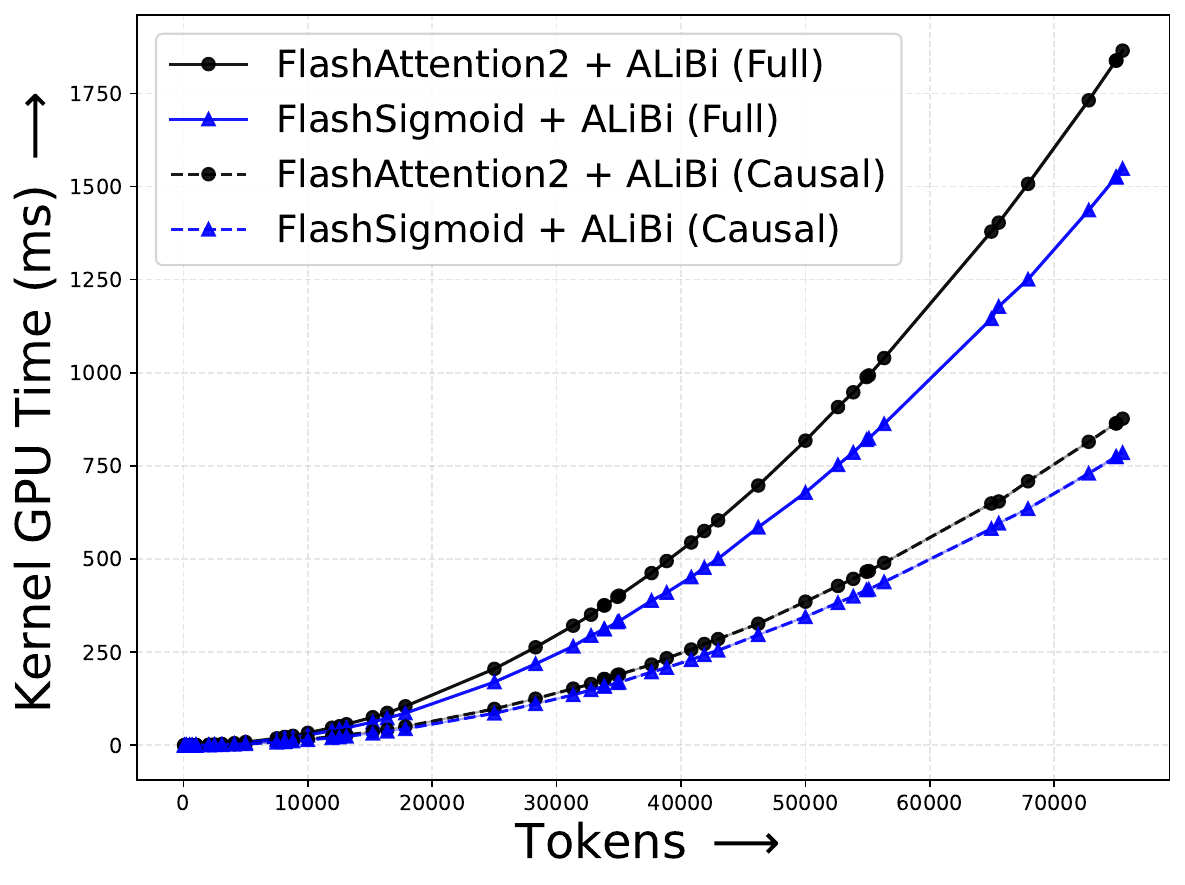}
        \captionsetup{justification=centering}
        \caption*{
            (a) Inference mode kernels on H100. 
        }
    \end{minipage}
    \hfill
    \begin{minipage}{0.46\textwidth}
        \centering        
        \includegraphics[trim={0 0 0 0}, width=\textwidth]{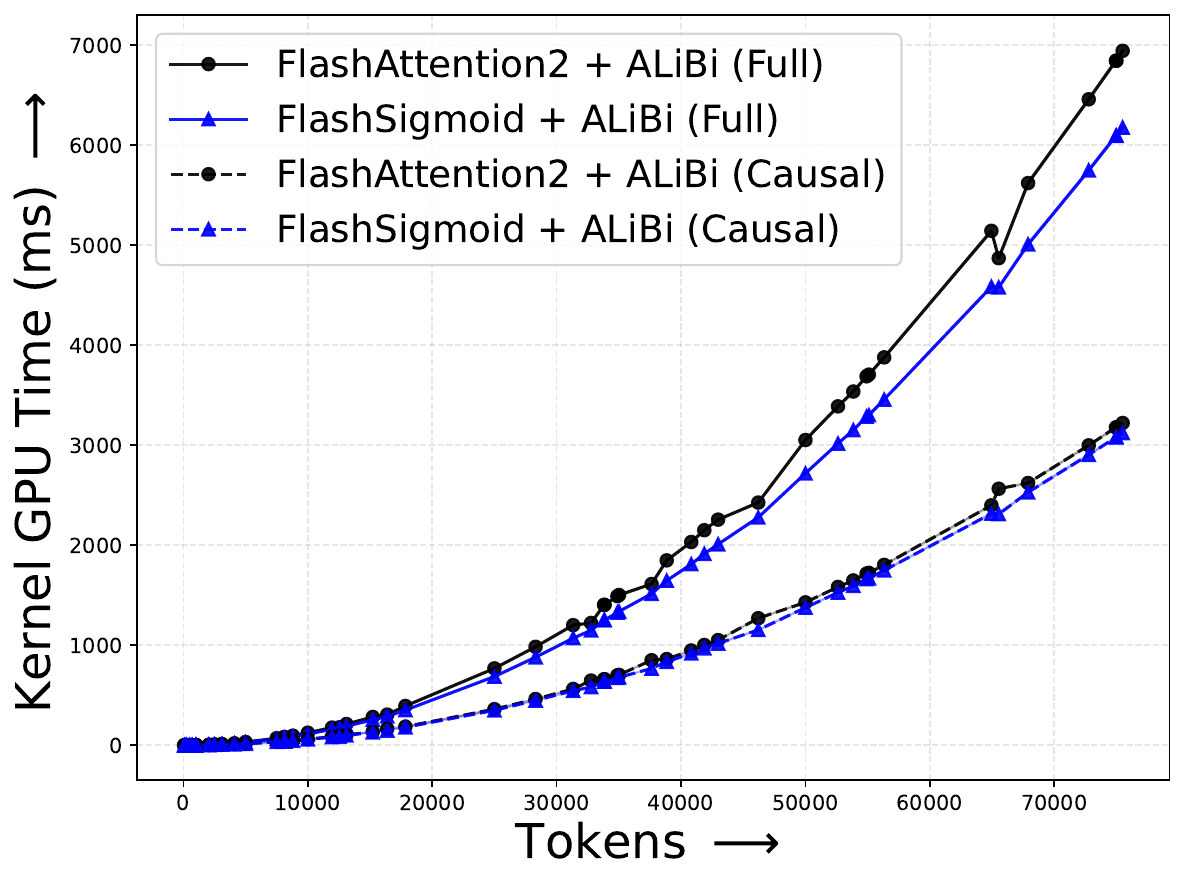}
        \captionsetup{justification=centering} 
        \caption*{
            (b) Training mode kernels on H100. 
        }
    \end{minipage}
    \caption{
        On average, for sequence lengths between $[64, 78\mathrm{k}]$, the inference mode kernel of \textsc{FlashSigmoid} is ${17.04}\%$ faster than \textsc{FlashAttention2} for self-attention and ${10.87}\%$ for causal attention.
        The training mode kernels of \textsc{FlashSigmoid} are ${8.91}\%$ faster than \textsc{FlashAttention2} for self-attention and ${4.72}\%$ for causal attention.
        Note that inference involves only the forward pass of the model and training involves both the forward and the backward pass of the model.
    }
    \label{fig:h100-softmax_alibi-sigmoid_alibi-fwd_bwd}
\end{figure}
\begin{figure}[!htbp]
    \centering
    \begin{minipage}{0.46\textwidth}
        \footnotesize
        \centering
        \includegraphics[trim={0 0 0 0}, width=\textwidth]{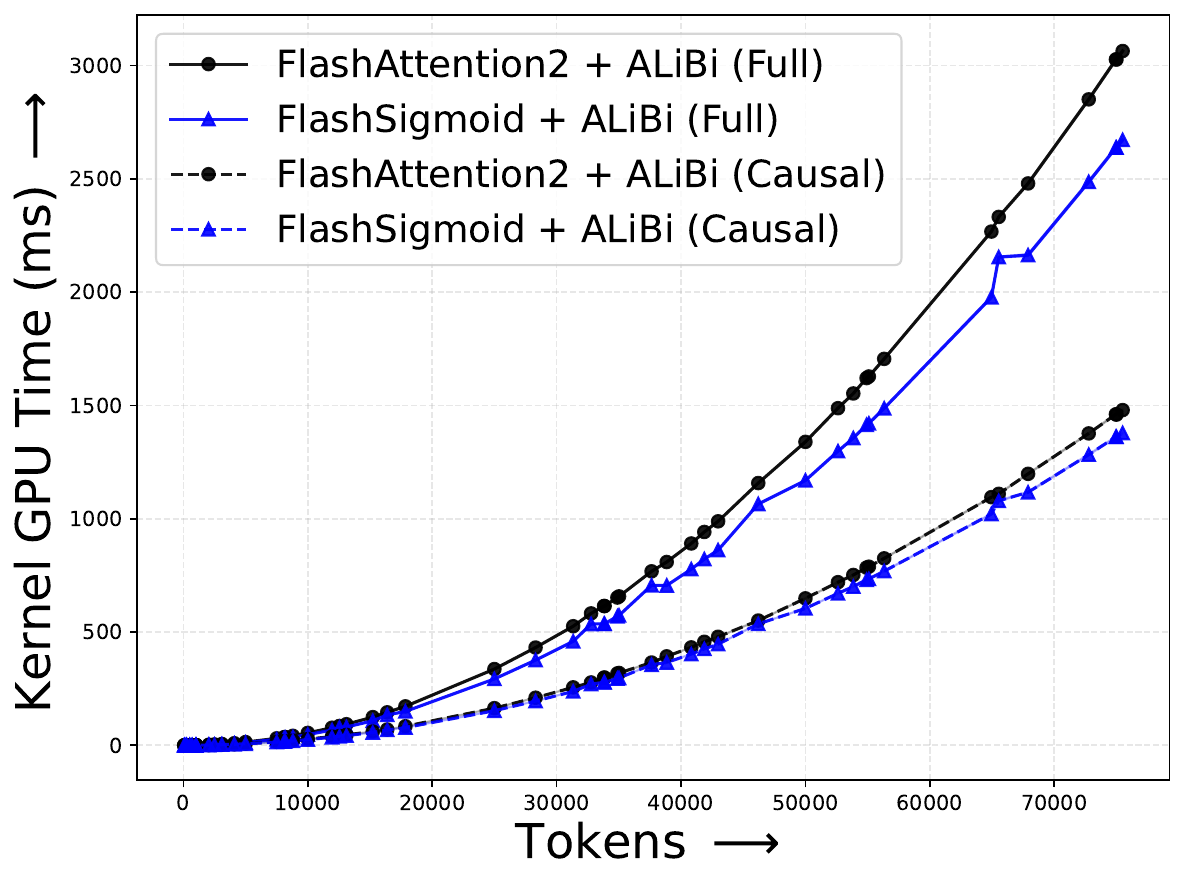}
        \captionsetup{justification=centering}
        \caption*{
            (a) Inference mode kernels on A100. 
        }
    \end{minipage}
    \hfill
    \begin{minipage}{0.46\textwidth}
        \centering        
        \includegraphics[trim={0 0 0 0}, width=\textwidth]{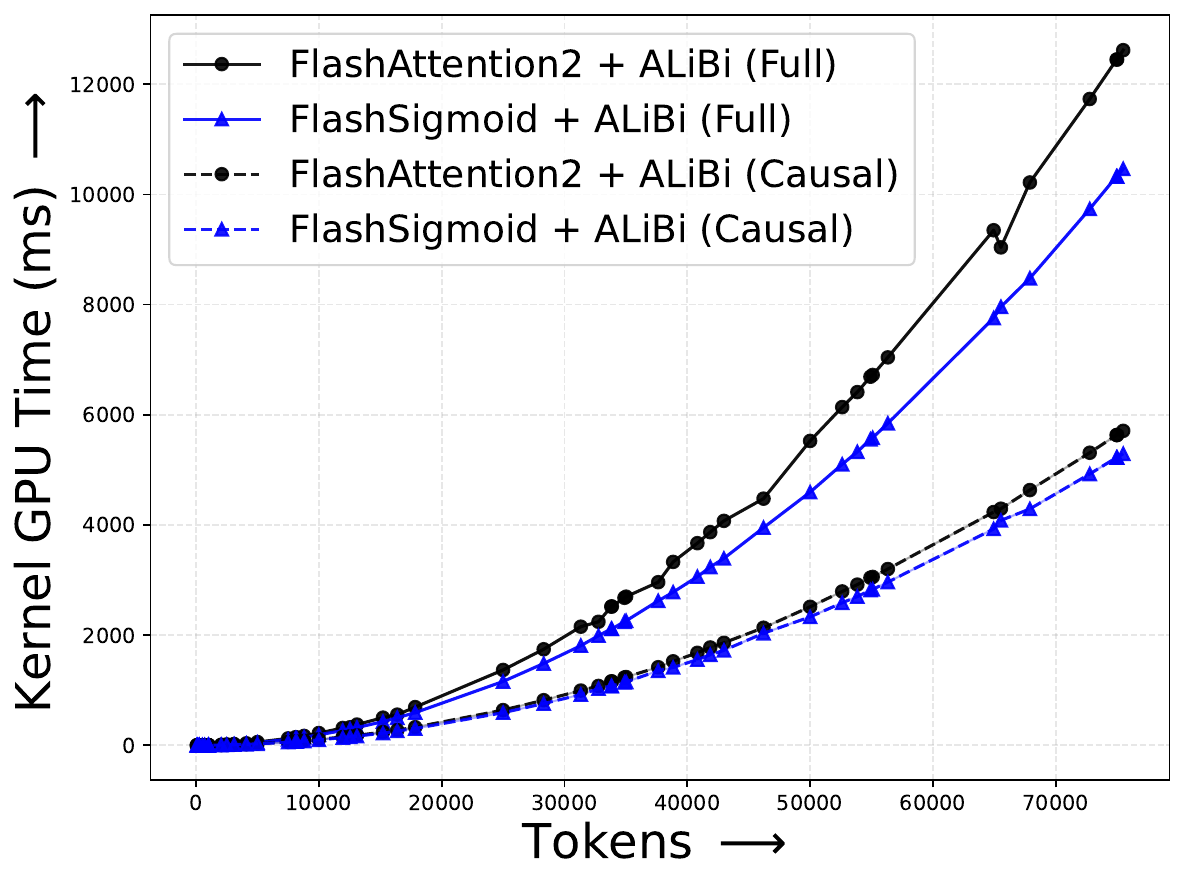}
        \captionsetup{justification=centering} 
        \caption*{
            (b) Training mode kernels on A100. 
        }
    \end{minipage}
    \caption{
        On average, for sequence lengths between $[64, 78\mathrm{k}]$, the inference mode kernel of \textsc{FlashSigmoid} is ${12.28}\%$ faster than \textsc{FlashAttention2} for self-attention and ${5.30}\%$ for causal attention.
        The training mode kernels of \textsc{FlashSigmoid} are ${14.64}\%$ faster than \textsc{FlashAttention2} for self-attention and ${6.80}\%$ for causal attention.
        Note that inference involves only the forward pass of the model and training involves both the forward and the backward pass of the model.
    }
    \label{fig:a100-softmax_alibi-sigmoid_alibi-fwd_bwd}
    \vspace{-0.1in}
\end{figure}
\paragraph{Significance of Wall-Clock Speed-Up of Inference:} Although \textsc{FlashSigmoid} provides only modest gains during training, the speed-up in inference is significant ($> 15\%$ for underlying kernels and $5-10\%$ during inference of full runs).
We posit that this speed-up in inference is extremely critical as well.
Contemporary large-scale models, once trained, spend a huge portion of the rest their lifetime in inference mode~\citep{DBLP:journals/corr/abs-2303-08774}. 
Thus, significant performance boosts in inference mode have immense potential for saving resources in deployment of large models for inference. 
\vspace{-0.1in}
\subsection{\textsc{FlashSigmoid} with ALiBi}
\label{sec:FlashSigmoidWithALiBi}
\noindent It is evident from the main text of the paper that improved positional embeddings, like ALiBi~\citep{DBLP:conf/iclr/PressSL22}, can be crucial for certain tasks and data modalities. 
Thus, we also provide a \textsc{FlashSigmoid} implementation that incorporates ALiBi.
We compare the \textsc{FlashSigmoid} with ALiBi implementation with the \textsc{FlashAttention2} with ALiBi implementation~\citep{DBLP:journals/corr/abs-2307-08691}.
\Cref{fig:a100-softmax_alibi-sigmoid_alibi-fwd_bwd,fig:h100-softmax_alibi-sigmoid_alibi-fwd_bwd} show the kernel GPU time for the forward and backward pass kernels of \textsc{FlashSigmoid} with ALiBi implementation versus \textsc{FlashAttention2} with ALiBi implementation.
Again, we observe that \textsc{FlashSigmoid} kernels for inference have significant speed-up in wall-clock time over those in \textsc{FlashAttention2} and the kernels for training also have modest wall-clock improvements. 

\subsection{Directions for Future Work on \textsc{FlashSigmoid}}
\label{sec:DirectionsForFutureWorkOnFlashSigmoid}
\noindent In this section, we discussed \textsc{FlashSigmoid}, a hardware-aware implementation of the $\sigmoidattn$ algorithm. Then, we demonstrated via kernel benchmarking and realistic setting runs that \textsc{FlashSigmoid} provides significant gains in inference as well as modest gains in training of models with attention mechanism. 
In this subsection we further discuss additional avenues for improving the implementation of \textsc{FlashSigmoid}, and point out some interesting directions for future work.
\paragraph{Optimization of Block Shapes for Different Input and GPU Settings:}\ 
As stated before, our \textsc{FlashSigmoid} implementation builds on \textsc{FlashAttention2} by adding functionality for forward and backward pass of sigmoid attention in place of the standard softmax attention.
In particular, for all \textsc{FlashSigmoid} results discussed so far, we inherit directly from \textsc{FlashAttention2} the details of optimal block shapes, grid shapes, and other kernel launch parameters, and keep them unchanged in our implementation.
For instance, this is the case for the block sizes $B _r, B_c$ in~\cref{alg:FlashSigmoidForward,alg:FlashSigmoidBackward}, which are identical in \textsc{FlashAttention2} and \textsc{FlashSigmoid}.
This choice is dictated by the need to ensure a fair comparison between the two implementations, and allows us to demonstrate the speed-up of sigmoid attention by minimizing confounders associated with parallel computations on different GPU architectures for different input shapes. 

\noindent Although \textsc{FlashSigmoid} kernels lead to speed-ups in inference and training for both H100 and A100 GPUs, we observe that the kernel timing speed-ups on A100 are not uniform across sequence lengths: for a small subset of these, our kernel provides significantly lower speed-up compared to the overall trend for other sequence lengths.
Ideally, the implementation of attention mechanisms should not assume any information on the token count in input, and it is then desirable to have uniform speed-ups across all input lengths. 
Here, we show that this is achievable by simply updating the block shape information in \textsc{FlashSigmoid} to values that are different than those in \textsc{FlashAttention2}. 
The implementation of \textsc{FlashAttention2} is templated according to block shapes, grid shapes, and other kernel launch parameters.
\noindent Note that \textsc{FlashAttention2} provides various tailored implementations, optimized for different input shapes (e.g., different ranges of feature dimension per head), input types (e.g., causal attention vs. self-attention, ALiBi vs. no ALiBi in attention, etc.), and GPU types (e.g., A100 vs. H100 via checking shared memory size on GPUs). This is achieved by opportunely selecting the kernel template parameters defining block shapes, grid shapes, and other kernel launch parameters for parallel computation on GPUs. 
In our case, we create a variant of \textsc{FlashSigmoid}, denoted by $\textsc{FlashSigmoid}^{\dagger}$, where we update the block sizes for query and key tensors from $\left(B_r, B_c\right) = \left(128, 128\right)$ of \textsc{FlashSigmoid} to $\left(B_r, B_c\right) = \left(128, 64\right)$ of $\textsc{FlashSigmoid}^{\dagger}$ \emph{only} for our input setting (template with features per head being $64$). 
\begin{figure}[!htbp]
    \centering
    \begin{minipage}{0.46\textwidth}
        \footnotesize
        \centering
        \includegraphics[trim={0 0 0 0}, width=\textwidth]{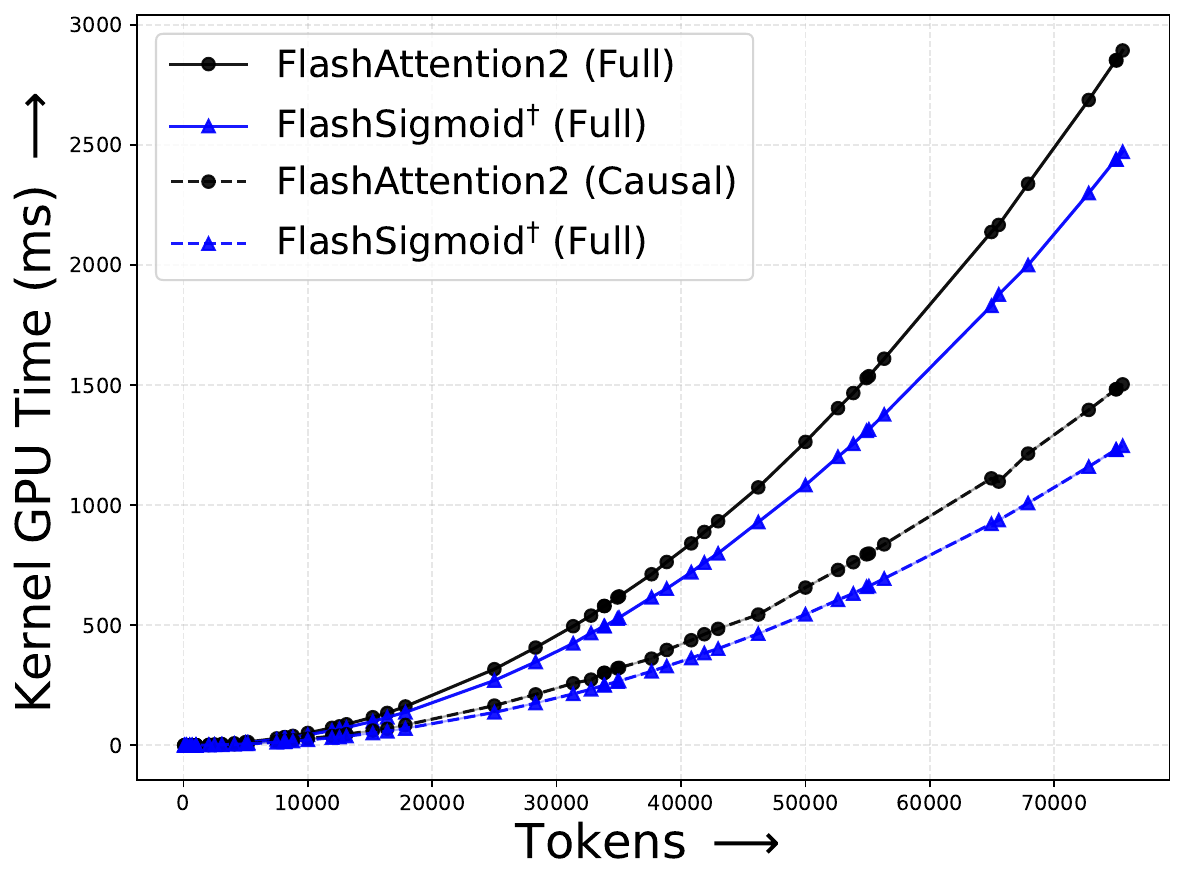}
        \captionsetup{justification=centering}
        \caption*{
            (a) Inference mode kernels on A100.
        }
    \end{minipage}
    \hfill
    \begin{minipage}{0.46\textwidth}
        \centering        
        \includegraphics[trim={0 0 0 0}, width=\textwidth]{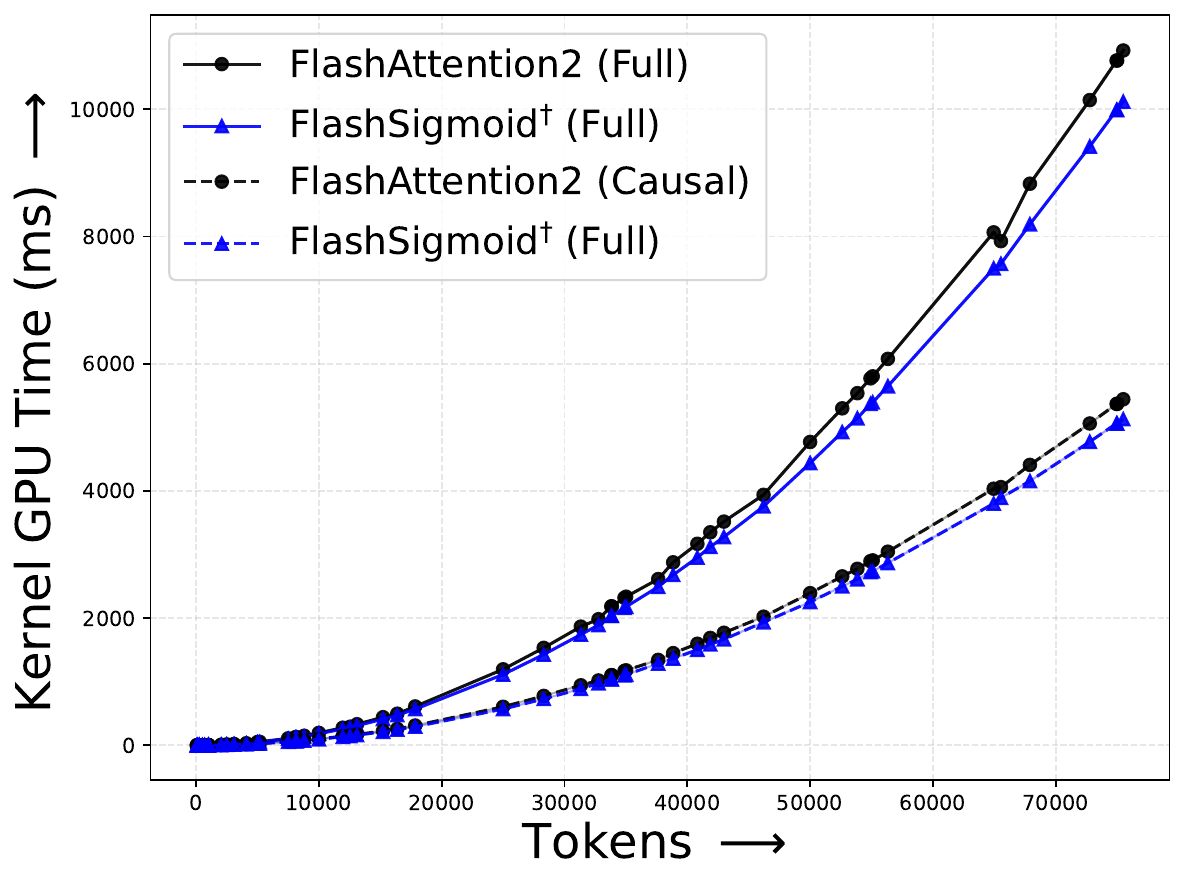}
        \captionsetup{justification=centering} 
        \caption*{
            (b) Training mode kernels on A100.
        }
    \end{minipage}
    \caption{
        On average, for sequence lengths between $[64, 78\mathrm{k}]$, the inference mode kernel of $\textsc{FlashSigmoid}^{\dagger}$ is ${14.82}\%$ faster than \textsc{FlashAttention2} for self-attention and ${18.02}\%$ for causal attention.
        The training mode kernels of $\textsc{FlashSigmoid}^{\dagger}$ are ${6.18}\%$ faster than \textsc{FlashAttention2} for self-attention and ${5.76}\%$ for causal attention.
        Note that inference involves only the forward pass of the model and training involves both the forward and the backward pass of the model. 
    }
    \label{fig:a100-softmax-sigmoid-fwd-bwd-special-variant}
    \vspace{-0.1in}
\end{figure}
\paragraph{Experimentation and Results:} For this variant, we perform kernel benchmarking as described in~\cref{sec:PerformanceAnalysisOfFlashSigmoidKernels}, and report the corresponding results in \cref{fig:a100-softmax-sigmoid-fwd-bwd-special-variant}.
Comparing the plots for kernel timing with \textsc{FlashSigmoid} plots from~\cref{fig:a100-softmax-sigmoid-fwd-bwd}, we observe that $\textsc{FlashSigmoid}^{\dagger}$ not only provides a more uniform inference and training kernel speed-up on \emph{all} sequence lengths, but also improves the average of these speed-ups across all lengths.
To further bolster our observations, \cref{fig:ComparisonsOfFlashSigmoidVariantsOnA100} shows the inference mode and training mode kernel speed-ups for a subset of sequence lengths under consideration.
This experiment indicates that it is possible to obtain higher and more uniform speed-ups in kernel timings across a wide range of tokens by investigating optimal block shape, grid shape, and other kernel launch parameters for each input setting and GPU type.
We leave this optimization for future work. 
\begin{table}[htbp]
    \tiny
    \centering
    \begin{sc}
    \resizebox{\columnwidth}{!}{%
    \begin{tabular}{@{\extracolsep{4pt}}ccccc}
        \toprule
            \multirow{2}{*}{Tokens}
            &
            \multicolumn{4}{c}{Kernel GPU Time Comparison}
        \\
        \cmidrule{2-5} 
            & 
            Kernels
            & 
            \textsc{FlashAttention2} (ms)
            & 
            \textsc{FlashSigmoid} (ms)
            & 
            $\textsc{FlashSigmoid}^{\dagger}$ (ms) 
        \\ 
        \toprule
            \multirow{2}{*}{4096} 
            & 
            $\textrm{fwd}$ 
            & 
            $\reading{8.32}{0.02}$
            & 
            $\reading{7.84}{0.03}\ \mathbf{\left(-5.79\%\right)}$
            & 
            $\reading{7.26}{0.02}\ \mathbf{\left(-13.21\%\right)}$
        \\
        \cmidrule{2-5}
            & 
            $\textrm{fwd} + \textrm{bwd}$  
            &  
            $\reading{31.81}{0.08}$
            & 
            $\reading{31.11}{0.08}\ \mathbf{\left(-2.19\%\right)}$
            &  
            $\reading{30.62}{0.09}\ \mathbf{\left(-4.03\%\right)}$
        \\
        \toprule
            \multirow{2}{*}{8100} 
            & 
            $\textrm{fwd}$ 
            & 
            $\reading{33.65}{0.09}$
            & 
            $\reading{27.92}{0.07}\ \mathbf{\left(-17.04\%\right)}$
            & 
            $\reading{28.54}{0.07}\ \mathbf{\left(-15.50\%\right)}$
        \\
        \cmidrule{2-5}
            & 
            $\textrm{fwd} + \textrm{bwd}$  
            & 
            $\reading{128.18}{0.13}$
            & 
            $\reading{119.04}{0.12}\ \mathbf{\left(-7.13\%\right)}$
            & 
            $\reading{119.85}{0.13}\ \mathbf{\left(-6.81\%\right)}$
        \\
        \toprule
            \multirow{2}{*}{10000} 
            & 
            $\textrm{fwd}$ 
            & 
            $\reading{51.17}{0.07}$
            & 
            $\reading{42.49}{0.06}\ \mathbf{\left(-16.96\%\right)}$
            & 
            $\reading{43.53}{0.09}\ \mathbf{\left(-15.32\%\right)}$
        \\
        \cmidrule{2-5} 
            & 
            $\textrm{fwd} + \textrm{bwd}$  
            & 
            $\reading{194.54}{0.14}$
            & 
            $\reading{180.59}{0.15}\ \mathbf{\left(-7.17\%\right)}$
            & 
            $\reading{181.97}{0.17}\ \mathbf{\left(-6.87\%\right)}$
        \\
        \toprule
            \multirow{2}{*}{16384} 
            & 
            $\textrm{fwd}$ 
            & 
            $\reading{134.19}{0.12}$
            & 
            $\reading{125.43}{0.10}\ \mathbf{\left(-6.53\%\right)}$
            & 
            $\reading{116.75}{0.10}\ \mathbf{\left(-13.40\%\right)}$
        \\
        \cmidrule{2-5} 
            & 
            $\textrm{fwd} + \textrm{bwd}$  
            & 
            $\reading{494.65}{0.28}$
            & 
            $\reading{482.08}{0.23}\ \mathbf{\left(-2.54\%\right)}$
            & 
            $\reading{474.52}{0.28}\ \mathbf{\left(-4.48\%\right)}$
        \\
        \bottomrule
        \\
    \end{tabular}
    }
    \caption{
        \textsc{FlashAttention2} vs.
        \textsc{FlashSigmoid} vs. $\textsc{FlashSigmoid}^{\dagger}$ on A100 nodes.
        The kernel GPU time for all three approaches are reported in milliseconds. 
        We observe that $\textsc{FlashSigmoid}^{\dagger}$ provides better and more uniform speed-ups across all example tokens.
    } 
    \label{fig:ComparisonsOfFlashSigmoidVariantsOnA100}
    \end{sc}
\end{table}

\vspace{-0.1in}
\section{Experiments}
\label{sec:appendix_experiments}
\vspace{-0.1in}
\subsection{Extra Ablations}
\label{sec:appendix_ablations}
\subsubsection{The Effect of Multiplicative Sequence Length Normalization}
\label{sec:appendix_normalization}
\begin{figure}[htbp]
    \centering
    \begin{minipage}{0.31\textwidth}
        \centering
        \includegraphics[width=\textwidth]{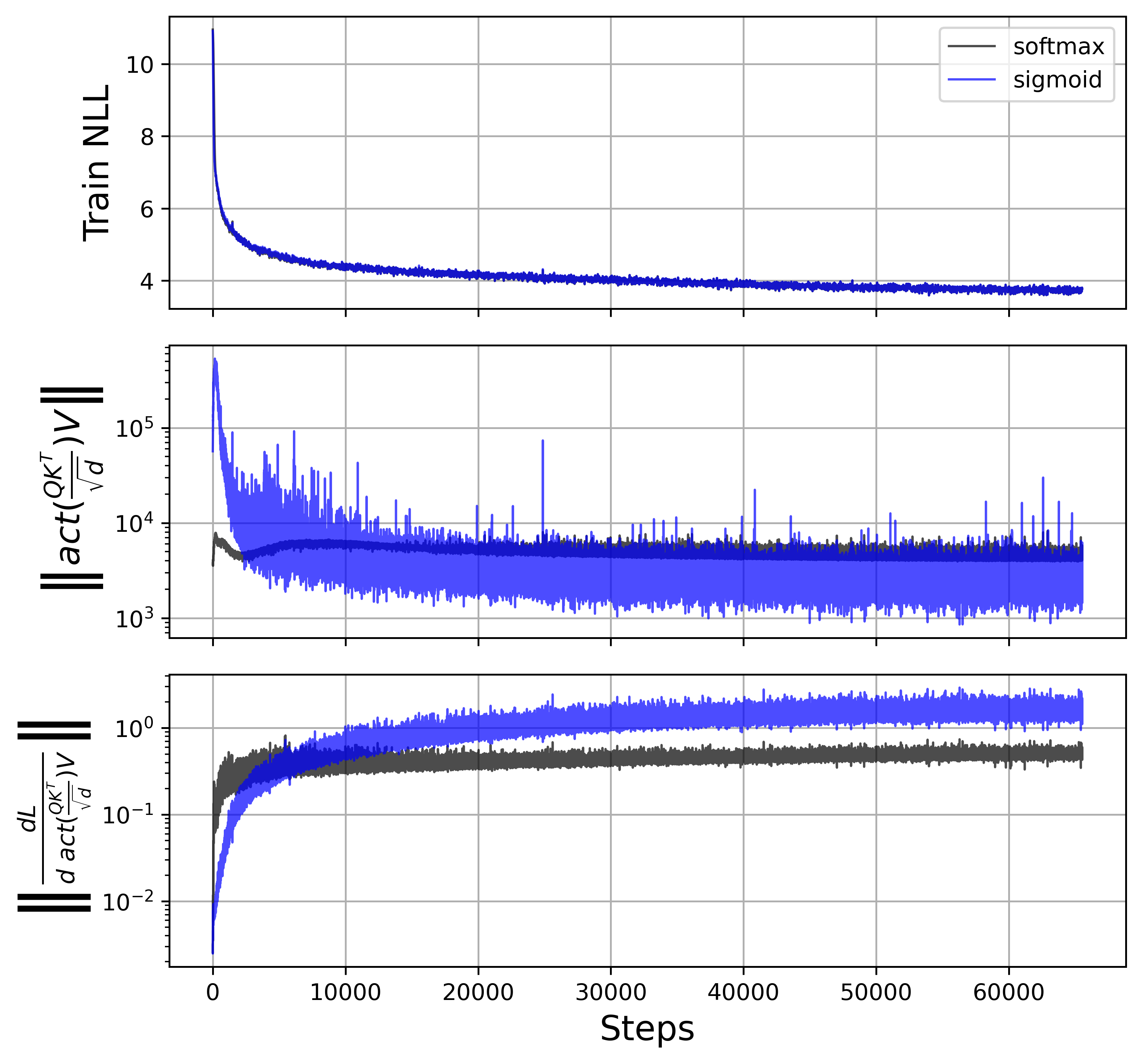}
        \captionsetup{justification=centering}
        \caption{$b = -\ln n$.}
        \label{fig:no_scaling}
    \end{minipage}\hfill
    \begin{minipage}{0.31\textwidth}
        \centering        
        \includegraphics[width=\textwidth]{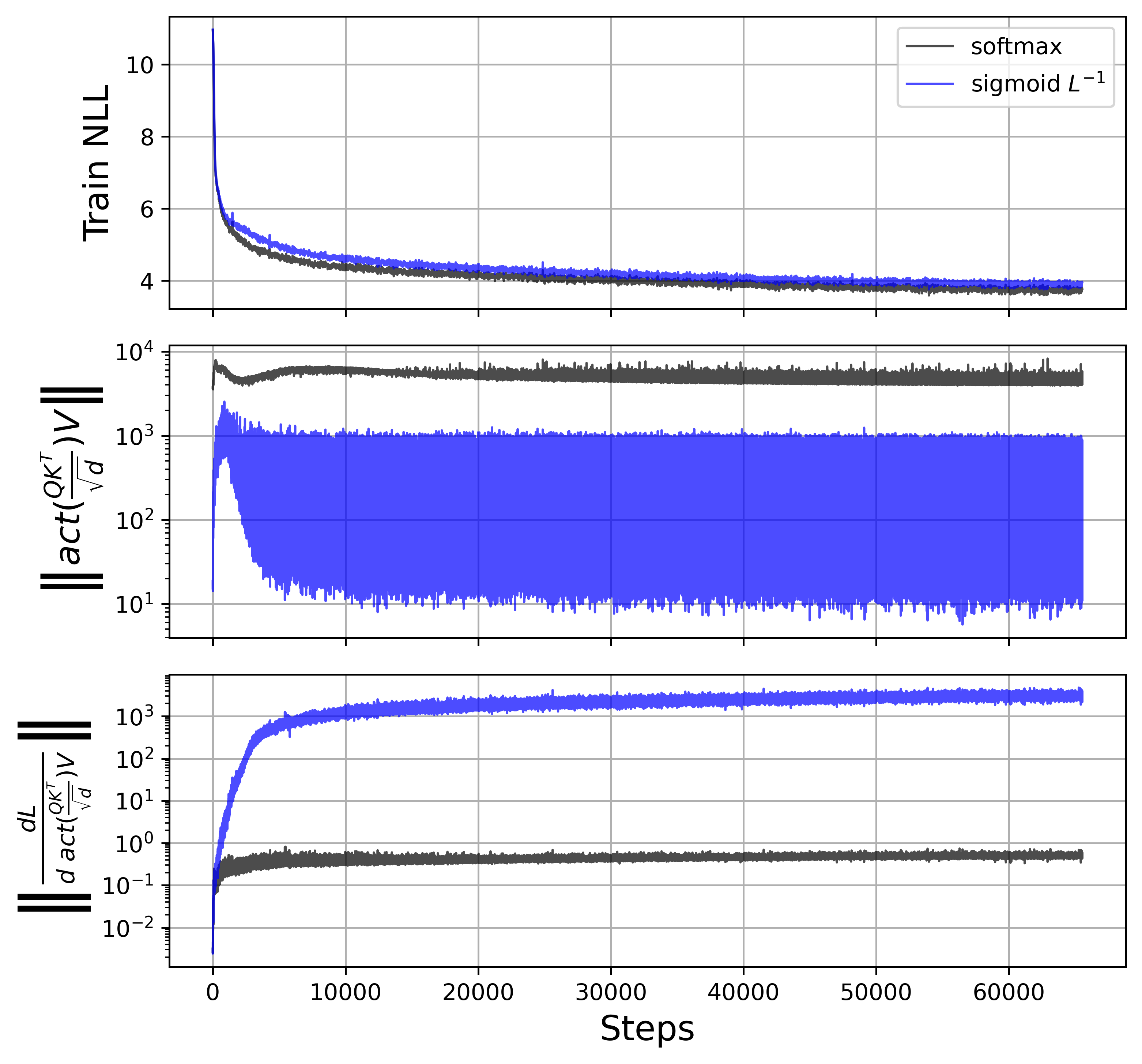}
        \captionsetup{justification=centering}
        \caption{$n^{-1}$ normalization.}
        \label{fig:seq_len_scaling}
    \end{minipage}\hfill
    \begin{minipage}{0.31\textwidth}
        \centering
        \includegraphics[width=\textwidth]{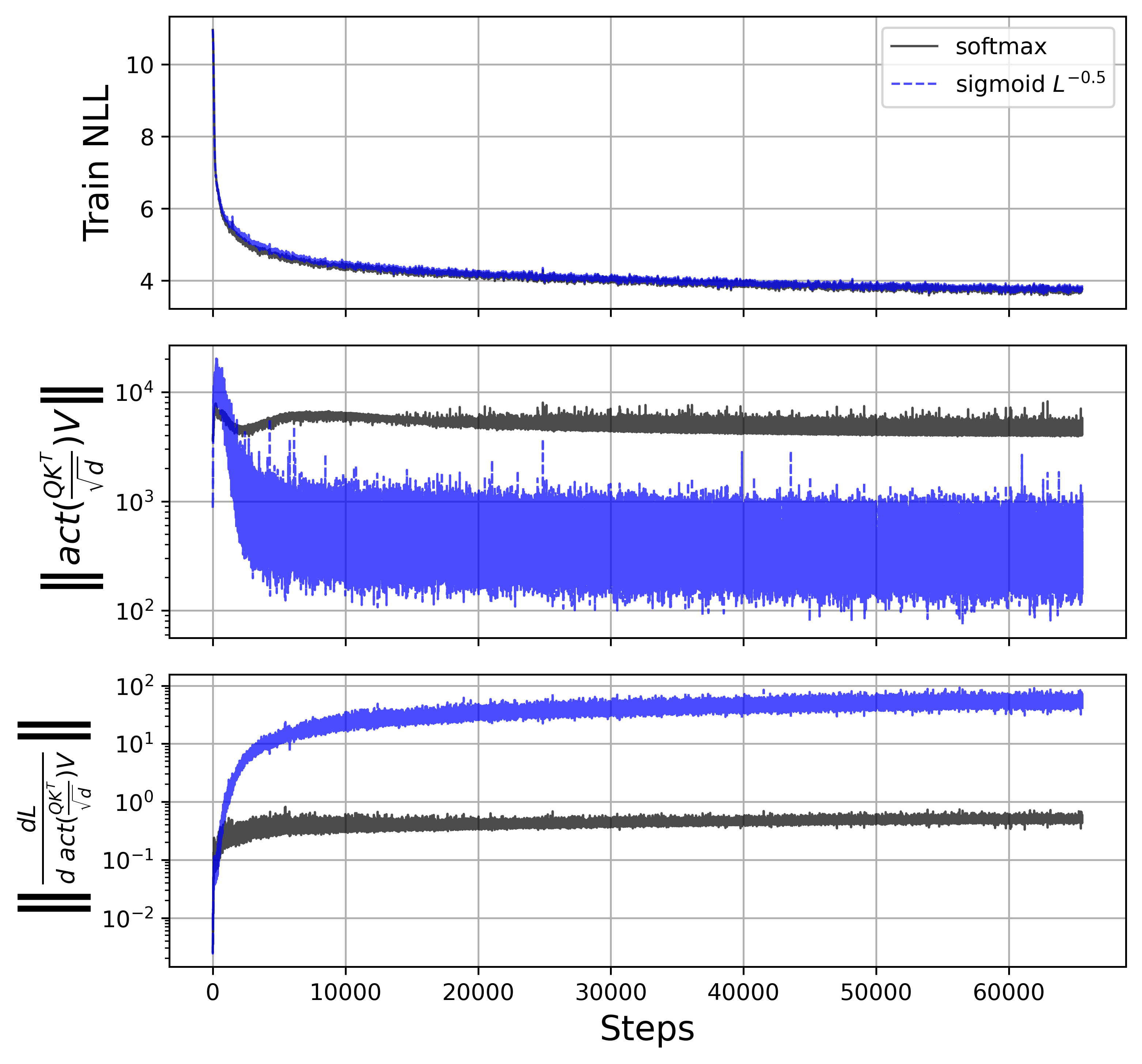}
        \captionsetup{justification=centering}
        \caption{$n^{-0.5}$ normalization.}
        \label{fig:sqrt_scaling}
    \end{minipage}
\end{figure}
\cite{wortsman2023replacing} notes that models trained with sigmoid or ReLU attention require scaling by the sequence length, $n^{-\alpha} \sigma(\mQ\mK^T / \sqrt{d_{qk}})\mV$. We ablate this by comparing the scaled solution to the one we propose in \cref{app:sigmoid_bias}. We also generalize the variant proposed in \citep{wortsman2023replacing} to variadic sequence lengths such that it works with auto-regressive (AR) training, for example for $n=3$:
\begin{align}
    \underbrace{\begin{bmatrix}
        1 & 1 & 1 \\
        0.5^{-\alpha} & 0.5^{-\alpha} & 1 \\
        0.33^{-\alpha} & 0.33^{-\alpha} & 0.33^{-\alpha} \\
    \end{bmatrix}}_{n^{-\alpha}} \odot
    \underbrace{\begin{bmatrix}
        1 & 0 & 0 \\
        1 & 1 & 0 \\
        1 & 1 & 1 \\
    \end{bmatrix}}_{\text{Causal Mask } \mM} \odot \   
     \sigma(\mQ \mK^T / \sqrt{d_{qk}}) \mV .
\label{eqn:ar_seq_len_normalization}
\end{align}
We repeat the experiment from \cref{fig:rope_vs_alibi}, using ALiBi positional embeddings for all trials. We apply $\alpha=\{1, 0.5\}$ AR normalization proposed in \Cref{eqn:ar_seq_len_normalization}. While there is an observable difference in terms of the attention norm, $\lVert \sigma(\mQ \mK^T / \sqrt{d_{qk}}) \mV \rVert$, we find that the train NLL is slightly worse for both normalized variants (\cref{fig:seq_len_scaling,fig:sqrt_scaling}) in comparison to the $b = -\ln n$  variant in \cref{fig:no_scaling}.
\subsubsection{Attention Bias Stability Ablation}
\label{sec:attn_bias_ablation}
To validate the stabilizing effects of attention bias we repeat the experiment from \cref{fig:qk_norm_ablation,fig:layerscale_ablation}, keeping all of the same hyper-parameters, while enabling QK norm and LayerScale (initialized at $10^{-4}$). We train with a range of constant bias offsets, $b \in \{-15, -10, -6, -4, -1 \}$ and visualize the results below in \cref{fig:const_attn_bias_ablation}.
\begin{figure}[htbp]
    \centering
    \includegraphics[width=\textwidth]{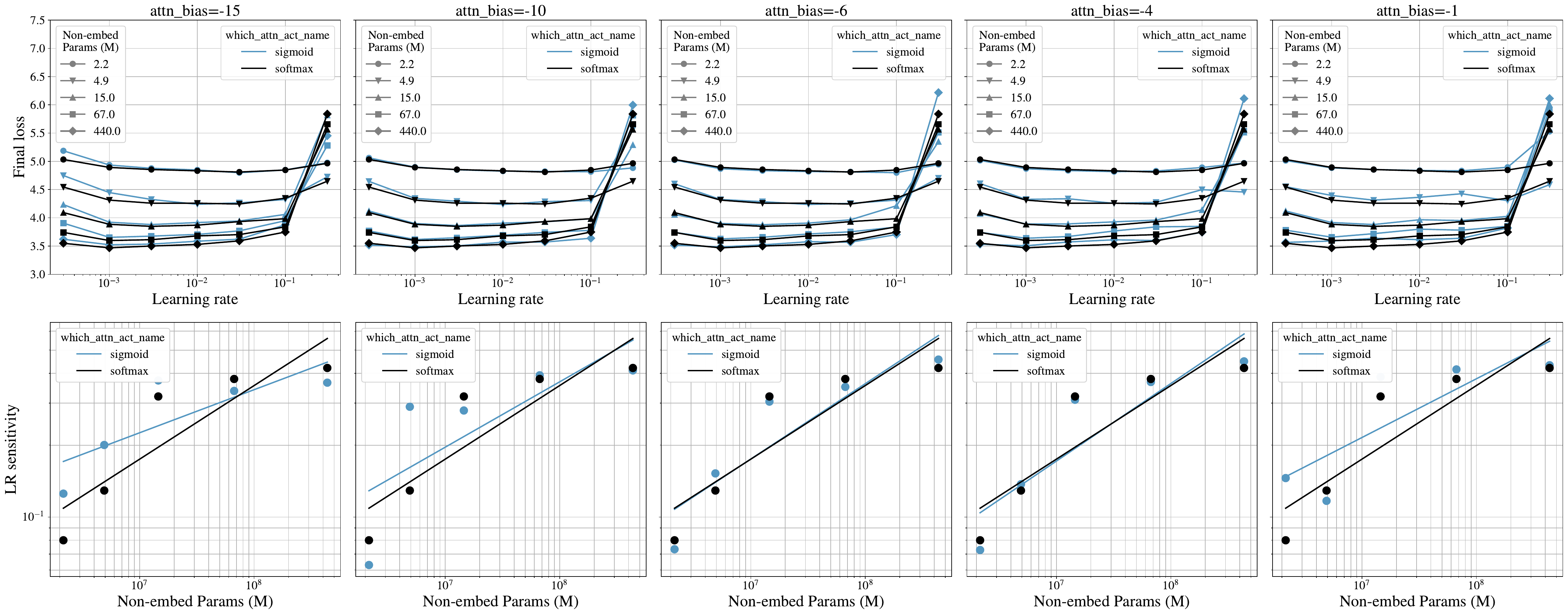}
    \captionsetup{justification=centering}
    \caption{Attention bias ablation.}
    \label{fig:const_attn_bias_ablation}
\end{figure}
We observe a systematic increase in stability (and lower $\sigmoidattn$ NLL) for values less than $-1$ up till $-10$, after which the $-15$ plot shows an over-regularizing effect with decreased performance.
\subsection{Vision}
\label{app:vision}
\subsubsection{Test ImageNet1k Top-1\%}
\label{app:top1_results}
\begin{figure}[htbp]
    \centering
    \includegraphics[width=\textwidth]{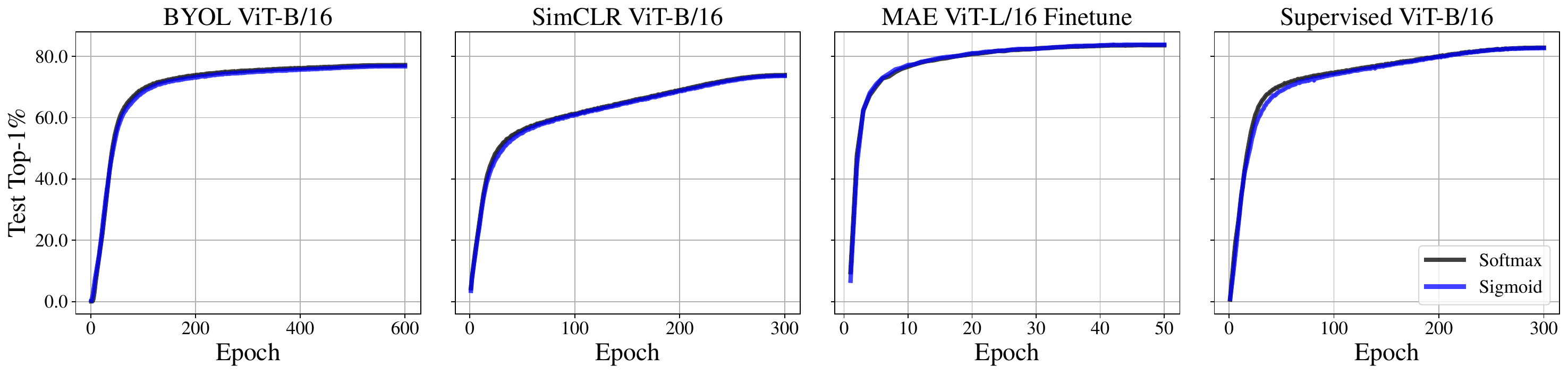}
    \caption{ImageNet1k test top-1\% for $\softmaxattn$ vs. $\sigmoidattn$ using models from \cref{fig:summary_nll}. %
    }
    \label{fig:test_top1_results}
\end{figure}
\cref{fig:test_top1_results} reports the test linear probe results for the ViT-B/16 BYOL \citep{DBLP:conf/nips/GrillSATRBDPGAP20, DBLP:conf/nips/BusbridgeRALDCW23}, ViT-B/16 SimCLR \citep{DBLP:conf/icml/ChenK0H20, DBLP:conf/icml/ZhaiLLBR0GS23} and the finetuned performance for the ViT-L/16 MAE \citep{DBLP:conf/cvpr/HeCXLDG22} and the test top-1\% results for for ViT-B/16 supervised model \citep{DBLP:conf/iclr/DosovitskiyB0WZ21}.
Across these wide range of SSL and supervised learning tasks, trained with contrastive (SimCLR), EMA distillation (BYOL) and reconstructive objectives (MAE), we find that $\sigmoidattn$ not only matches the training dynamics (\cref{fig:summary_nll}), but also the linear probe and finetuned performance of the baseline $\softmaxattn$.
\subsubsection{LayerScale Free Sigmoid Attention}
\label{sec:layerscale_free_sigmoid}
\begin{figure}[ht]
  \begin{minipage}{0.58\textwidth}
    \centering
    \includegraphics[width=\linewidth]{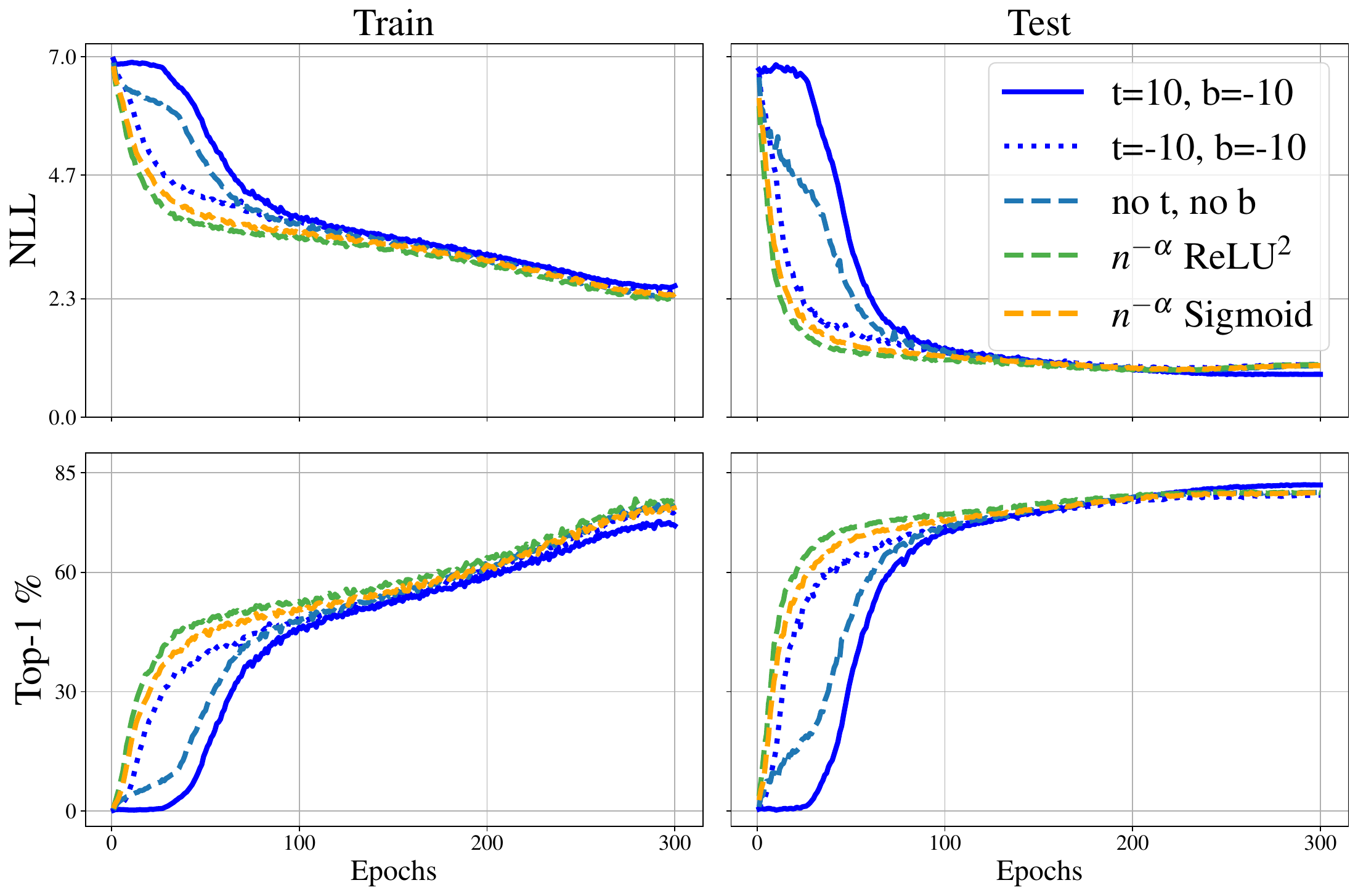}
  \end{minipage}%
  \hfill
  \begin{minipage}{0.4\textwidth}
    \caption{A competitive $\sigmoidattn$ ViT-B/16 model can be learned without LayerScale or QK norm using a large initial \emph{learnable} scalar temperature $t=10$ and bias $b=-10$ (similar to SigLIP \citep{DBLP:journals/corr/abs-2303-15343}): $\sigma(e^t [\mQ\mK^T / \sqrt{d_{qk}}] + b)\mV, \{b, t\} \in \mathbb{R}$. This regularizes the model, as it must move the temperature to a learnable regime. The $t=10,b=-10$ curve makes no progress in train NLL or test top-1 for $\sim$25 epochs (near max LR), but ultimately outperforms baselines.}    
    \label{fig:layerscale_free_sigmoid}
  \end{minipage}  
\end{figure}
While \cref{fig:layerscale_free_sigmoid} demonstrates the possibility of learning $\sigmoidattn$ without LayerScale, it involves task specific tuning of $\{t, b\}$. We also explored gating attention from learning (through a simple multiply by zero) for $\sim$25 epochs and were able to match the $t = 10, b = -10$ training curves from above. However, we opted for the LayerScale method due to its simplicity.
\subsubsection{Sigmoid Attention vs. Attention Relaxations}
\label{sec:attention_relaxations}
\vspace{-0.1in}
\begin{figure}[ht]
   \begin{minipage}{0.5\textwidth}
    \centering
    \includegraphics[width=\linewidth]{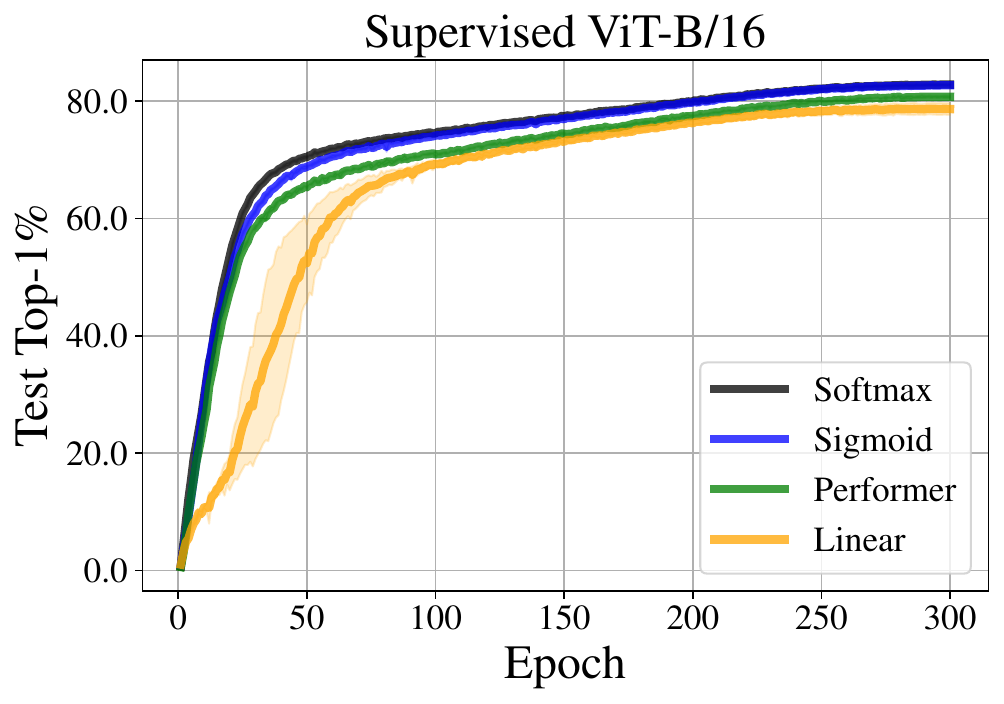}
  \end{minipage}%
  \hfill
  \begin{minipage}{0.48\textwidth}
    \caption{Supervised ViT-B/16 ImageNet1k classification. We contrast $\sigmoidattn$ and $\softmaxattn$ against (a) linear attention with no activation: $\mQ \mK^T / \sqrt{d_{qk}}$ and (b) fast attention via positive orthogonal random features, used in Performer \citep{DBLP:conf/iclr/ChoromanskiLDSG21}. $\sigmoidattn$, like $\softmaxattn$, differs from attention relaxations like Performer which uses low-rank representations of the attention matrix. $\sigmoidattn$ maintains performance parity with $\softmaxattn$, while outperforming other efficient attention variants.}
    \label{fig:attention_relaxations}
  \end{minipage}  
\end{figure}
\subsubsection{Hyper-Parameters}
\label{sec:appendix_vision_hyperparams}
\begin{table}[H]
\centering
\small
\caption{$\sigmoidattn$ SimCLR and BYOL ViT-B/16 hyperparameters.}
\label{tab:sigmoid_simclr_byol_attn_recipe}
\resizebox{0.7\textwidth}{!}{%
\begin{tabular}{lcc}
\toprule
Parameter & SimCLR & BYOL \\
\midrule
Attention bias & None & None \\
LayerScale Init & $10^{-4}$ & $10^{-4}$ \\
QK Norm & Yes & Yes \\
Pos Embed & SinCos & Learnable \\
\midrule
Freeze Patcher & Yes & No \\
Weight init & MocoV3~\citep{DBLP:conf/iccv/ChenXH21} & \texttt{trunc\_normal(.02)} \\
Normalization & LayerNorm & LayerNorm \\
LR schedule & Single Cycle Cosine & Single Cycle Cosine \\
LR warmup & 10 Epochs & 40 Epochs \\
Min LR & $1\times 10^{-6}$ & $1\times 10^{-6}$ \\
Training duration & 300 Epochs & 600 Epochs \\
Optimizer & AdamW & AdamW \\
Optimizer scaling rule & Linear & Linear \\
Base Adam ($\beta_1, \beta_2$) & (0.9, 0.95) & (0.9, 0.95) \\
Base LR & $2\times 10^{-4}$ & $1\times 10^{-4}$ \\
Base batch size & 256 & 256 \\
Total batch size & 4096 & 4096 \\
Base teacher momentum & - & 0.996 \\
Weight decay & 0.1 & 0.3 \\
Weight decay skip bias & Yes & Yes \\
Numerical precision & \texttt{bf16} & \texttt{bf16} \\
Stochastic depth & 0.0 & 0.2 \\
Augmentation stack & SimCLR~\citep{DBLP:conf/icml/ChenK0H20} & DINO multicrop~\citep{DBLP:conf/iccv/CaronTMJMBJ21} \\
Color Jitter Scaling & 0.5~\citep{DBLP:conf/iccv/ChenXH21} & 1.0 \\
\bottomrule
\end{tabular}
}
\end{table}
\begin{table}[H]
\centering
\small
\caption{$\sigmoidattn$ Supervised ViT-B/16 and MAE ViT-L/16 hyperparameters.}
\label{tab:sigmoid_mae_sup_attn_recipe}
\resizebox{0.7\textwidth}{!}{%
\begin{tabular}{lcc}
\toprule
Parameter & Supervised & MAE \\
\midrule
Attention bias & None & $b = - \ln{n}$ \\
LayerScale Init & $10^{-4}$ & $10^{-4}$ \\
QK Norm & Yes & Yes \\
Pos Embed & Learnable & Learnable \\
\midrule
Architecture & ViT-B/16 & ViT-L/16 \\
Mask Ratio & - & 0.75 \\
Freeze Patcher & No & No \\
Weight init & \texttt{trunc\_normal(.02)} & \texttt{trunc\_normal(.02)} \\
Normalization & LayerNorm & LayerNorm \\
LR schedule & Single Cycle Cosine & Single Cycle Cosine \\
LR warmup & 20 Epochs & 40 Epochs \\
Min LR & $1\times 10^{-6}$ & $0.0$ \\
Training duration & 300 Epochs & 400 Epochs \\
Optimizer & AdamW & AdamW \\
Optimizer scaling rule & Linear & Linear \\
Base Adam ($\beta_1, \beta_2$) & (0.9, 0.95) & (0.9, 0.95) \\
Base LR & $1\times 10^{-4}$ & $1.5\times 10^{-4}$ \\
Base batch size & 256 & 256 \\
Total batch size & 4096 & 4096 \\
Weight decay & 0.3 & 0.05 \\
Weight decay skip bias & Yes & Yes \\
Numerical precision & \texttt{bf16} & \texttt{bf16} \\
Stochastic depth & 0.28 & 0.0 \\
Augmentation stack & RandAug~\citep{DBLP:conf/nips/CubukZS020} & RRC + HFLIP \\
\bottomrule
\end{tabular}
}
\end{table}

\subsection{Language Model}
\label{sec:llm_appendix}
\subsubsection{Hyper-Parameters}
\begin{table}[h]
\centering
\caption{Training details for the Llama-style 1B LM training.}
\label{tab:lm_training_details}
\resizebox{0.45\textwidth}{!}{%
\begin{tabular}{@{}lll@{}}
\toprule
 & Parameter          & Value     \\ \midrule
 & Params             & 1B        \\
 & Context Length     & 2048      \\
 & Total Tokens       & 300B        \\
 & Batch size         & 4M tokens \\
 & LR Schedule        & Cosine    \\
 & LR Warmup Steps    & 5000      \\
 & Peak LR            & 1e-2      \\
 & Final LR           & 10\% of peak \\
 & Optimizer          & AdamW     \\
 & Optimizer momentum & 0.9, 0.95 \\
 & Weight decay       & 1e-4       \\
 & Gradient clipping  & 1.0       \\
 & Position encoding  & ALiBi     \\
 & Q/K Norm           & Applied   \\
 & Norm type          & RMSNorm \citep{zhang-sennrich-neurips19} \\
 & Norm structure     & Pre-norm  \\
 & Num layers         & 24        \\
 & Num heads          & 32        \\
 & Hidden dim         & 2048      \\
 \bottomrule
\end{tabular}
}
\end{table}
\cref{tab:lm_training_details} shows the hyper-parameters for the final comparison. MuP-simple \citep{DBLP:journals/corr/abs-2309-14322} is used, where the peak learning rate is set to 1e-2. Weight decay is decoupled, following \cite{loshchilov2017decoupled}. In addition, to confirm that applying QK-Norm does not hurt the baseline, we show training parity with and without QK-Norm in \cref{fig:lm_1b_qknorm}.
\begin{figure}[h]
    \centering
    \begin{minipage}{0.46\textwidth}
        \footnotesize
        \centering
        \includegraphics[trim={0 0 0 0}, width=\textwidth]{
            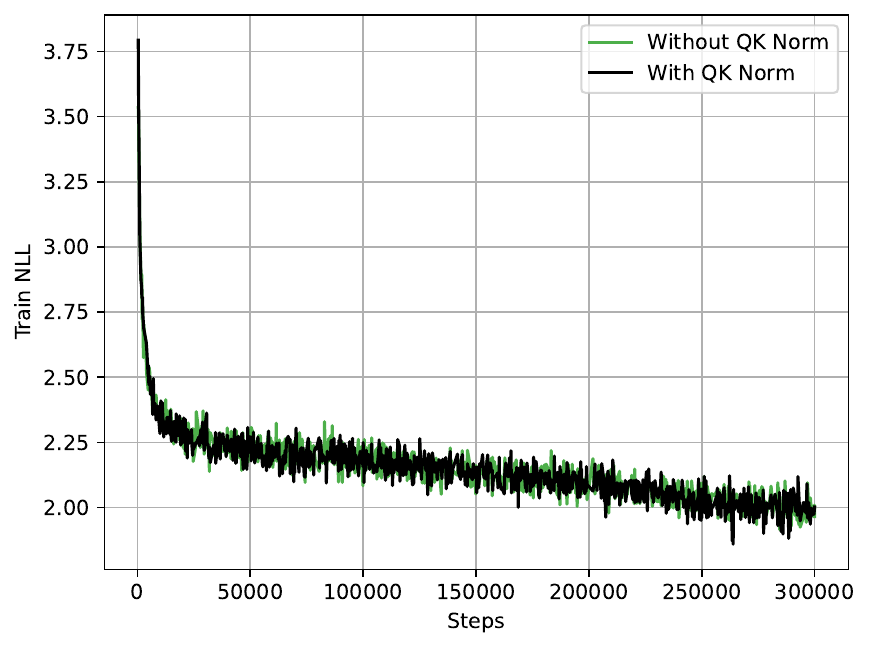
        }
        \captionsetup{justification=centering}
        \caption{
            1B $\softmaxattn$ LLM training with and without QK Norm, converging to the same loss.
        }
        \label{fig:lm_1b_qknorm}
    \end{minipage}
    \hfill
    \begin{minipage}{0.46\textwidth}
        \centering        
        \includegraphics[trim={0 0 0 0}, width=\textwidth]{
            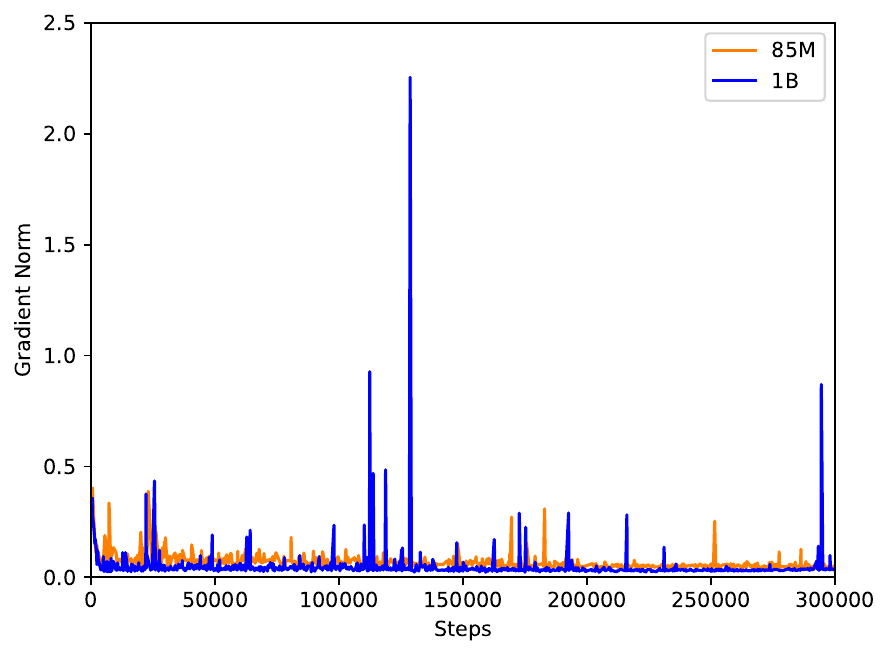
        }
        \captionsetup{justification=centering} 
        \caption{85M and 1B LLM training using $\sigmoidattn$ (n = 4096). Smooth training loss curves, but gradient norm shows spikes.}
        \label{fig:lm_grad_norm}
    \end{minipage}
\end{figure}
\begin{figure}[H]
    \centering
    \begin{minipage}{0.46\textwidth}
        \footnotesize
        \centering
        \includegraphics[trim={0 0 0 0}, width=\textwidth]{
            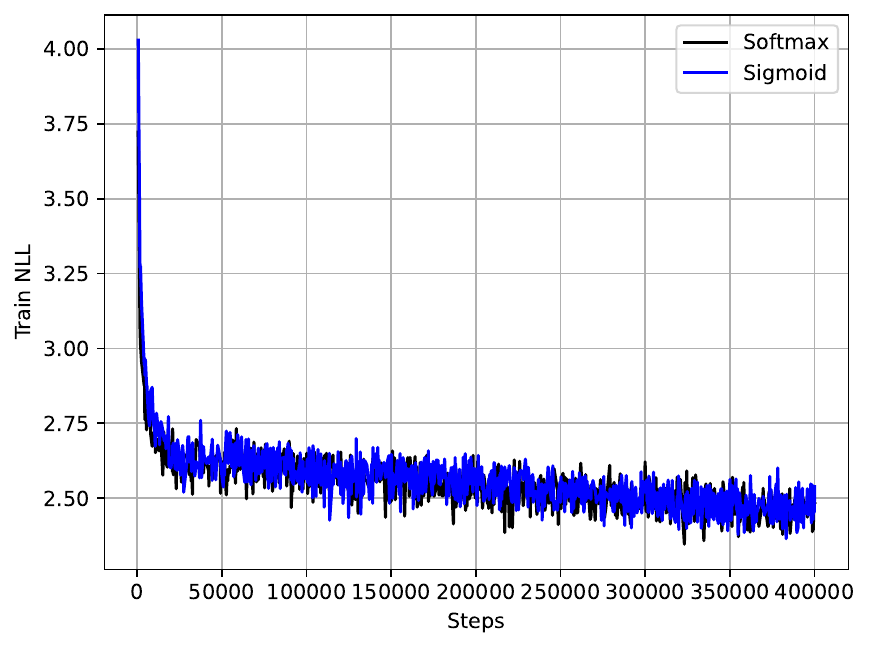
        }
        \captionsetup{justification=centering}
        \caption{
            85M training using $\sigmoidattn$ and $\softmaxattn$ (n = 4096). Training loss matches.
        }
        \label{fig:85m_4k_nll}
    \end{minipage}
    \hfill
    \begin{minipage}{0.46\textwidth}
        \centering        
        \includegraphics[trim={0 0 0 0}, width=\textwidth]{
            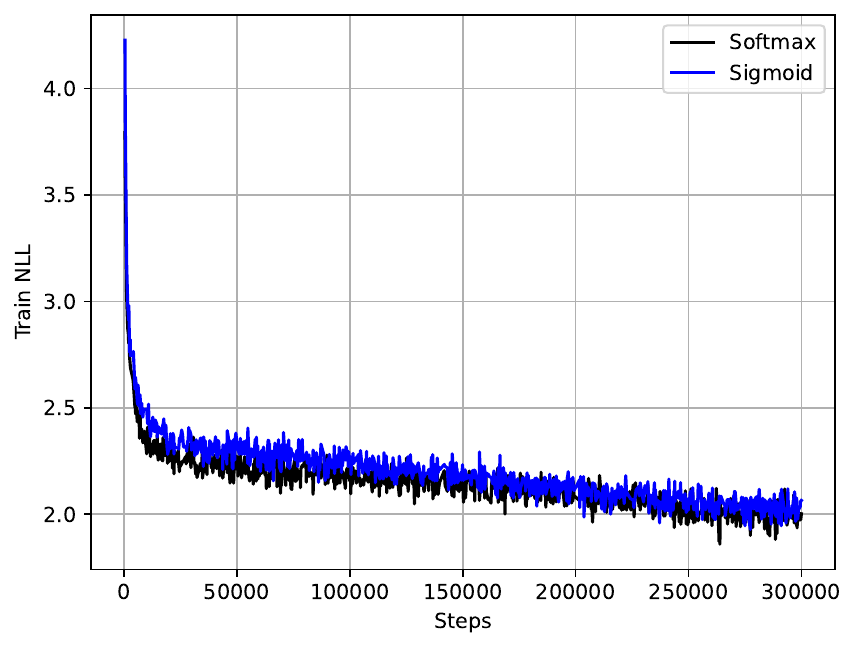
        }
        \captionsetup{justification=centering} 
        \caption{1B training using $\sigmoidattn$ (n = 4096). Higher sequence length with a larger model shows a slightly different loss curve.}
        \label{fig:1b_4k_nll}
    \end{minipage}
\end{figure}
\subsubsection{Gradient Norm}
While a $\sigmoidattn$ based LM using aforementation hyper-parameters has a smooth loss curve, we do see more gradient norm fluctuations. See \cref{fig:lm_grad_norm}, where spikes larger than $0.5$ are not visible in the $\softmaxattn$ equivalent.
\subsubsection{Norm structure}
\label{app:norm_structure}
Due to the slight performance difference observed at 4096 context length when using $\sigmoidattn$ versus $\softmaxattn$, and marginally lower downstream results, we evaluated various norm structures to address potential instabilities (see \cref{tab:lm_norm_ablations}). Some of these structures replace the required attention bias (in this case, column 'Attn. Bias' is 'No'). All use QK-norm with RMSNorm \citep{zhang-sennrich-neurips19}, without LayerScale. We examined pre-norm and hybrid-norm (where we do both pre-norm and normalization of the output of the attention layer following \cite{xiong2020layer}): $\text{norm}(\sigma(\mQ\mK^T / \sqrt{d_{qk}})\mV)$. Post-norm, which normalizes the combined residual data stream, $\text{norm}(x + \sigma(\mQ\mK^T / \sqrt{d_{qk}})\mV)$, is omitted from our analysis as it did not train stably for $\sigmoidattn$. 
\begin{table}[t]
\centering
\caption{Different norm structure ablations for $\sigmoidattn$ with 1B language-modeling.}
\label{tab:lm_norm_ablations}
\begin{sc}
\begin{scriptsize}
\bgroup
\setlength{\tabcolsep}{.35em}
\begin{tabular}{@{}lllllllllllllllll@{}}
\toprule
Model   & \makecell{Seq.\\Len.} &\makecell{Attn.\\Bias} &\makecell{Pos.\\Encod.} &\makecell{Norm} & \makecell{ARC\\Easy} & \makecell{ARC\\Chall.} & \makecell{Hella-\\swag} & Piqa & Sciq & \makecell{Wino-\\grande} & \makecell{Lambada\\OpenAI} & \makecell{TriviaQA\\(1-shot)} & \makecell{WebQS\\(1-shot)} & AVG \\ \midrule
Soft. & 2k & No & ALiBi & Pre & 62.2       &     26.8           &    42.4       &  59.0    &   72.3   &     88.1       &     58.4           &      19.9             &    15.4            &    49.4    \\
Soft. & 2k & No & RoPE & Pre & 64.5       &     30.4           &    43.9       &  61.0    &   71.9 &     88.7       &     59.3           &      21.1             &    15.0            &    50.6    \\
Sigm. & 2k & Yes & ALiBi & Pre &  62.8       &      28.8         &    42.5       &  59.7    &   70.3   &     88.6       &      59.7          &       19.1            &   13.8             &       49.5     \\
Sigm. & 2k & Yes & RoPE & Pre &  62.2       &      26.9         &    41.4       &  57.9    &   71.1   &     87.8       &      57.3          &       17.3            &   12.8             &       48.3     \\
Sigm. & 2k & No & RoPE & Pre &  59.3       &      26.5         &    39.4       &  55.1    &   69.4   &     88.2       &      59.2          &       11.7            &   7.5             &       46.0     \\
\midrule
Soft. & 2k & No & ALiBi & Hybrid & 64.9       &     30.5           &    43.3       &  61.9    &   71.6   &     88.4       &     60.9           &      23.6             &    12.8            &    50.9    \\
Sigm. & 2k & Yes & ALiBi & Hybrid &  60.5       &      26.9         &    42.2       &  59.2    &   70.8   &     89.6       &      57.9          &       17.7            &   13.4             &       48.7     \\
Sigm. & 2k & No & ALiBi & Hybrid &  62.8       &      28.2         &    42.0       &  59.1    &   70.3   &     88.7       &      59.8          &       18.6            &   15.1             &       49.4     \\
\midrule
Soft. & 4k &  No & RoPE & Pre & 63.3       &     29.3           &    43.3       &  58.1    &   71.3   &     86.9       &     58.8           &  20.4             &    15.6            &    49.7     \\
Soft. & 4k & No & ALiBi & Pre & 62.6       &     27.7           &    42.4       &  58.6    &   71.1   &     88.2       &     58.6           &      18.9             &    14.7            &    49.2     \\
Sigm. & 4k & Yes & ALiBi & Pre & 60.5       &      27.3         &    41.3       &  57.8    &   70.5   &     87.0       &      57.6          &       18.9            &   12.6             &       48.2  \\
\midrule
Soft. & 4k & No & RoPE & Hybrid & 64.1       &      27.2         &    43.3       &  61.4    &   71.2   &     88.5       &      60.0          &       21.4            &   15.3             &       50.3  \\
Soft. & 4k & No & ALiBi & Hybrid & 61.7       &      26.8         &    43.4       &  59.4    &   70.6   &     88.6       &      60.8          &       20.5            &   12.9             &       49.4  \\
Sigm. & 4k & No & RoPE & Hybrid & 63.3       &      27.1         &    43.4       &  61.3    &   70.4   &     88.2       &      57.5          &       20.5            &   14.8             &       49.6  \\
Sigm. & 4k & Yes & ALiBi & Hybrid & 63.5       &      28.1         &    43.5       &  60.7    &   70.8   &     88.9       &      59.0          &       20.9            &   16.0             &       50.2  \\
Sigm. & 4k & No & ALiBi & Hybrid & 62.4       &      28.9         &    43.5       &  60.8    &   71.3   &     89.6       &      59.2          &       20.2            &   14.3             &       50.0  \\ \bottomrule
\end{tabular}
\egroup
\end{scriptsize}
\end{sc}
\vspace{-0.4cm}
\end{table}

\subsection{Automatic Speech Recognition}
\label{sec:asr_hps}

\subsubsection{Training Details}
All acoustic models are fed 80 channel log-mel filterbanks with a 25ms sliding window strided by 10ms. 

The transformer-based encoder model has 255M parameters: 1D convolution of kernel 7 and stride 3 followed by CAPE positional embedding if it is used and 36 transformer blocks with pre-LayerNorm, an embedding dimension of 768, 4 heads, 3072 units in the MLP layers.
The model is trained with CTC loss and a character vocabulary, including apostrophe (`). 
In additional experiments, we vary the depth to 12 and 24 layers, and change pre-LayerNorm to post-LayerNorm.

We implemented our own conformer-based encoder model, also trained  with a CTC loss and a character vocabulary.
The conformer model has 104M parameters and consists of 1D convolution of kernel 7 and stride 3 followed by 16 conformer blocks with an embedding dimension of 512, 4 heads, 2048 units in the MLP layers. 
Variational noise is not used and RoPE is used as a relative positional embedding instead of relative sinusoidal positional embedding.

For all models, SpecAugment~\citep{DBLP:conf/interspeech/ParkCZCZCL19} is used for augmentation with 2 frequency masks (max width 30) and 10 time masks (max width 50, ratio 0.1). 
All models are trained with dynamic batching and mixed precision with BF16.
Models are trained with different configurations of optimizers and hyperparameters to have diverse coverage of use-cases. 
\textbf{\textit{We first optimize every configuration for $\softmaxattn$ and then change only attention to the introduced configuration of $\sigmoidattn$ while all other parameters are kept the same.}}
Detailed configurations are shown in~\Cref{tab:asr-training-details}.
We train models until the greedy WER stops improving on the validation sets (\textit{dev-clean, dev-other}) and report final test sets (\textit{test-clean, test-other}) greedy WER without integration of any external language model.

For the bias term $b=-\log n$ in $\sigmoidattn$, we do not use max sequence length as in language model experiments. Instead, for every audio sample we use its own duration as a bias terms resulting into non-trainable bias vector for the minibatch. For experiments with sequence normalization, we also use not the max sequence length in the minibatch but rather the ground truth sample duration to properly normalize encoder attention.

\begin{table}[t!]
\centering
\caption{Training details for the ASR models on LibriSpeech 100h (LS-100) and LibriSpeech 960h (LS-960) for transformers and conformers.}
\label{tab:asr-training-details}
\resizebox{\textwidth}{!}{
\begin{tabular}{@{}llllll@{}}
\toprule
 & Parameter          & Transformer LS-960 & Conformer LS-960 & Transformer LS-100 & Transformer LS-100     \\ \midrule
 & Params             & 255M & 104M & 255M / 170M / 85M        & 255M \\
 & LayerNorm & pre & pre + post & pre & post \\
 & Dropout & 0.1 & 0.1 & 0.3 & 0.3 \\
 & Layer drop & 0.1 & 0.0 & 0.3 & 0.3 \\
 & Training steps      & 400k & 400k & 400k & 500k \\
 & Batch size         & 3.56h & 4.44h & 1.1h & 1.1h \\
 & LR schedule        & step-wise & step-wise & step-wise &  step-wise   \\
 & SpecAugment start & 0k & 10k & 0k & 0k \\
 & LR Warmup Steps    & 64k & 10k & 64k & 64k     \\
 & Peak LR            &  1e-3 & 2e-3 & 0.1 & 0.03   \\
 & LR start decay    & 250k & 250k & 200k & 330k     \\
 & LR decay step  & 50k & 50k & 30k & 50k     \\
 & Optimizer          & AdamW & AdamW & Adagrad & Adagrad    \\
 & Optimizer momentum & 0.9, 0.999 & 0.9, 0.98 & - & - \\
 & Weight decay       & 1e-6 & 1e-6 &   0 & 0     \\
 & Gradient clipping  & 1.0 & 0.5 & 1.0  & 1.0     \\
 & Position encoding  & CAPE / ALiBi / RoPE & RoPE & CAPE & CAPE / ALiBi / RoPE     \\
 & Q/K Norm $\softmaxattn$          & Not Applied & Not Applied &  Not Applied & Not Applied  \\
 & Q/K Norm $\sigmoidattn$          & Applied & Applied &  Not Applied & Applied  \\
 & Num layers         & 36 & 16 & 36 / 24 / 12  & 36      \\
 & Num heads          & 4 & 4 & 4 & 4        \\
 \bottomrule
\end{tabular}
}
\end{table}

To evaluate behaviour for length generalization we use TED-LIUM v3 dataset~\citet{hernandez2018ted} as its validation and test sets have longer audio duration than LibriSpeech: LibriSpeech has in average 10-15s duration, while in TED-LIUM there are audio longer than 30s (the max duration of LibriSpeech).
To perform evaluation on TED-LIUM v3, we combine together validation and test sets of TED-LIUM v3 (we don't use them for training and hyper-parameters search and just perform final evaluation) and split them into 4 datasets according to the duration: 0-10s, 10-20s, 20-30s, and 30s+.

For positional embeddings we use not only CAPE, but change it to AliBi or RoPE. 
As ALiBi was originally introduced for the decoder only models and there is no official adoption of it yet\footnote{See discussion in \url{https://github.com/ofirpress/attention_with_linear_biases/issues/5}.} for the encoder models (without causal masking), we follow the best practices found in \url{https://iclr-blogposts.github.io/2024/blog/alibi-mlm/} of nonsymmetric ALiBi with different slopes instead of symmetric version used by~\citep{lee2022littlebird}.

\begin{table}[t!]
\centering
\caption{Word error rate (\%) on LibriSpeech dev/test sets and TED-LIUM v3~\citep{hernandez2018ted} (``TED'', joint validation and test sets with split according to audio duration) for pre-LayerNorm transformer (255M~/ 170M / 85M params) with CAPE and with either $\softmaxattn$ or $\sigmoidattn$ (w/ LayerScale, w/o QK norm, w/ $b=-\log n$) trained on LibriSpeech 100h data (average duration is 10-15s). Hyper-parameters can be found in \Cref{tab:asr-training-details}.}
\label{tab:asr-pre-100h}
\begin{center}
\begin{scriptsize}
\begin{sc}
\resizebox{\columnwidth}{!}{
\begin{tabular}{lcrrrrrrrr}
\toprule
 attn & \# layers & dev-clean & test-clean & dev-other & test-other & ted 0-10s & ted 10-20s & ted 20-30s & ted 30s+  \\
\midrule 
softmax &  36 & 6.7 & 7.1 & 20.0 & 20.4 & 26.4 & 22.4 & 23.3 & 21.8 \\
sigmoid & 36 & 7.0 & 7.3 & 20.3 & 20.5 & 26.2 & 23.4 & 23.6 & 21.8 \\
\,\,\,\, $b=0$  & 36 & 6.8	& 7.1	& 19.8	& 20.3
\\
\midrule
softmax &  24 & 6.4 & 6.8 & 20.2 & 20.5 & 25.4 & 22.1 & 23.3 & 21.8 \\
sigmoid &  24 & 7.1 & 7.3 & 21.0 & 21.3 &  26.6 & 23.3 & 24.0 & 22.0 \\
\,\,\,\, $b=0$ & 24 & 6.7	& 6.9	& 20.2	& 20.7
 \\
\midrule
softmax &  12 & 8.2 & 8.7 & 25.0 & 25.4 & 29.0 & 25.6 & 27.1 & 27.4 \\
sigmoid & 12 & 8.3 & 8.7 & 24.8 & 25.2 & 29.0 & 25.7 & 26.3 & 25.5 \\
\,\,\,\, $b=0$ & 12 & 8.7	& 8.5	& 24.4	& 24.7
\\
\bottomrule
\end{tabular}
}
\end{sc}
\end{scriptsize}
\end{center}
\end{table}

\begin{table}[t!]
\centering
\caption{Word error rate (\%) on LibriSpeech dev/test sets for post-LayerNorm transformer (255M) with either $\softmaxattn$ (w/o QK norm) or $\sigmoidattn$ (by default w/ LayerScale, w/ QK norm, w/ $b=-\log n$) trained on LibriSpeech 100h data. Hyper-parameters can be found in \Cref{tab:asr-training-details}.}
\label{tab:asr-post-100h}
\begin{center}
\begin{scriptsize}
\begin{sc}
\begin{tabular}{lcrrrr}
\toprule
 attn & PE & dev-clean & test-clean & dev-other & test-other \\
\midrule 
softmax & CAPE & 6.4 & 6.5 & 18.4 & 18.2 \\
\,\,\,\, + QK norm & & 6.1 & 6.3 & 18.2 & 18.1  \\
sigmoid & & 8.0 & 8.4 & 22.7 &  22.7\\
\,\,\,\, - QK norm & & 7.5 & 7.9 & 22.1 & 27.6 \\
\,\,\,\, - LayerScale & & \multicolumn{4}{c}{unstable, gradient norm and loss spikes} \\
\,\,\,\, - QK norm - LayerScale & & 6.5 & 6.9 & 19.9 & 20.1 \\ 
sigmoid ($b=-10$, learnable)  & & 8.7 & 9.4 & 23.5 & 24.0 \\
\midrule
softmax & RoPE & 6.6 & 6.9 & 18.3 & 18.5 \\
sigmoid & & 6.8 & 7.1 & 20.8 & 20.8\\
sigmoid ($b=-10$, learnable)  & & 8.7 & 9.4 & 23.5 & 24.0 \\
\midrule
softmax & AliBi & 6.4 & 6.9 & 18.3 & 18.3 \\
sigmoid & & 6.9 & 7.2 & 20.8 & 21.1 \\
sigmoid ($b=-10$, learnable)  & & 6.8 & 7.1 & 20.4 & 20.5\\
\bottomrule
\end{tabular}
\end{sc}
\end{scriptsize}
\end{center}
\end{table}

\begin{table}[t!]
\centering
\caption{Word error rate (\%) on LibriSpeech dev/test sets and TED-LIUM v3~\citep{hernandez2018ted} (``TED'', joint validation and test sets with split according to audio duration) for conformer (104M) with RoPE and with either $\softmaxattn$ or $\sigmoidattn$ (w/ LayerScale, w/ QK norm, w/ $b=-\log n$) trained on LibriSpeech 960h data (average duration is 10-15s). Hyper-parameters can be found in \Cref{tab:asr-training-details}.}
\label{tab:asr-conformer}
\begin{center}
\begin{scriptsize}
\begin{sc}
\resizebox{\columnwidth}{!}{
\begin{tabular}{lrrrrrrrr}
\toprule
 attn & dev-clean & test-clean & dev-other & test-other & ted 0-10s & ted 10-20s & ted 20-30s & ted 30s+  \\
\midrule 
softmax & 2.2	&2.5	& 5.4	&5.6	&13.0 & 11.1	&13.2	&7.1 \\
sigmoid  & 2.3	&2.5	&5.6	&5.8	&13.5	&10.8	&13.3	&10.2 \\
sigmoid ($b=-10$, learnable)  &  2.4&	2.7&	5.8&	5.8&	12.9&	11.1&	14.1&	54.9  \\
\bottomrule
\end{tabular}
}
\end{sc}
\end{scriptsize}
\end{center}
\end{table}

\begin{figure}[t]
    \centering
    
        \centering
        \includegraphics[width=0.6\textwidth]{
            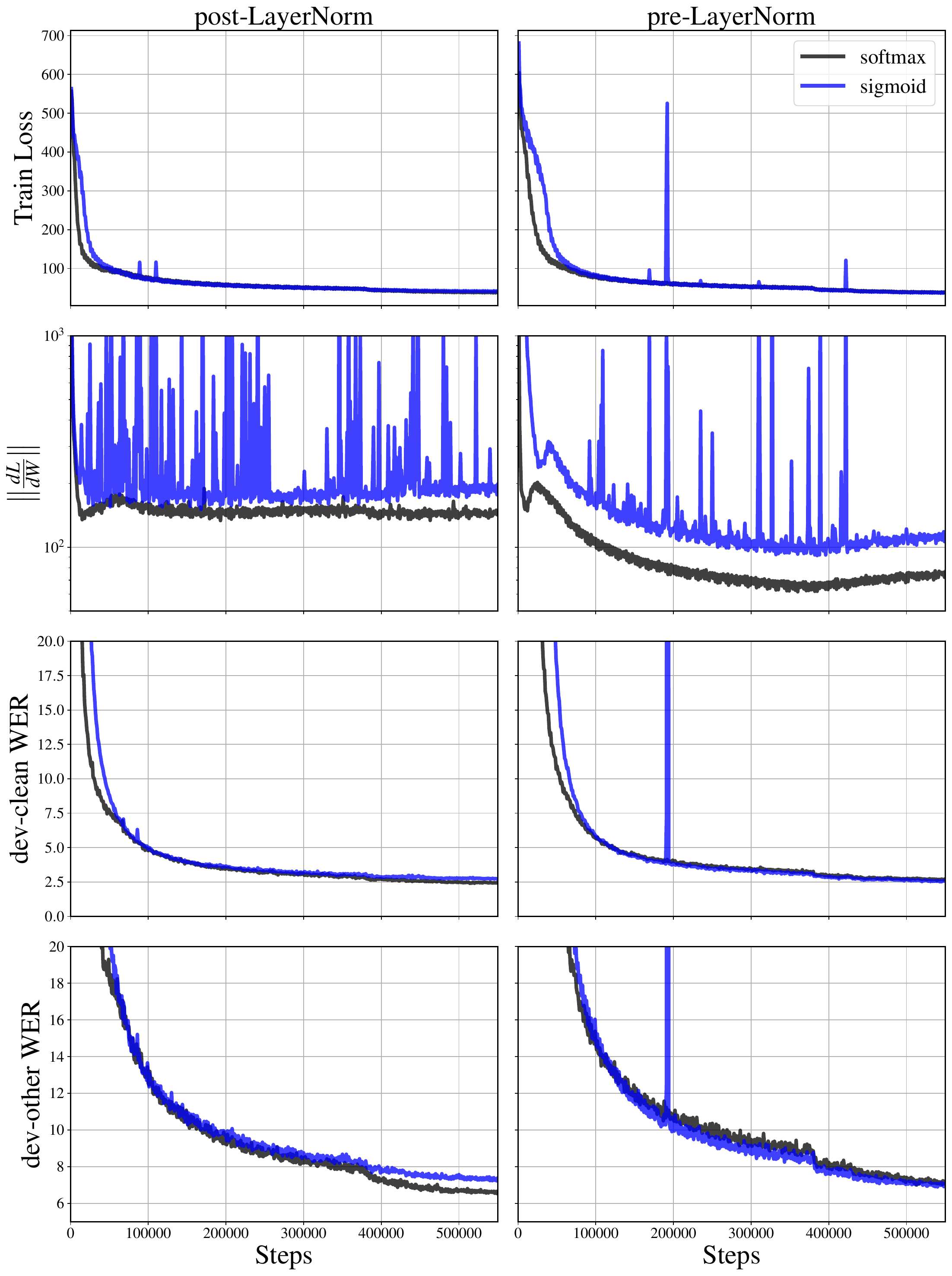
        }
        \caption{
            ASR Transformer model (255M) training with post-LayerNorm (left) and pre-LayerNorm (right) on LibriSpeech 960h with $\sigmoidattn$ (w/ bias term, $b=0$, w/o QK norm, w/ LayerScale) or with $\softmaxattn$. Huge gradient norms and training loss spikes are observed for $\sigmoidattn$ which can result in worse final model performance hence models for $\sigmoidattn$ are unstable.}
        \label{fig:asr-spikes}
        \vspace{-0.4cm}
\end{figure}

\begin{figure}[ht]
  \begin{minipage}{0.58\textwidth}
    \centering
    \includegraphics[width=0.6\linewidth]{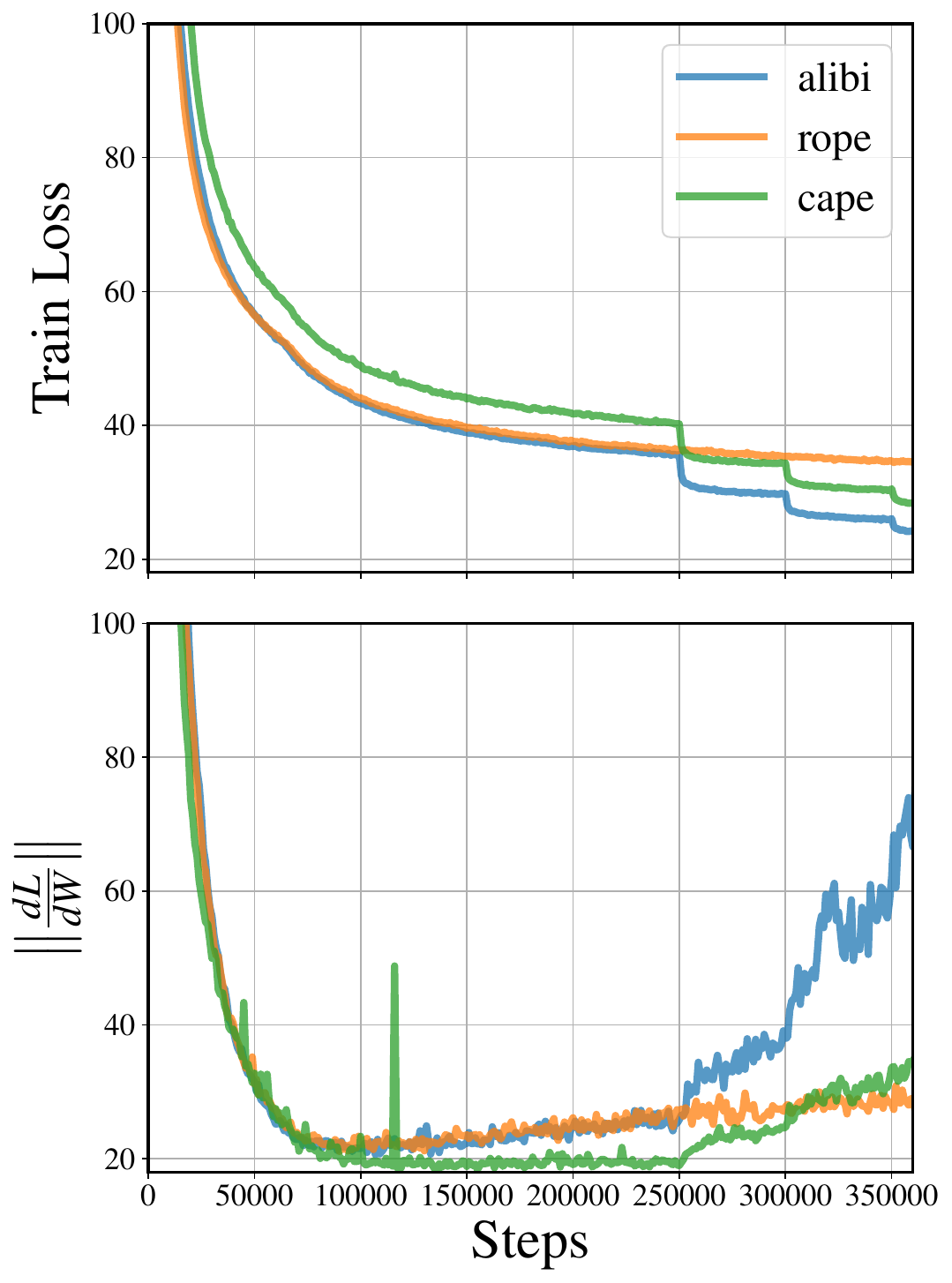}
  \end{minipage}%
  \hfill
  \begin{minipage}{0.4\textwidth}
    \caption{ASR Transformer model (255M) training with pre-LayerNorm on LibriSpeech 960h with $\sigmoidattn$ (w/ bias term, $b=-\log n$, w/ QK norm, w/ LayerScale) and different positional embeddings CAPE, RoPE, ALiBi. The bias $b$ is able to stabilize $\sigmoidattn$ training: smooth training loss and only marginal rare spikes in gradient norms are observed.}    
    \label{fig:asr-logn}
  \end{minipage}  
\end{figure}

\subsubsection{Results and Ablations}
\label{sec:asr_appendix_ablations_and_results}
Initial investigation on post-LayerNorm and pre-LayerNorm transformers on both LibriSpeech 100h and 960h revealed that $\sigmoidattn$ without any bias is unstable resulting in huge and frequent gradient norm and training loss spikes throughout the training which in turn result in spikes of validation and test WER, see~\Cref{fig:asr-spikes}. Neither LayerScale nor QK norm were able to stabilize the training, though we did not observe any model divergence.

Further experiments with bias term in the $\sigmoidattn$ definition for post-LayerNorm transformers on LibriSpeech 100h reveal that training is now stable (only few marginal spikes in gradient norm occur, while train loss is smooth all the time). However, both LayerScale and QK norm restrict model capacity thus not matching $\softmaxattn$. Moreover, some combination of them is needed for the stable training, though w/o both of them we got the best performance for $\sigmoidattn$ (still behind $\softmaxattn$), see~\Cref{tab:asr-post-100h}.
We believe, further adaptation and deeper investigation is needed for $\sigmoidattn$ and post-LayerNorm, though recent advances in machine learning do not use post-LayerNorm models due to high training instability even for $\softmaxattn$.

Switching to pre-LayerNorm transformers and varying the depth of the models lead to stable training with $\sigmoidattn$ and bias term $b=-\log n$ with few (2-5 times) spikes in the gradient norm and smooth loss. In this case, $\sigmoidattn$ matches results for $\softmaxattn$ and they both generalize to TED-LIUM data similarly, see~\Cref{tab:asr-pre-100h}. If the bias term is removed, $\sigmoidattn$ can still match $\softmaxattn$ but large spikes in gradient norm and loss can occur.

Finally, we experiment with a conformer model, in \Cref{tab:asr-conformer}. Again, we found that bias term $b=-\log n$ stabilizes training. The learnable $b=-10$ works though we see significant gradient norm spikes while the train loss remains smooth. Besides,  $b=-\log n$ generalizes well to longer sequences while learnable $b=-10$ fails to do so with RoPE for conformer. Overall, $\sigmoidattn$ is able to match $\softmaxattn$ having stable training with  $b=-\log n$.

In experiments with different variants of bias term for $\sigmoidattn$, the bias $b=-\log n$ is found to be the most stable (only few marginal gradient norm spikes are observed with the train loss being smooth) and it provides similar performance as $\softmaxattn$ in most settings. 
The source of instability is coming from the larger attention output norms (80k for CAPE, 40k for RoPE and 20k for AliBi while being 200 for $\softmaxattn$). This happens due to high attention weight of every token which can be biased towards zero with a bias term in $\sigmoidattn$ definition.
Preliminary results to connect this to the local attention property needed at the beginning of the training for stable training failed, as local attention did not converge well at all (it is deactivated after some initial training).

To fully benefit from the improved throughput of \textsc{FlashSigmoid}, for the bias term $b=-\log n$ in $\sigmoidattn$, we experimented with configuration when the maximum audio duration in the minibatch is used as $n$ resulting into non-trainable bias scalar which changes between minibatches as we use dynamic batching. Comparison between the bias vector with per sample own duration normalization and the bias scalar as maximum duration in the minibatch is shown in ~\Cref{tab:asr-results-ext}: final model performance is similar and stability is same (only 2-3 minor spikes in CAPE for gradient norms are observed). Thus, per batch maximum audio duration can be used with $b=-\log n$ as the final configuration.

We also experimented with hybrid-norm (see Table 16) to check if it is able to stabilize the attention magnitudes as well as gradient norms. 
We did ablation with configuration similar to Table 16 with the following changes: LayerScale after attention is replaced to LayerNorm, only RoPE is used for positional embedding; we either keep or remove QK-norm and we either keep or remove LayerScale in MLP part of transformer block. 

First, for all variants we observe that training loss and gradient norms are smooth without any spikes during training while we see abnormally large attention activations compared to all prior experiments. Second, while we observe that QK-norm or its removal behave similarly, the LayerScale on top of MLP output is necessary to get performance on par with $\softmaxattn$ or with $\sigmoidattn$ with bias term.

\begin{table}[t!]
\centering
\caption{Word error rate (\%) on LibriSpeech dev/test sets and TED-LIUM v3~\citep{hernandez2018ted} (``TED'', joint validation and test sets split according to  duration) for transformer (255M params) with either $\softmaxattn$ or $\sigmoidattn$ (LayerScale and QK norm are used with $b=-\log n$) trained on LibriSpeech 960h data (mean duration is 10-15s). Hyper-parameters are in~\cref{sec:asr_hps}. H-Norm corresponds to hybrid-norm, no LS-Attn corresponds to removing the LayerScale from the attention outputs, and no LS corresponds to removing the LayerScale from both the attention and MLP outputs.}
\label{tab:asr-results-ext}
\begin{center}
\begin{scriptsize}
\begin{sc}
\resizebox{\columnwidth}{!}{%
\begin{tabular}{lcrrrrrrrr}
\toprule
 attn & PE & dev-clean & test-clean & dev-other & test-other & ted 0-10s & ted 10-20s & ted 20-30s & ted 30s+  \\
\midrule 
softmax & \multirow{3}{*}{CAPE} & 2.2 & 2.3 & 5.6 & 5.7 & 12.4 & 10.5 & 11.9 & 9.1 \\
 sigmoid &  & 2.2 & 2.4 & 5.2 & 5.5 & 12.4 & 10.3 & 12.3 & 9.7 \\
 sigmoid, $b=-\log(\max_{batch} n)$ &  & 2.1 & 2.3 & 5.2 & 5.3 & 12.2 & 10.6 & 12.0 & 9.3 \\
\midrule
softmax & \multirow{6}{*}{RoPE} & 2.2 & 2.2 & 5.4 & 5.5 & 12.7 & 10.6 & 12.8 & 9.5 \\
 sigmoid &  & 2.0 & 2.3 & 5.2 & 5.4 & 12.3 & 10.1 & 12.3 & 8.6 \\
 sigmoid, $b=-\log(\max_{batch} n)$ & & 2.1 & 2.3 & 5.0 & 5.1  & 12.3 & 10.1 & 12.1 & 10.4 \\
 sigmoid (h-norm), no QK-norm, no LS-attn & & 2.1 &	2.2 &	5.0	& 5.0	& 11.8	& 10.2	& 12.3	& 10.8 \\
 sigmoid (h-norm), no LS-attn & & 2.1 &	2.3 &	5.0	& 5.1	& 12.0	& 10.2	& 12.4	& 11.4 \\
 sigmoid (h-norm), no QK-norm, no LS & & 2.2 &	2.3 &	5.6	& 5.6	& 13.2	& 10.9	& 13.5	& 11.5 \\
\midrule
 softmax & \multirow{3}{*}{ALiBi} & 2.1 & 2.2 & 5.3 & 5.4 & 12.3 & 10.7 & 12.1 & 8.6 \\
 sigmoid &  & 2.1 & 2.3 & 5.0 & 5.1 & 12.3 & 10.5 & 12.6 & 9.1 \\
 sigmoid, $b=-\log(\max_{batch} n)$ &  & 2.0 & 2.3 & 5.2 & 5.2 & 12.3 & 10.5 & 11.9 & 10.2 \\
\bottomrule
\vspace{-0.4cm}
\end{tabular}
}
\end{sc}
\end{scriptsize}
\end{center}
\end{table}

\subsection{Simple Experiments}
\label{sec:appendix_simple_experiments}

\subsubsection{k--Summation Problem Definition}
\label{sec:a_se_prob_def}

Here we look at a synthetic, simple task in order to investigate the behavior of softmax and sigmoid attention activations.  The problem chosen is to minimize the MSE loss of a $\mathbb{R}^{n} \rightarrow \mathbb{R}$ target function.  In the first half of each input are samples from a $\mathcal{N}(0,1)$ distribution, and the second half is a $k$-hot binary vector indicating which values in the first half to sum.  

The results presented here are for the $n=40$ problem with various values for $k$.  Where a transformer is used, the transformer is a single layer to aid visualization.  In all cases (unless noted otherwise), the optimizer is Adam with a constant learning rate of 0.001, and the training data is continuously generated to preclude over-fitting.

A few examples for $n=10$ (not drawn from $\mathcal{N}(0,1)$) are shown below.  Inputs in the second half of the input are show in {\color{orange} orange} only as a visual aid.

\begin{table}[h]
    \centering
    \begin{tabular}{c}
        \texttt{1 2 3 4 5 {\color{orange} 0 0 0 0 1 }$\rightarrow$ 5} \\
        \texttt{1 2 3 4 5 {\color{orange} 1 0 0 0 1 }$\rightarrow$ 6} \\
        \texttt{8 1 2 0 5 {\color{orange} 0 1 1 1 0 }$\rightarrow$ 3} \\
        \texttt{2 0 2 2 2 {\color{orange} 1 1 0 1 0 }$\rightarrow$ 4} \\
    \end{tabular}
    \label{tab:k_sum_examples}
\end{table}

\subsubsection{Comparison to Softmax}
\label{sec:a_se_compare}

In \Cref{fig:final_loss_k_summation}, we see the performance of three architectures on the $k$-summation problem as $k$ increases.  The sigmoid activated transformer has similar scaling to the softmax activation.

\begin{figure}[h]
    \centering
    \includegraphics[width=0.6\textwidth]{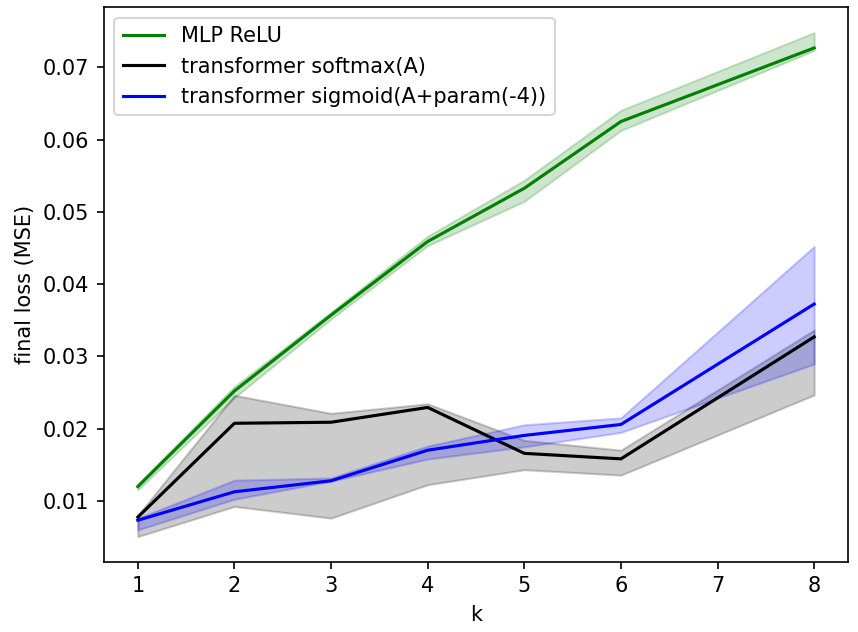}
    \caption{Final loss is shown after training convergence as k-summation problem complexity increases. The ReLU MLP has two hidden layers (900, 300) for 307k parameters, while the transformer has an embedding dimension of 120, 8 heads, and an MLP ratio of 4, giving 187k parameters. The $\sigmoidattn$ is applied after a learned offset initialized to -4, \texttt{A+param(-4)}.}
    \label{fig:final_loss_k_summation}
\end{figure}

\subsubsection{Attention Evolution}
\label{sec:a_se_evo}

In \Cref{fig:attn_evolve,fig:attn_metric_evolve}, forty samples are used to monitor the single head, single layer post-activation attention matrix as training progresses.   In \Cref{fig:attn_evolve}, the distribution of values is visualized over time; note the sigmoid attention is more variable but reaches comparable values at convergence.  The main difference at convergence is that the sigmoid has fewer high magnitude values than softmax indicating a more distributed attention.  

\begin{figure}[h]
  \centering
  \begin{minipage}[b]{0.45\textwidth}
    \includegraphics[width=\textwidth]{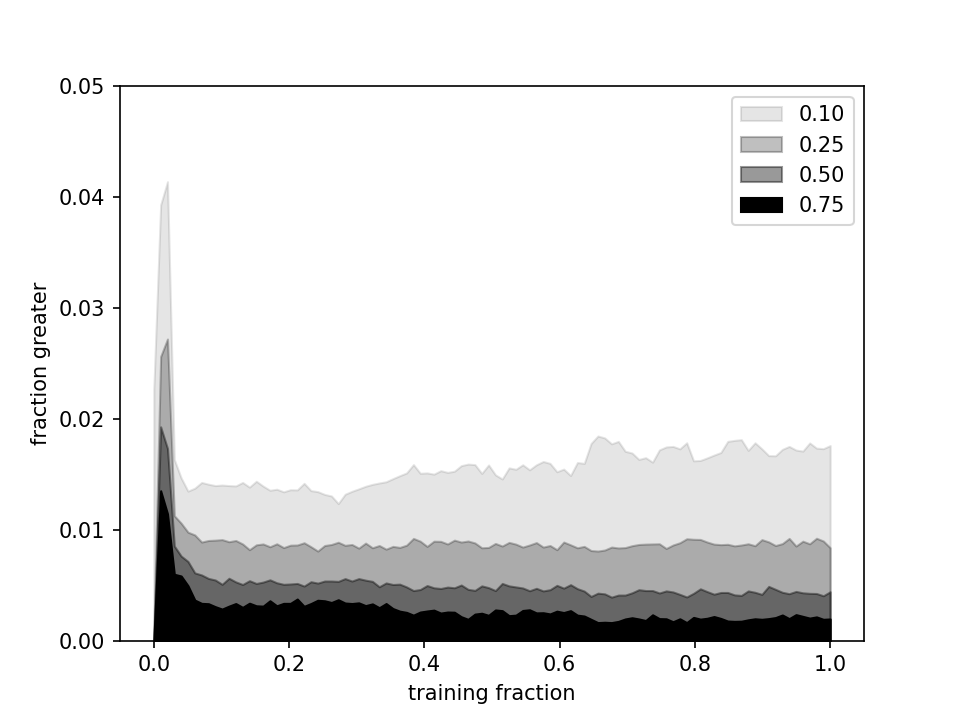}
    \captionsetup{labelformat=empty}
    \caption{Softmax}
    \addtocounter{figure}{-1}
  \end{minipage}
  \hfill
  \begin{minipage}[b]{0.45\textwidth}
    \includegraphics[width=\textwidth]{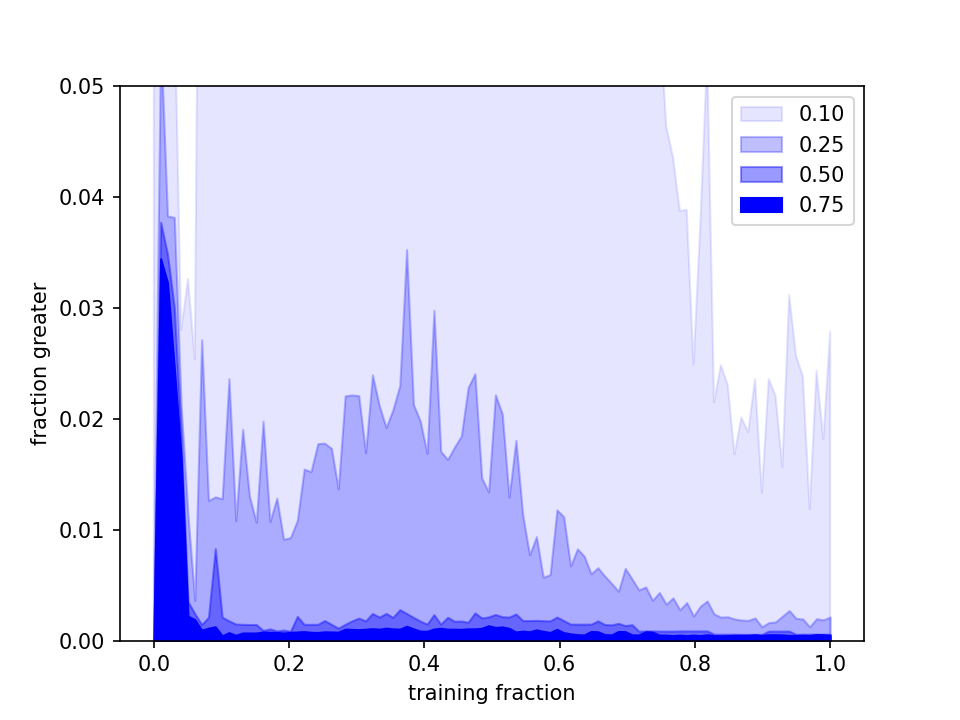}
    \captionsetup{labelformat=empty}
    \caption{Sigmoid}
    \addtocounter{figure}{-1}
  \end{minipage}
  \caption{The post-activation attention evolves during training on the $k=1, n=40$ summation problem.  The model has one head to simplify the visualization. Forty repeated test samples are used.}
  \label{fig:attn_evolve}
\end{figure}

In \Cref{fig:attn_metric_evolve}, metrics on the post-activation attention matrices are used and show comparable behavior in the first half of training.  
In the second half of training, the $\sigmoidattn$ can be seen to reduce in norm and in sparsity. (see following discussion of \Cref{fig:attn_by_sample} for further insights).

\begin{figure}[h]
  \centering
  \begin{minipage}[b]{0.45\textwidth}
    \includegraphics[width=\textwidth]{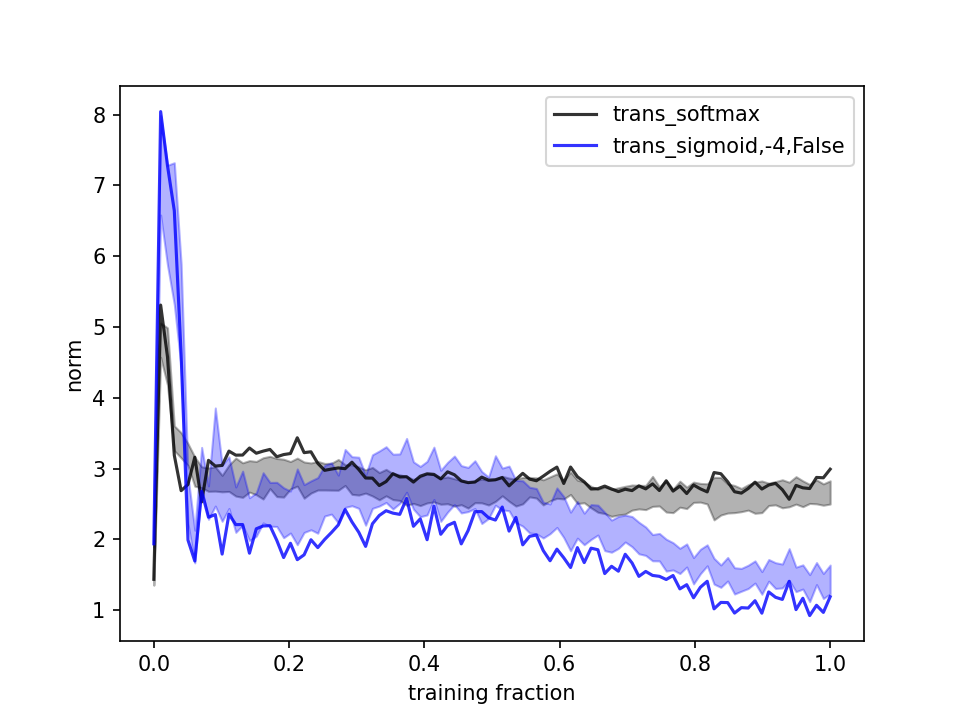}
    \captionsetup{labelformat=empty}
    \caption{Norm}
    \addtocounter{figure}{-1}
  \end{minipage}
  \hfill
  \begin{minipage}[b]{0.45\textwidth}
    \includegraphics[width=\textwidth]{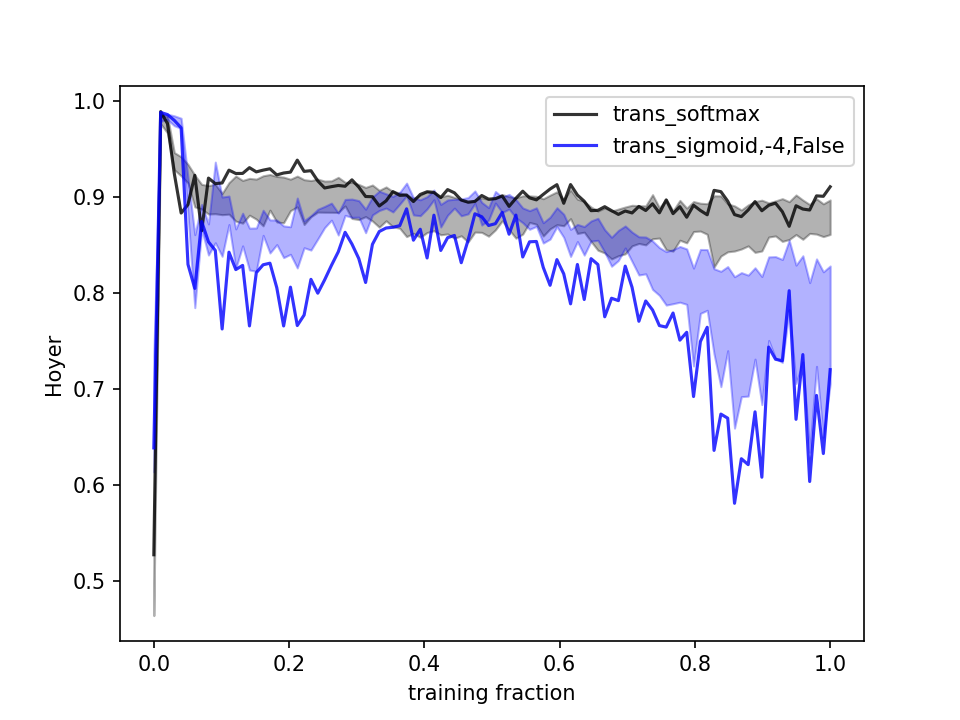}
    \captionsetup{labelformat=empty}
    \caption{Hoyer Sparsity}
    \addtocounter{figure}{-1}
  \end{minipage}
  \caption{ Metrics on the post-activation attention evolve during training on the $k=1, n=40$ summation problem.  The model has one head to simplify the visualization. Quartiles and mean from 40 repeated test samples are shown. On the right, the Hoyer Sparsity~\citep{hurley2009comparing} is used to measure the change in sparsity as training progresses: $\text{Hoyer} := \left(\sqrt{n}-\frac{\sum_j c_j}{\sqrt{\sum_j c_j^2}}\right)(\sqrt{n}-1)^{-1}$.
  }
  \label{fig:attn_metric_evolve}
\end{figure}

In \Cref{fig:attn_by_sample}, we see post-activation attention values for eight samples at training progresses.  The most notable difference between the activations is, that by the end of training, the $\sigmoidattn$ is less sparse in the $\mathcal{N}(0,1)$ self-attention in the upper-left quadrant.  We can see that softmax tends to produce sparser values (as it is designed to) while sigmoid controls the magnitude and location of peak attention independently, leading to a less sparse attention at the end of training.

\begin{figure}[h]
    \centering
    \includegraphics[width=\textwidth]{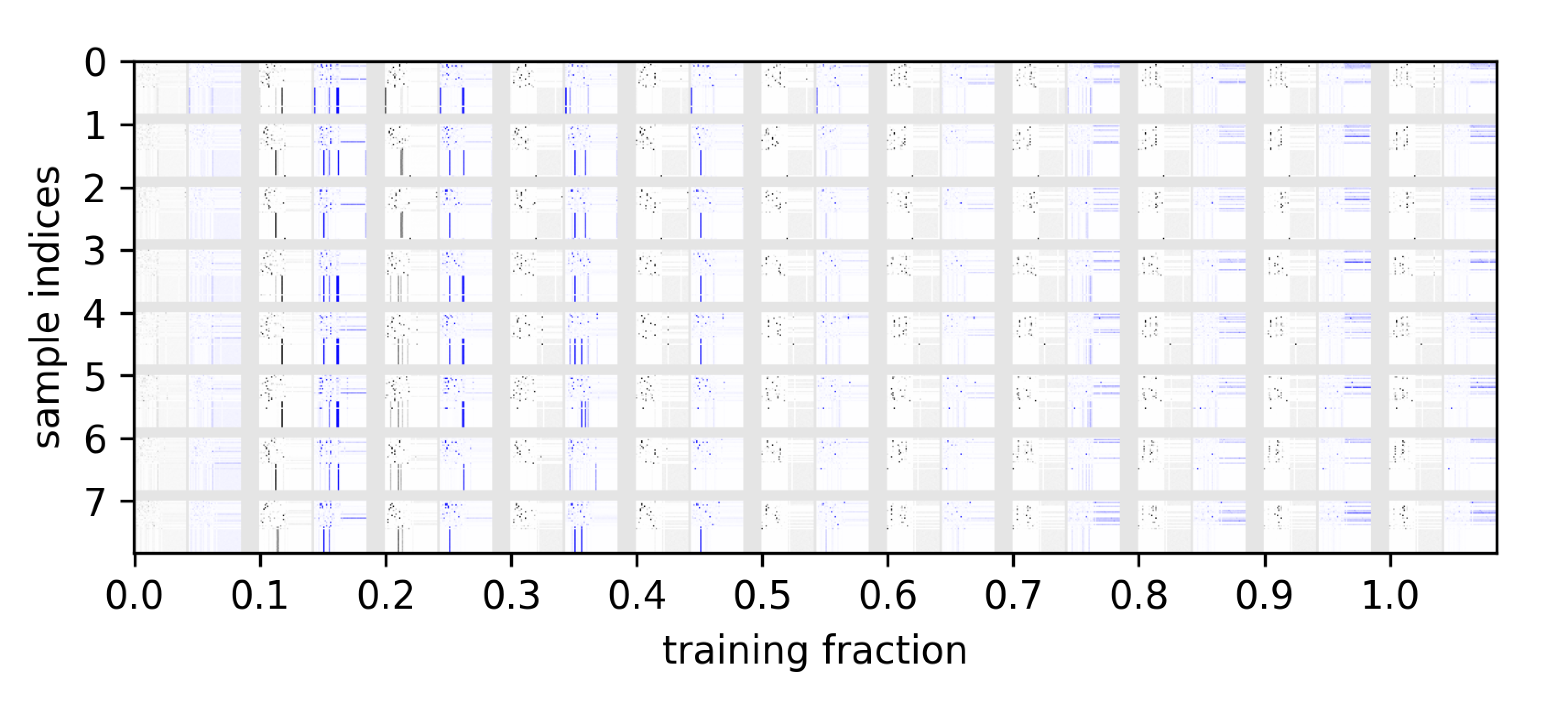}
    \caption{For 8 samples, the post-activation attentions is visualized as training progresses on the $k=1, n=40$ summation problem.  The model has one head to simplify the visualization.  The attention is shown in pairs for each sample with softmax attention is in black and sigmoid is in blue.  A $2 \times 2$ block structure is evident in both cases, resulting from each halve of the input containing different information.}
    \label{fig:attn_by_sample}
\end{figure}

\subsubsection{Pair Repeat Problem}
\label{sec:a_se_pair_repeat_prob}

We define a synthetic task of identifying if the first two symbols in a sequence repeat.  The symbols, $s_i$ below come from a fixed vocabulary of size $K$, and the repeat location (when present) is uniformly distributed in the sequence.

\begin{center}
\begin{math}
f(s_0, s_1, s_2, ..., s_N) = \begin{cases}
1, & \text{ if } \exists \ n>1 \mid (s_0, s_1) = (s_n, s_{n+1}), \\
0 & \text{ otherwise}
\end{cases}
\end{math}
\end{center}

A simple two layer transformer is trained on this problem.  The model has an embedding dimension of 160, MLP ratio of 4, QK norm, and layers with eight heads.  The results for different model architectures are shown in Figure \Cref{fig:repeat_seq_results}.  The maximum input length is 22, $K=9$, shorter lengths are padding with value $K$, and the training set only contains lengths 14 and 15.  A cosine learning rate schedule with 5\% linear warmup and a maximum learning rate of 1e-3 is used with the Adam optimizer.  

In this result, we see the sigmoid activation has higher data efficiency and similar fall-off in the out of distribution cases.  From shorter runs, we estimate that the softmax network would fit the training with 4--5x more data.  Our conjecture is that the two layer transformer more easily learns the pair finding task with sigmoid because softmax is biased to focus on single values, though it is unclear why multiple heads are not able to compensate for this proposed cause in the softmax case.

\begin{figure}[h]
    \centering
    \includegraphics[width=0.7\textwidth]{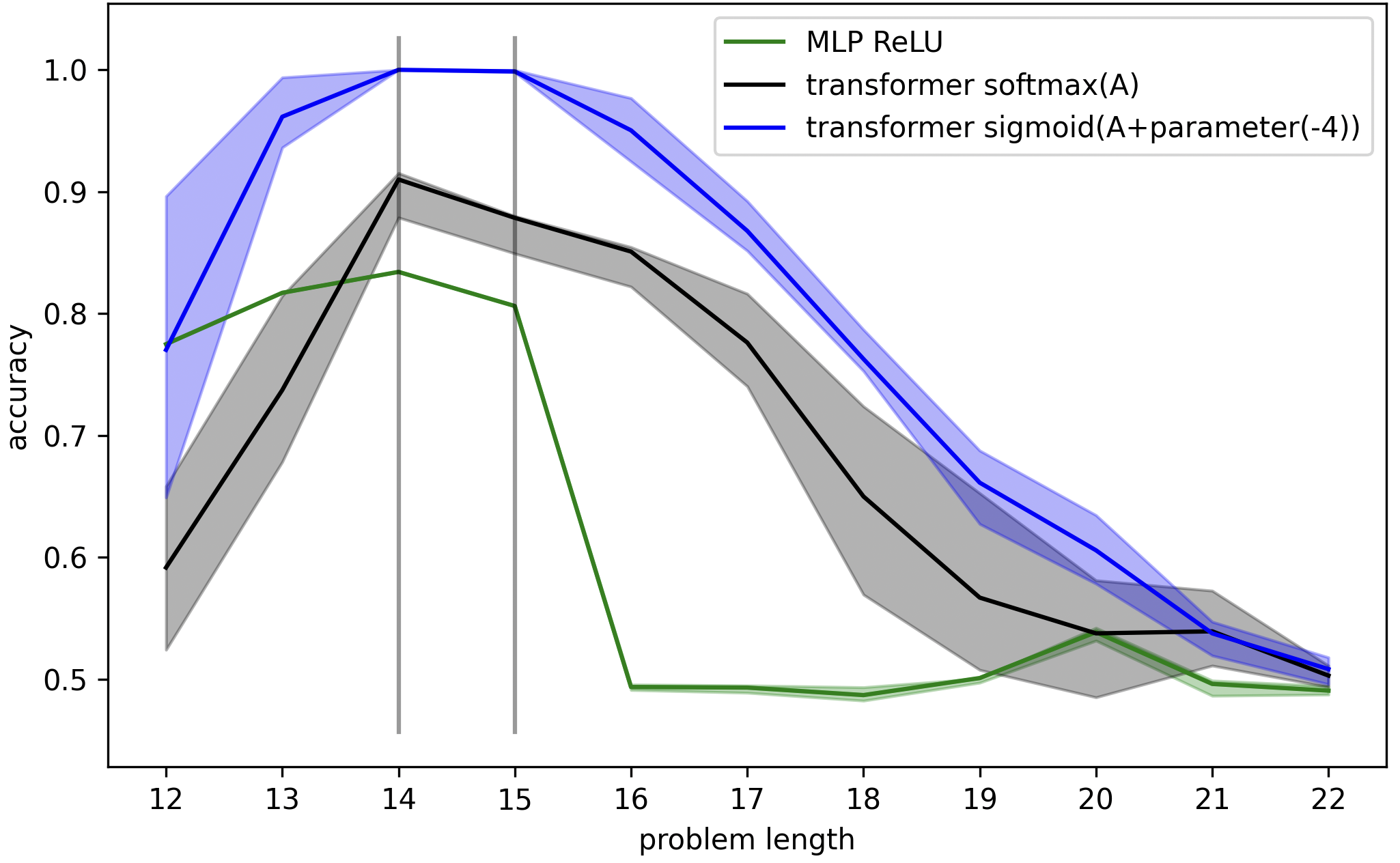}
    \caption{Validation accuracy for out of distribution sequence lengths is shows after 5M samples of training; trained lengths are shown with vertical lines.  Quartiles and means are shown from six trials.  The MLP has two hidden layers, ReLU activation, and a similar number of parameters.  The sigmoid transformer has a learned offset initialized to -4.}
    \label{fig:repeat_seq_results}
\end{figure}

\subsection{Practitioner's Guide}
\label{sec:practioners_guide}
In \Cref{tab:practioners} we summarize recommended settings for practitioners who aim to use $\sigmoidattn$ for training in their respective domains / learning scenarios. While each setting has fully enumerated hyper-parameters listed in the Appendix, we highlight some sane $\sigmoidattn$ choices below. 
\begin{table}[ht!]
\centering
\caption{Simplified recipe for different domains and tasks with Sigmoid Attention. S/C/A/R refers to using any of SinCos, CAPE, ALiBi or RoPE positional encoding methods.}
\label{tab:practioners}
\begin{center}
\begin{scriptsize}
\begin{sc}
\resizebox{\columnwidth}{!}{
\begin{tabular}{lcrrrrrr}
\toprule
 Domain & Objective & Model Size & Pos Embed & QK Norm & LayerScale & Sigmoid Bias & Norm Strategy  \\
\midrule 
Vision & Supervised & 87M & Learnable & Yes & Yes & No & Pre-Norm \\
 & BYOL & 87M & Learnable & Yes& Yes & No & Pre-Norm \\
 & SimCLR & 87M & SinCos & Yes& Yes & No & Pre-Norm \\
 & MAE & 304M & Learnable & Yes & Yes & Yes & Pre-Norm \\
\midrule
ASR & Supervised (CTC) & 100M-250M & S/C/A/R & Yes & Yes & Yes & Pre-Norm \\
& Supervised (CTC) & 100M-250M & RoPE & No & Yes & No & Hybrid-Norm \\
\midrule
AR Language & Next-token (<=2k seq len) & 1B & ALiBi  & Yes & No & Yes & Pre-norm \\
 & Next-token (<=2k seq len) & 1B & ALiBi  & Yes & No & No & Hybrid-norm \\
 & Next-token (>2k seq len) & 1B & ALiBi  & Yes & No & No & Hybrid-norm \\
\bottomrule
\end{tabular}
}
\end{sc}
\end{scriptsize}
\end{center}
\end{table}
\paragraph{Stabilizing larger models beyond sequence length 2048:} We propose a non-learned scalar bias to mitigate large attention norms with $\sigmoidattn$ (\Cref{app:sigmoid_bias}), but observe instabilities at sequence length $n=4096$ for autoregressive language modeling (\Cref{sec:llm}). Hybrid-norm (without learnable affine parameters) resolves these instabilities (\Cref{tab:lm_norm_ablations}). 
\begin{table}[h]
\centering
\begin{tabular}{cc}
\toprule
Post-norm & Hybrid-norm \\
\hline
$\text{norm}(x + \sigma(\mQ\mK^T / \sqrt{d_{qk}})\mV)$ & x + $\text{norm}(\sigma(\mQ\mK^T / \sqrt{d_{qk}})\mV)$ \\
\bottomrule
\end{tabular}
\end{table}
Hybrid-norm differs from post-norm, which normalizes the combined residual data stream and attention block output. Hybrid-norm is used in models such as Grok-1 \citep{grok1} and frameworks such as Praxis \citep{praxis} under the normalization strategy "primer\_hybrid". When both $\softmaxattn$ and $\sigmoidattn$ use hybrid-norm, we observe similar kernel speedup times as highlighted in \Cref{sec:FlashSigmoidHardwareAwareImplementation}. 
However, with LayerNorm~\citep{lei2016layer} only for $\sigmoidattn$, a token length of 10,000 is needed to achieve a performance gain of $\sim 5.04\%$ for full self-attention and a token length of 1024 is needed to achieve a performance gain of $8.36\%$ for causal self-attention on H100 GPUs. 
For LayerNorm (with and without affine terms), we summarize approximate regimes for positive throughput gains in \Cref{fig:appendix:flash-sigmoid-plus-layer-normalization-torchcompile}.

\begin{table}[htbp]
    \tiny
    \centering
    \begin{sc}
    \resizebox{\columnwidth}{!}{%
    \begin{tabular}{@{\extracolsep{4pt}}ccccc}
        \toprule
            \multirow{3}{*}{Attention Type}
            &
            \multicolumn{4}{c}{
                \textsc{FlashSigmoid} with LayerNorm versus \textsc{FlashAttention2 comparison}
            }
        \\
        \cmidrule{2-5}
            &
            \multicolumn{2}{c}{A100} 
            &
            \multicolumn{2}{c}{H100} 
        \\
        \cmidrule{2-3}
        \cmidrule{4-5}
            &
            Affine Projection
            &
            No Affine Projection
            &
            Affine Projection
            &
            No Affine Projection
        \\
        \toprule
            Full
            &
            {16384\ $\left(5.22\%\uparrow\right)$}
            &
            {12544\ $\left(5.08\%\uparrow\right)$}
            &
            {10000\ $\left(4.82\%\uparrow\right)$}
            &
            {10000\ $\left(5.04\%\uparrow\right)$}
        \\
        \cmidrule{1-5}
            Causal
            &
            {12544\ $\left(4.18\%\uparrow\right)$}
            &
            {5184\ $\left(4.14\%\uparrow\right)$}
            &
            {2048\ $\left(7.65\%\uparrow\right)$}
            &
            {1024\ $\left(8.36\%\uparrow\right)$}
        \\
        \bottomrule
        \\
    \end{tabular}
    }
    \caption{
        \textsc{FlashSigmoid} along with LayerNorm vs. \textsc{FlashAttention2} on A100 GPUs.
        Based on benchmarking on a set of randomly sampled tokens from the range $\left[64, 60000\right]$, we report the token $T^\ast$ after which \textsc{FlashSigmoid} with normalization consistently outperforms \textsc{FlashAttention2}, along with the total CUDA time speed-up averaged over subsequent tokens ($T > T^\ast$).
    } 
    \label{fig:appendix:flash-sigmoid-plus-layer-normalization-torchcompile}
    \end{sc}
\end{table}

\section{Contributions}
\label{sec:attribution}

All authors contributed to writing this paper, designing the experiments, discussing results at each stage of the project.

\paragraph{Preliminary work} 
Preliminary viability of $\sigmoidattn$ done by Jason Ramapuram.

\paragraph{Universal Function Approximation} 
Proof of UFA (\Cref{sec:ufa,app:UAP_proof}) sculpted by Federico Danieli.

\paragraph{Lipschitzness of Sigmoid Attention}
Lipschitzness analysis (\Cref{sec:regularity,app:lipschitz_proof}) molded by Pierre Ablin.

\paragraph{FlashSigmoid}
Implementation and analysis driven by Eeshan Dhekane in collaboration with Jagrit Digani (\Cref{sec:FlashSigmoidHardwareAwareImplementation,sec:DetailsOfFlashSigmoid}).

\paragraph{Bias Analysis}
Theoretical grounding for bias (\Cref{app:sigmoid_bias}) done by Amitis Shidani in discussion with Pierre Ablin.

\paragraph{Language Modeling Results}
All large scale language model pretraining and evaluation (\Cref{sec:llm,sec:llm_appendix}) driven by Floris Weers.

\paragraph{Stability Analysis}
QK norm (\Cref{fig:qk_norm_ablation}), LayerScale (\Cref{fig:layerscale_ablation}) and bias (\Cref{fig:const_attn_bias_ablation}) ablations crafted by Dan Busbridge using Attention Simulator. Attention Simulator written by Jason Ramapuram and used to validate norm growth (\Cref{fig:rope_vs_sincos,fig:rope_vs_rope,fig:rope_vs_alibi,fig:rope_vs_rope_b-10,fig:seq_len_scaling,fig:sqrt_scaling}).

\paragraph{ASR Results}
All ASR experiments (\Cref{sec:asr}) and ablations (\Cref{sec:asr_appendix_ablations_and_results}) are conducted by Tatiana Likhomanenko in discussions with Jason Ramapuram and Zijin Gu. Baseline ASR models code is written by Zijin Gu and Tatiana Likhomanenko. Baseline models are optimized by Zijin Gu to be close to state-of-the-art results.

\paragraph{Vision Results}
All vision experiments (\Cref{sec:supervised_image_classification,sec:ssl}) and ablations (\Cref{fig:imagenet_top_1_ablations,app:top1_results,fig:layerscale_free_sigmoid,fig:attention_relaxations}) conducted and written by Jason Ramapuram.

\paragraph{Simple Experiments}
Simple experiments to compare $\sigmoidattn$ to $\softmaxattn$, including visualizing attention evolution and simple sequence length generalization analysis (\Cref{sec:appendix_simple_experiments}) conducted by Russ Webb.

\end{document}